\documentclass[10pt,twocolumn,letterpaper]{article}

\usepackage{cvpr}              

\usepackage{amsmath}
\usepackage{amsthm}
\usepackage{booktabs}
\usepackage{algorithm}
\usepackage{algpseudocode}
\usepackage{graphicx}
\usepackage{subcaption}
\usepackage{multirow}
\usepackage{adjustbox}
\usepackage{enumitem}
\usepackage{float}
\usepackage{makecell}

\usepackage{microtype}

\definecolor{cvprblue}{rgb}{0.21,0.49,0.74}
\usepackage[pagebackref,breaklinks,colorlinks,allcolors=cvprblue]{hyperref}

\def\paperID{*****} 
\def\confName{CVPR}
\def\confYear{2026}

\title{Let Triggers Control: Frequency-Aware Dropout for Effective Token Control}

\author{Junyoung Koh\textsuperscript{1,2}\thanks{Corresponding author}\quad
Hoyeon Moon\textsuperscript{1} \quad
Dongha Kim\textsuperscript{1} \quad
Seungmin Lee\textsuperscript{1} \quad
Sanghyun Park\textsuperscript{1} \quad
Min Song\textsuperscript{1,2}\\
\textsuperscript{1}Yonsei University \quad \textsuperscript{2}Onoma AI\\
{\tt\small solbon1212@yonsei.ac.kr}
}

\begin{document}
\maketitle
\begin{abstract}
Text-to-image models such as Stable Diffusion have achieved unprecedented levels of high-fidelity visual synthesis. As these models advance, personalization of generative models---commonly facilitated through Low-Rank Adaptation (LoRA) with a dedicated trigger token---has become a significant area of research.
Previous works have naively assumed that fine-tuning with a single trigger token to represent new concepts. However, this often results in poor controllability, where the trigger token alone fails to reliably evoke the intended concept.
We attribute this issue to the frequent co-occurrence of the trigger token with the surrounding context during fine-tuning, which entangles their representations and compromises the token's semantic distinctiveness.
To disentangle this, we propose Frequency-Aware Dropout (FAD)---a novel regularization technique that improves prompt controllability without adding new parameters. FAD consists of two key components: co-occurrence analysis and curriculum-inspired scheduling.
Qualitative and quantitative analyses across token-based diffusion models (SD~1.5 and SDXL) and natural language--driven backbones (FLUX and Qwen-Image) demonstrate consistent gains in prompt fidelity, stylistic precision, and user-perceived quality.
Our method provides a simple yet effective dropout strategy that enhances controllability and personalization in text-to-image generation. Notably, it achieves these improvements without introducing additional parameters or architectural modifications, making it readily applicable to existing models with minimal computational overhead.
\end{abstract}

\begin{figure}[t]
    \centering
    \includegraphics[width=0.95\linewidth]{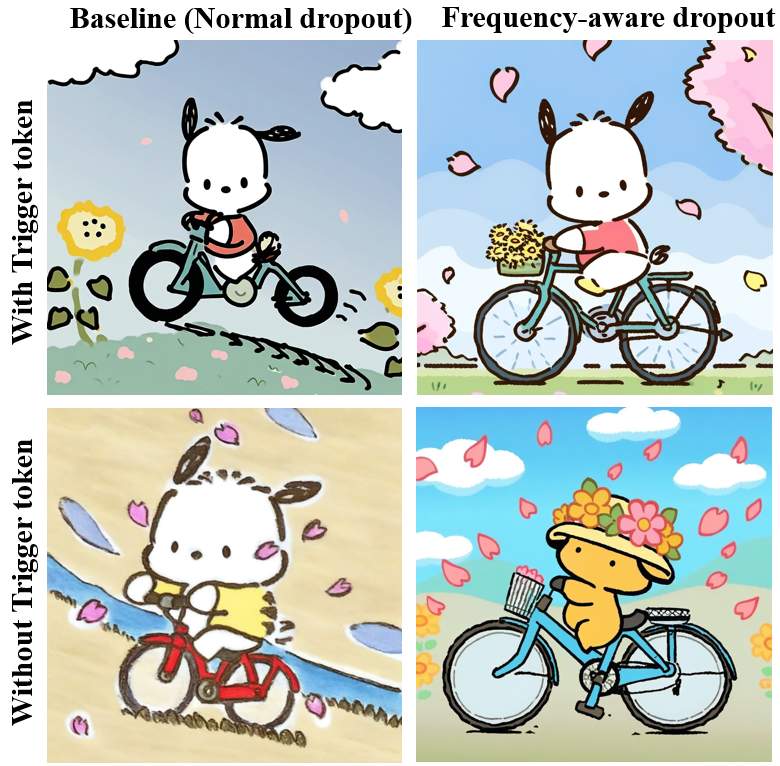}
    \caption{Comparison of images generated using Normal Dropout and Frequency-Aware Dropout with the prompt: \texttt{pochacco, riding a bike, sunny day, flower petals}, where \texttt{pochacco} serves as the trigger token.}
    \label{fig:fig1}
\end{figure}
\begin{figure*}[t]
    \centering
    \includegraphics[width=0.9\textwidth]{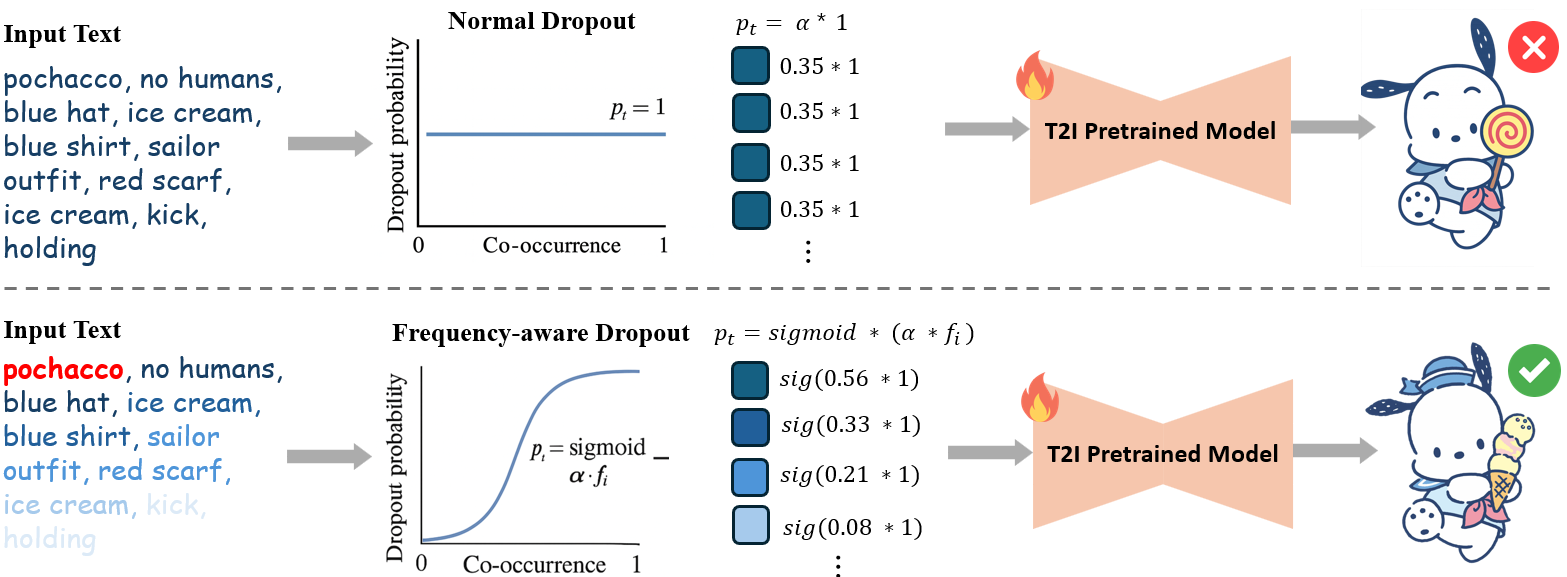}
    \caption{Comparison between Normal Dropout (top) and Frequency-Aware Dropout (FAD) (bottom).
    \emph{Top}: Normal Dropout lets style cues scatter and the trigger token is ignored.
    \emph{Bottom}: FAD raises the dropout of tokens that often co-occur with the trigger, binding the style to the trigger token $t$.}
    \label{fig:fig2}
\end{figure*}

\section{Introduction}

Text-to-Image (T2I) generation models~\cite{ramesh2022hierarchicaltextconditionalimagegeneration,ramesh2021zeroshottexttoimagegeneration,saharia2022photorealistictexttoimagediffusionmodels,esser2024scalingrectifiedflowtransformers,gao2024luminat2xtransformingtextmodality,mou2023t2iadapterlearningadaptersdig,labs2025flux1kontextflowmatching} have demonstrated remarkable capabilities in synthesizing high-resolution images with faithful adherence to textual descriptions. These models learned semantic priors from large-scale image--caption datasets, enabling them to associate and compose diverse concepts effectively.
However, this generalization~\cite{podell2023sdxlimprovinglatentdiffusion,rombach2022highresolutionimagesynthesislatent,c:22} comes at the cost of \emph{subject specificity}: the base model alone cannot reliably reproduce the
fine-grained visual identity of a particular person or character from a small set of reference images.

Consequently, personalization techniques are required to adapt pretrained diffusion models toward user-defined concepts.
To address this limitation, DreamBooth~\cite{dreambooth} fine-tunes a diffusion model on a few subject-specific images and
introduces a dedicated \emph{trigger token} that binds the subject's identity to a single textual representation.
Building upon this idea, Low-Rank Adaptation (LoRA)~\cite{hu2021loralowrankadaptationlarge} has become a standard parameter-efficient fine-tuning (PEFT) strategy for T2I personalization, allowing models to efficiently learn new
concepts without updating the full network weights.
Despite their effectiveness, these methods implicitly assume that the trigger token alone can represent the target subject~\cite{houlsby2019parameterefficienttransferlearningnlp,li2021prefixtuningoptimizingcontinuousprompts,zaken2022bitfitsimpleparameterefficientfinetuning}.
In practice, however, the trigger token frequently co-occurs with descriptive context words in training captions
(e.g., \textit{red hat}, \textit{asian}, \textit{male}),
which entangles the learned identity across multiple tokens and weakens the token's independent controllability.

This entanglement problem is exacerbated in PEFT settings, where limited training data and frozen backbones
encourage the LoRA~\cite{hu2021loralowrankadaptationlarge} parameters to memorize correlations rather than isolating the subject into the trigger token itself.
As a result, providing the trigger token alone during inference often fails to evoke the desired identity or style~\cite{HyperDreamBooth,StyleGANT}.

Following the DreamBooth convention within LoRA-based personalization~\cite{zhang2024personalizedlorahumancenteredtext}, a trigger token \textit{t} is typically defined so that providing \textit{t} alone during inference can evoke the desired style.
However, when \textit{t} frequently co-occurs with other terms such as \textit{red hat} or \textit{human} in captions, the model tends to bind crucial style-related information mapping between \textit{t} and its surrounding tokens~\cite{wu2024contrastivepromptsimprovedisentanglement,xu2024cusconceptcustomizedvisualconcept,park2025fairgenerationunfairdistortions,wang2024tokencomposetexttoimagediffusiontokenlevel}, making \textit{t} alone insufficient for reliable control~\cite{sohn2023styledroptexttoimagegenerationstyle,brooks2023instructpix2pixlearningfollowimage,dong2025dreamartistcontrollableoneshottexttoimage}.

In this paper, we propose a \textbf{Frequency-Aware Dropout} that strengthens the control of the trigger token.
We dynamically adjust the dropout rate during training to partially block the gradient flow of tokens whose co-occurrence frequencies with trigger token have been precomputed.
This encourages the model to focus on the style representation within the trigger token itself.
Our proposed method consists of the following key contributions:
\begin{itemize}
\raggedright
    \item It operates without additional computational overhead or any modification of the model architecture, which solely relies on statistical information extracted from the training data.
    \item It remarkably enhances the controllability of the trigger token, thereby preserving high stylistic consistency even when the prompt consists of the trigger token by itself.
\end{itemize}

\section{Preliminaries}
\subsection{Diffusion-based Text-to-Image Generation}
Let $x_0$ be an image sampled from the data distribution $p_{\text{data}}$:
\begin{equation}
x_0 \sim p_{\text{data}}(x).
\label{eq:sample_from_data}
\end{equation}

A diffusion model defines a Markov chain that gradually perturbs $x_0$ with Gaussian noise, producing $\{x_t\}_{t=1}^T$.
Following DDPM~\cite{ddpm}, define a variance schedule $\{\beta_t\}_{t=1}^T$ with $\alpha_t = 1-\beta_t$ and the cumulative product
$\bar{\alpha}_t = \prod_{s=1}^{t}\alpha_s$.
The forward noising process is:
\begin{equation}
q(x_t \mid x_{t-1}) = \mathcal{N}\!\left(x_t;\sqrt{\alpha_t}\,x_{t-1},\, (1-\alpha_t)I\right),
\end{equation}
which admits the closed form:
\begin{equation}
q(x_t \mid x_0)=\mathcal{N}\!\left(x_t;\sqrt{\bar{\alpha}_t}\,x_0,\,(1-\bar{\alpha}_t)I\right),
\end{equation}
\begin{equation}
x_t = \sqrt{\bar{\alpha}_t}\,x_0 + \sqrt{1-\bar{\alpha}_t}\,\epsilon,
\quad \epsilon\sim\mathcal{N}(0,I).
\label{eq:forward_closed_form}
\end{equation}

A neural network $\epsilon_\theta(x_t, t, y)$ is trained to predict the noise $\epsilon$ at diffusion timestep $t \in \{1,\dots,T\}$,
conditioned on a text embedding $y$~\cite{diffusionbeatsgan,ddpm,ddim}.
In Stable Diffusion, the diffusion process is applied in a learned latent space via a KL-regularized VAE~\cite{razavi2019generatingdiversehighfidelityimages}, as in Latent Diffusion Models.

\subsection{Low-Rank Adaptation}
Fine-tuning the full parameters of a diffusion U-Net~\cite{ronneberger2015unetconvolutionalnetworksbiomedical} is computationally expensive.
LoRA mitigates the resource demands by introducing trainable low-rank matrices while freezing the original weights $W \in {\rm I\!R}^{d \times d}$ and introducing a learnable update $\Delta W = BA$ where $A \in {\rm I\!R}^{r \times d}$ and $B \in {\rm I\!R}^{d \times r}$ with rank $r \ll d$.
However, despite LoRA's parameter-efficiency, prior work has shown that under low-data regimes its additional low-rank parameters can still memorize the limited training set, leaving LoRA-based fine-tuning prone to overfitting~\cite{loradropout}.

\subsection{Trigger Token}
Prior studies have highlighted the role of the trigger token, referring to it as a ``pseudo-word token'' in Textual Inversion~\cite{textualinversion}, a ``unique identifier token'' in Dreambooth, and a ``pseudo token'' in Custom Diffusion.
The trigger token \textit{t} is either randomly initialized~\cite{kocmi2017explorationwordembeddinginitialization} or derived from a semantically adjacent token in the text encoder~\cite{pang2023crossinitializationpersonalizedtexttoimage,Dobler_2023}.
To serve effectively as an independent control signal, \textit{t} must (1) exhibit higher variance in its cross-attention scores $A_t$ relative to neighboring tokens~\cite{ren2025unveilingmitigatingmemorizationtexttoimage,zhang2024enhancingsemanticfidelitytexttoimage}, and (2) possess a sufficiently large gradient norm while maintaining a low similarity coefficient $\rho$ with other token embeddings.

\section{Related Work}

\paragraph{Parameter-Efficient Fine-Tuning for Diffusion Models.}
LoRA~\cite{hu2021loralowrankadaptationlarge} achieves near full fine-tuning performance while substantially reducing computational overhead, making it the de facto standard for text-to-image personalization~\cite{xu2023parameterefficientfinetuningmethodspretrained}.
DoRA~\cite{liu2024dora} further bridges the gap by analytically decomposing weight matrices, while LyCORIS~\cite{lycoris} incorporates unconventional decomposition algorithms such as the Kronecker product to expand the design space of rank adaptation.
Recent efforts continue to improve LoRA efficiency and generalization~\cite{cheng2025revisitingloralensparameter,wang2024loragalowrankadaptationgradient,soboleva2025tlorasingleimagediffusion}.
However, these methods primarily operate at the parameter level and do not address how individual input tokens interact with the adapted weights during training.

\paragraph{Token--Object Alignment in Personalization.}
A separate line of work focuses on aligning specific tokens with their intended visual concepts.
MC$^2$~\cite{jiang2024mc2multiconceptguidancecustomized} mitigates token--object interference in multi-concept generation by adaptively reweighting background separation through token-wise mask tuning.
NaviDet~\cite{navidet} demonstrates that a single trigger token can induce distinctive neural activations during early diffusion steps, highlighting the significant influence of input-level tokens on the generation process.
LoRA-Dropout~\cite{loradropout} alleviates overfitting and promotes sparsity by applying dropout to the low-rank matrices themselves.
While these approaches advance token-level understanding and regularization, none of them directly addresses the controllability loss that arises when a trigger token frequently co-occurs with contextual words during fine-tuning~\cite{wu2024contrastivepromptsimprovedisentanglement,xu2024cusconceptcustomizedvisualconcept,park2025fairgenerationunfairdistortions,wang2024tokencomposetexttoimagediffusiontokenlevel}.

\paragraph{Our Approach.}
Although empirical analyses suggest that trigger tokens lose controllability due to co-occurrence with surrounding words~\cite{shen2024promptstealingattackstexttoimage,witteveen2022investigatingpromptengineeringdiffusion,chefer2023hiddenlanguagediffusionmodels,yu2024imageworth32tokens}, systematic input-level interventions remain scarce.
We introduce Frequency-Aware Dropout into LoRA training as a principled approach that directly addresses co-occurrence entanglement, enabling consistent style reproduction with a single trigger token without architectural modifications.

\section{Methodology}
We propose \textbf{Frequency-Aware Dropout (FAD)}, which ensures the characteristic information is exclusively tied to the newly introduced trigger token.
As illustrated in \cref{fig:fig2}, FAD scans training captions to identify tokens that often co-occur with trigger and assigns them higher dropout rates, thereby putting all stylistic cues into the single token~$t$.
This selective suppression binds the model to invoke $t$---and $t$ alone---whenever the target feature is required.
The key idea is to concentrate the learned feature information in a newly introduced trigger token $t$, so that this single token can reproduce the desired feature on its own.
Algorithm~\ref{alg:overall_algorithm} summarizes the overall procedure with Frequency-Aware Dropout.

The dropout probability for each token $w$ is defined as:
\begin{equation}
\begin{aligned}
p_{\text{drop}}(w) &= p_{\min} + \Delta p\ *\sigma\bigl(\alpha\,[r(w)-c]\bigr)\\
\end{aligned}
\label{eq:dropout}
\end{equation}

\begin{equation}
\begin{aligned}
\Delta p\ = p_{\max} - p_{\min}
\\
\end{aligned}
\label{eq:dropout_minmax}
\end{equation}

In \cref{eq:dropout} and \cref{eq:dropout_minmax}, $p_{\min}$ and $p_{\max}$ denote the minimum and maximum dropout probabilities, respectively, $c$ is the center ratio at which the sigmoid function begins to rise most sharply, and $\alpha$ controls the sigmoid's slope. This means when $r(w)$ with the trigger token is close to $c$, its dropout probability starts to increase rapidly.
Consequently, a word that almost always co-occurs with the trigger token receives a dropout probability $p_{\text{drop}}(w)$ close to $p_{\max}$, whereas a rare word gets $p_{\text{drop}}(w)$ near $p_{\min}$.
We do not apply dropout to the trigger token itself, ensuring that $t$ is always present in the input.

For LoRA fine-tuning, we apply these dropout probabilities with a step-wise scaling factor $p_{\text{step}}(i)$ that varies over the training steps $i$.
The factor $p_{\text{step}}(i)\in[0,1]$ starts at 0 and gradually increases to 1 as training progresses.
We begin with no dropout for an initial warm-up period to allow the model to first adapt to the new concept and then introduce dropout after step $i_{\text{w}}$, increasing $p_{\text{step}}(i)$ until it reaches 1 at step $i_{\text{f}}$, where $i_{\text{w}}$ is the step when the dropout begins and $i_{\text{f}}$ is the step when the dropout scale reaches its maximum value.
A representative schedule is defined as follows:
\begin{equation}
p_{\text{step}}(i)=
\begin{cases}
0, & i<i_w,\\
1-e^{-\beta\,s}, & i_w\le i<i_f,\\
1, & i\ge i_f
\end{cases}\quad
s=\dfrac{i-i_w}{i_f-i_w}
\label{eq:step}
\end{equation}

Using the schedule in \cref{eq:step} as a multiplier, at each training step $i$ we independently drop each token $w$ except the trigger token with probability $p_{\text{drop}}(w)\cdot p_{\text{step}}(i)$.
In other words, the caption fed into the text encoder at step $i$ stochastically omits each non-trigger word according to its probability, leaving only the trigger token in later stages.
As training progresses, the cross-attention module is thus forced to rely almost solely on the trigger token for the feature representation, guiding the model to bind the feature information in the trigger token.

Our proposed algorithm performs simple dropout operations guided by statistics derived from the captions. It introduces no additional GPU-bound computational overhead while significantly improving the controllability of the trigger token.
Consequently, after fine-tuning, even a prompt containing only the trigger token can produce images in the learned feature.

\begin{algorithm}[t]
\caption{Frequency-Aware Dropout}
\textbf{Input}: Caption set $\mathcal{D}$, trigger token $t$, total training steps $N$

\begin{algorithmic}[1]
  \State \textbf{Compute Co-occurrence Frequency:}
  \Statex \vspace{0.2em}
  \State \quad $\mathcal{C} \gets \{C \in \mathcal{D} \mid t \in C\}$ \Comment{Captions containing $t$}
  \State \quad $N_t \gets |\mathcal{C}|$ \Comment{\# captions containing the trigger}
  \State \quad $\text{count}[w] \gets \sum_{C\in\mathcal{C}}\mathbf{1}_{w\in C}$ \Comment{Co-occurrence count}
  \State \quad $r(w) \gets \text{count}[w] / N_t$
  \State \quad $p_{\text{drop}}(w) \gets f_{\text{drop}}\bigl(r(w)\bigr)$ \Comment{\cref{eq:dropout}}
  \State \quad $p_{\text{drop}}(t) \gets 0$

  \Statex \vspace{0.5em}
  \State \textbf{Training with Frequency-Aware Dropout:}
  \For{$i = 1$ \textbf{to} $N$}
      \State \quad $p_{\text{step}} \gets \text{schedule}(i/N)$ \Comment{\cref{eq:step}}
      \State \quad Sample a training caption $C$
      \ForAll{tokens $w \in C$ with $w \neq t$}
          \State \quad \quad Drop $w$ with probability $p_{\text{drop}}(w) \cdot p_{\text{step}}$
      \EndFor
  \EndFor
\end{algorithmic}
\label{alg:overall_algorithm}
\end{algorithm}

\begin{table*}[t]
\centering
\small
\caption{FID ($\downarrow$) and DINO ($\uparrow$) across different diffusion models. Three different approaches are compared: Normal Dropout, Frequency-Aware Dropout (FAD) and Frequency-aware dropout with time-step (sFAD). Lower FID and higher DINO are better.
\textit{For \textbf{pikachu} and \textbf{pochacco}, we fine-tune LoRA on the SDXL-based Illustrious model (instead of vanilla SDXL).}}
\setlength{\tabcolsep}{3pt}
\begin{adjustbox}{max width=\textwidth}
\begin{tabular}{llcccccccccccc}
\toprule
\multicolumn{2}{c}{\textbf{Dataset}}
& \multicolumn{2}{c}{\textbf{faker}}
& \multicolumn{2}{c}{\textbf{reeves}}
& \multicolumn{2}{c}{\textbf{hsng}}
& \multicolumn{2}{c}{\textbf{mbst}}
& \multicolumn{2}{c}{\textbf{pikachu}}
& \multicolumn{2}{c}{\textbf{pochacco}} \\
\cmidrule(lr){3-4} \cmidrule(lr){5-6} \cmidrule(lr){7-8} \cmidrule(lr){9-10} \cmidrule(lr){11-12} \cmidrule(lr){13-14}
\textbf{Model} & \textbf{Method}
& \textbf{FID ($\downarrow$)} & \textbf{DINO ($\uparrow$)}
& \textbf{FID ($\downarrow$)} & \textbf{DINO ($\uparrow$)}
& \textbf{FID ($\downarrow$)} & \textbf{DINO ($\uparrow$)}
& \textbf{FID ($\downarrow$)} & \textbf{DINO ($\uparrow$)}
& \textbf{FID ($\downarrow$)} & \textbf{DINO ($\uparrow$)}
& \textbf{FID ($\downarrow$)} & \textbf{DINO ($\uparrow$)} \\
\midrule
\multirow{3}{*}{SD 1.5}
& Normal Dropout   & 207.10 & 0.897  & 210.53 & 0.924  & 164.25 & 0.896  & 220.03 & 0.898  & 137.47 & 0.915  & 136.01 & 0.940  \\
& FAD              & 197.46 & 0.907  & 203.07 & 0.929  & 161.20 & 0.893  & 214.13 & 0.903  & 137.76 & 0.917  & 133.72 & \textbf{0.944} \\
& sFAD             & \textbf{195.86} & \textbf{0.910}  & \textbf{198.52} & \textbf{0.932}  & \textbf{160.47} & \textbf{0.893}  & \textbf{212.87} & \textbf{0.907}  & \textbf{136.74} & \textbf{0.922}  & \textbf{131.51} & 0.943  \\
\midrule
\multirow{3}{*}{SDXL}
& Normal Dropout   & 199.31 & 0.9037 & 208.11 & 0.9238 & 166.35 & 0.9013 & 236.94 & 0.9025 & 170.04 & 0.8509 & 193.02 & 0.8582 \\
& FAD              & \textbf{194.69} & 0.9061 & \textbf{196.30} & 0.9322 & \textbf{160.80} & 0.9005 & \textbf{216.49} & 0.9115 & \textbf{160.95} & 0.8628 & \textbf{182.56} & \textbf{0.8828} \\
& sFAD             & 197.76 & \textbf{0.9107} & 199.09 & \textbf{0.9325} & 162.65 & \textbf{0.9034} & 217.34 & \textbf{0.9124} & 164.22 & \textbf{0.8671} & 184.08 & 0.8668 \\
\midrule
\multirow{3}{*}{FLUX}
& Normal Dropout           & 168.09 & 0.6754 & 187.64 & 0.6702 & 	142.11     & 0.6945     & 180.26 & 0.6671 & 123.92 & 0.8064 & \textbf{119.47} & 0.8408 \\
& FAD    & 168.31 & 0.6638 & 187.79 & 0.6744 & 	135.36     & 0.7219     & 171.30 & 0.6817 & 114.77 & 0.8273 & 121.09 & \textbf{0.8486} \\
& sFAD   & \textbf{162.23} & \textbf{0.7135} & \textbf{181.93} & \textbf{0.7015} & \textbf{129.47}     & \textbf{0.7716}     & \textbf{171.04} & \textbf{0.7149} & \textbf{110.78} & \textbf{0.8370} & 124.27 & 0.8290 \\
\midrule
\multirow{3}{*}{Qwen-Image}
& Normal Dropout
& 175.52 & 0.6373
& 201.83 & 0.6543
& 138.13 & 0.7347
& 175.47 & 0.6784
& 115.99 & 0.8255
& 148.20 & \textbf{0.8606} \\
& FAD
& 174.81 & \textbf{0.6427}
& 192.62 & 0.6502
& 133.64 & 0.7297
& 175.99 & 0.6822
& 127.63 & 0.8156
& 145.39 & 0.8487 \\
& sFAD
& \textbf{174.75} & 	0.6385
& \textbf{181.95} & \textbf{0.7015}
& \textbf{133.06} & \textbf{0.7371}
& \textbf{171.02} & \textbf{0.7149}
& \textbf{110.74} & \textbf{0.8370}
& \textbf{124.23} & 0.8290 \\
\bottomrule
\end{tabular}
\label{tab:fid_dino_table}
\end{adjustbox}
\end{table*}

\begin{table*}[t]
\centering
\caption{Comprehensive Evaluation: InsightFace ($\downarrow$), CCIP ($\uparrow$), and GPT-4.1 Evaluation (CS / CI). Lower is better for InsightFace, higher is better for CCIP and GPT-4.1 Evaluation scores.}
\begin{adjustbox}{max width=\textwidth}
\begin{tabular}{llcccccc}
\toprule
\multicolumn{2}{c}{\textbf{Dataset}}
& \textbf{faker} & \textbf{reeves} & \textbf{hsng} & \textbf{mbst} & \textbf{pikachu} & \textbf{pochacco} \\
\cmidrule(lr){3-6} \cmidrule(lr){7-8}
\textbf{Model} & \textbf{Method} & \multicolumn{4}{c}{\textbf{InsightFace ($\downarrow$)}} & \multicolumn{2}{c}{\textbf{CCIP ($\uparrow$)}} \\
\midrule
\multirow{3}{*}{SD 1.5}
& Normal Dropout   & 28.064  & 30.619  & 27.792  & 28.015  & 0.9987 & 0.9431 \\
& FAD              & 26.997  & 30.236  & 27.511  & 28.058  & 0.9987 & \textbf{0.9528} \\
& sFAD    & \textbf{26.954}  & \textbf{30.192}  & \textbf{27.433}  & \textbf{27.835}  & \textbf{0.9987} & 0.9487 \\
\midrule
\multirow{3}{*}{SDXL}
& Normal Dropout   & 25.510  & 27.070  & 24.960  & 27.280  & 0.9992 & 0.8996 \\
& FAD              & \textbf{24.470}  & 26.290  & \textbf{24.410}  & 25.440  & 0.9996 & 0.9091 \\
& sFAD    & 25.130  & \textbf{25.930}  & 24.530  & \textbf{24.550}  & \textbf{1.0000} & \textbf{0.9295} \\
\midrule
\multirow{3}{*}{FLUX}
& Normal
& 20.19
& 21.22
& 20.98
& 19.68
& 0.9903
& 0.9674 \\
& FAD
& 20.14
& 21.43
& 20.48
& \textbf{19.53}
& 0.9909
& \textbf{0.9696} \\
& sFAD
& \textbf{19.70}
& \textbf{20.61}
& \textbf{19.34}
& 19.63
& \textbf{0.9925}
& 0.9682 \\
\midrule
\multirow{3}{*}{Qwen-Image}
& Normal
& 22.35
& 25.27
& 21.81
& 22.56
& 0.9911
& 0.9674 \\
& FAD
& 22.00
& 25.01
& 22.32
& 22.17
& 0.9897
& \textbf{0.9693} \\
& sFAD
& \textbf{21.66}
& \textbf{20.61}
& \textbf{21.63}
& \textbf{19.63}
& \textbf{0.9925}
& 0.9682 \\
\midrule
\multicolumn{8}{c}{\textbf{GPT-4.1 Evaluation (Character Similarity / Composition \& Image Quality)}} \\
\midrule
\multirow{3}{*}{SD 1.5}
& Normal Dropout    & 6.98 / 7.54 & 6.48 / 6.36 & 7.10 / 7.38 & 7.50 / \textbf{7.62} & 8.32 / 7.84 & 7.98 / 8.12 \\
& FAD               & 7.70 / 7.90 & 6.52 / 6.94 & 7.16 / 7.50 & \textbf{7.56} / 7.48 & 8.46 / 7.72 & \textbf{8.32} / \textbf{8.52} \\
& sFAD     & \textbf{7.70} / \textbf{7.90} & \textbf{6.62} / \textbf{6.98} & \textbf{7.26} / \textbf{7.56} & 7.24 / 7.24 & \textbf{8.46} / \textbf{7.84} & 8.28 / 8.38 \\
\midrule
\multirow{3}{*}{SDXL}
& Normal Dropout    & 8.44 / 8.32 & 7.46 / \textbf{7.96} & 8.12 / 8.69 & 7.57 / \textbf{9.11} & \textbf{2.64} / 8.86 & \textbf{8.62} / 8.55 \\
& FAD               & \textbf{8.66} / \textbf{8.44} & \textbf{7.64} / 7.82 & \textbf{8.59} / \textbf{9.22} & \textbf{8.08} / 8.33 & 1.48 / 8.94 & 8.41 / 8.46 \\
& sFAD     & 8.42 / 8.14 & 7.46 / 7.56 & 8.38 / 9.08 & 7.40 / 8.13 & 1.90 / \textbf{8.96} & 8.39 / \textbf{8.74} \\
\midrule
\multirow{3}{*}{FLUX}
& Normal Dropout
& 8.44 / 9.30
& \textbf{7.92} / \textbf{9.04}
& 7.83 / 9.15
& \textbf{8.38} / \textbf{9.22}
& 8.52 / 8.82
& 7.72 / 8.58 \\
& FAD
& 8.22 / 9.20
& 7.68 / 8.96
& 7.88 / 9.16
& 8.34 / 9.16
& 8.42 / 9.02
& 8.06 / \textbf{8.90} \\
& sFAD
& \textbf{8.65} / \textbf{9.33}
& 7.86 / 8.82
& \textbf{8.19} / \textbf{9.57}
& 8.04	/ 9.09
& \textbf{8.94} / \textbf{9.18}
& \textbf{8.18} / 8.16 \\
\midrule
\multirow{3}{*}{Qwen-Image}
& Normal Dropout
& 8.12 / 8.90
& 7.35 / 8.79
& 7.92 / 8.84
& 8.02 / 8.69
& \textbf{9.05} / \textbf{9.18}
& \textbf{8.51} / \textbf{8.86} \\
& FAD
& 8.04 /8.94
& 7.58 / 8.80
& 7.86 / 9.20
& 8.13 / 8.82
& 8.90 / 8.84
& 8.39 / 8.58 \\
& sFAD
& \textbf{8.20} / \textbf{9.06}
& \textbf{7.78} / \textbf{8.84}
& \textbf{8.18} / \textbf{9.22}
& \textbf{8.18}	/ \textbf{9.22}
& 8.91 / \textbf{9.18}
& 8.24 / 8.26 \\
\bottomrule
\end{tabular}
\label{tab:ccip_insight_gpt_table}
\end{adjustbox}
\end{table*}

\section{Experiments}
\subsection{Experiments Setup}
\begin{figure}[t]
    \centering
    \begin{subfigure}{0.32\linewidth}
        \centering
        \includegraphics[width=\linewidth]{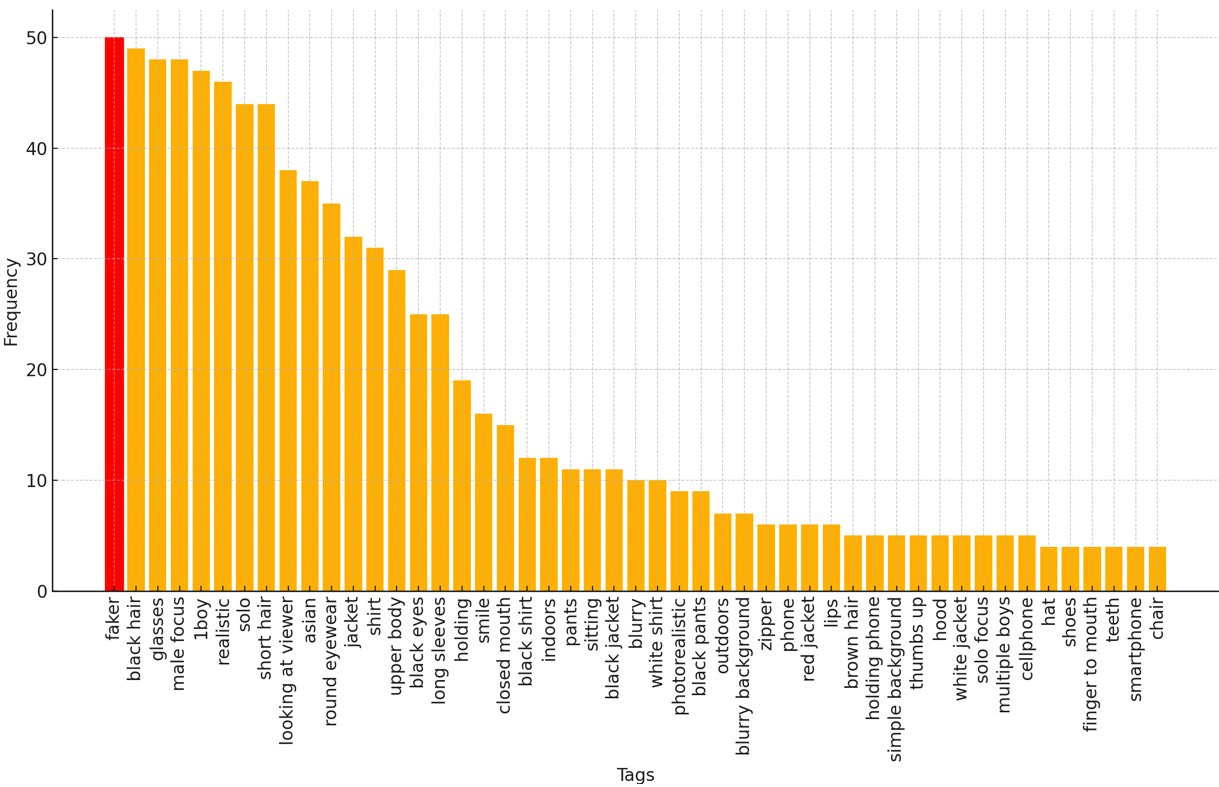}
        \caption{faker}
    \end{subfigure}%
    \hfill
    \begin{subfigure}{0.32\linewidth}
        \centering
        \includegraphics[width=\linewidth]{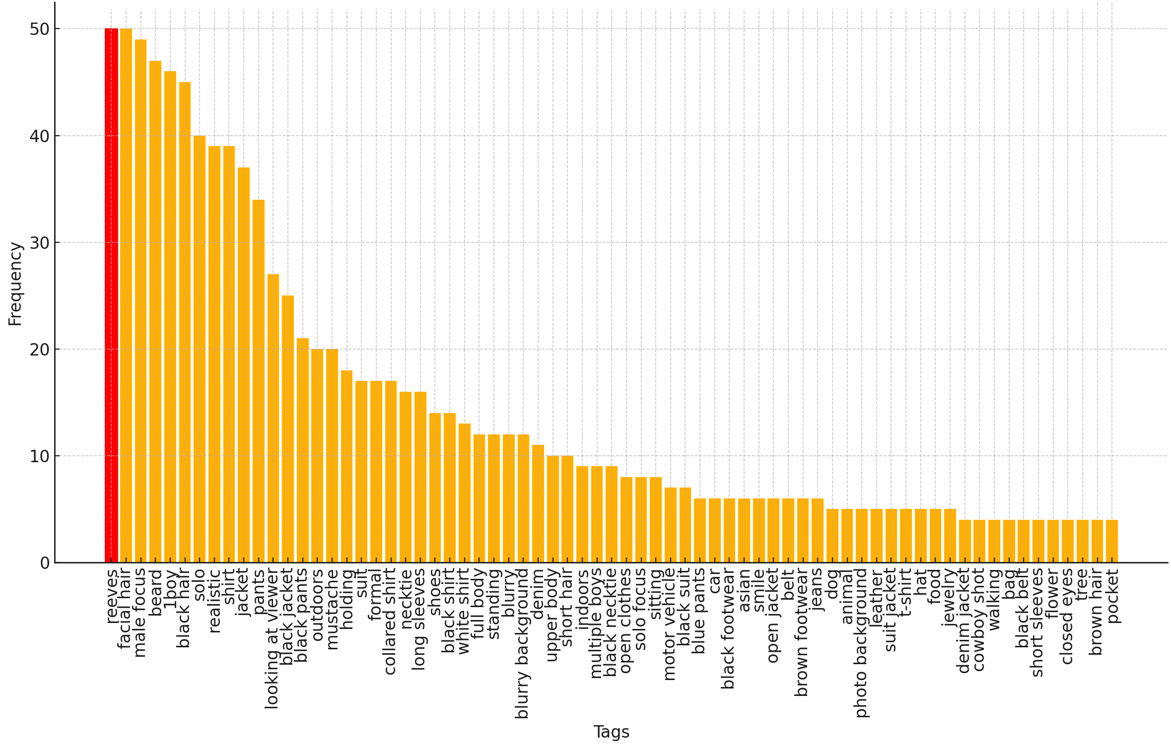}
        \caption{reeves}
    \end{subfigure}%
    \hfill
    \begin{subfigure}{0.32\linewidth}
        \centering
        \includegraphics[width=\linewidth]{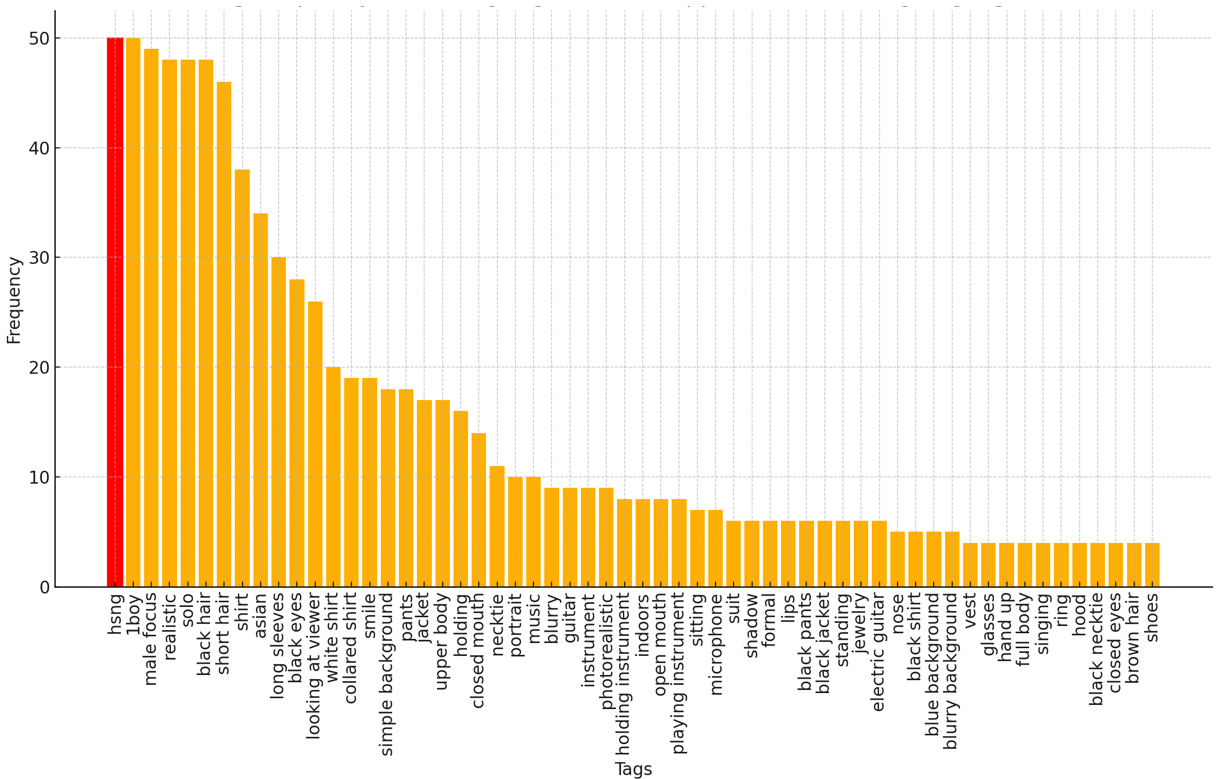}
        \caption{hsng}
    \end{subfigure}

    \begin{subfigure}{0.32\linewidth}
        \centering
        \includegraphics[width=\linewidth]{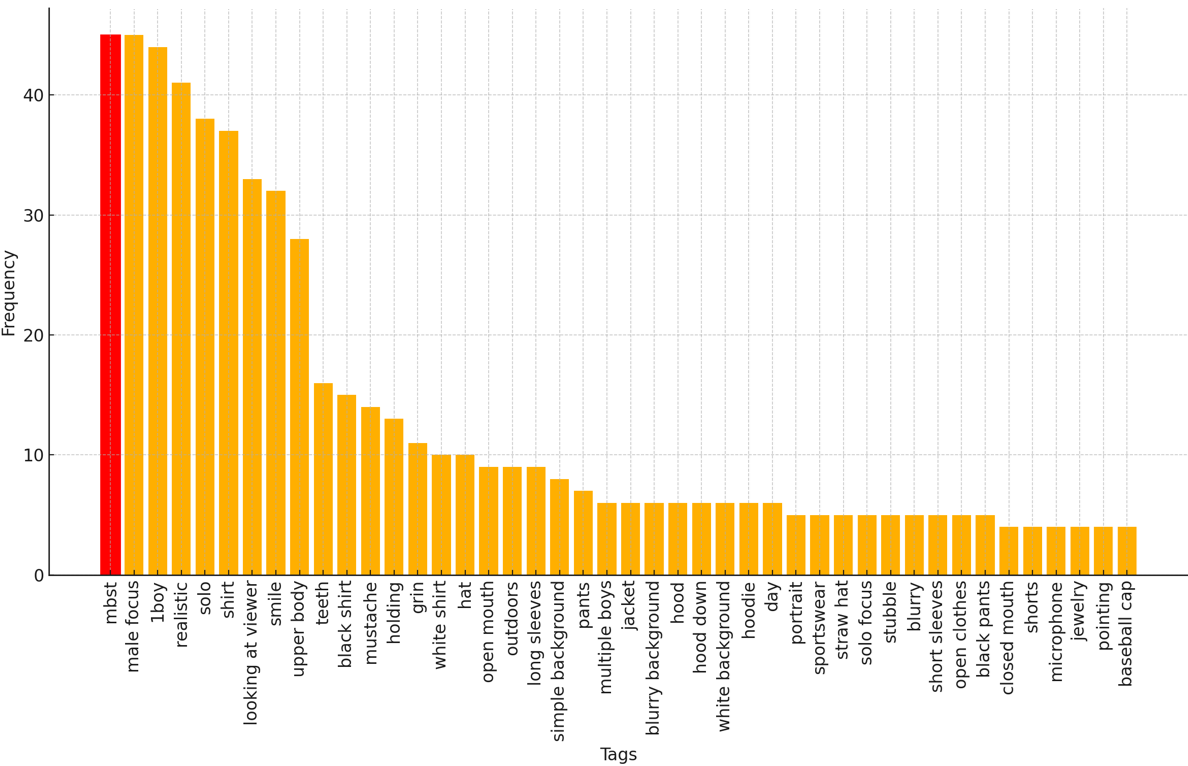}
        \caption{mbst}
    \end{subfigure}%
    \hfill
    \begin{subfigure}{0.32\linewidth}
        \centering
        \includegraphics[width=\linewidth]{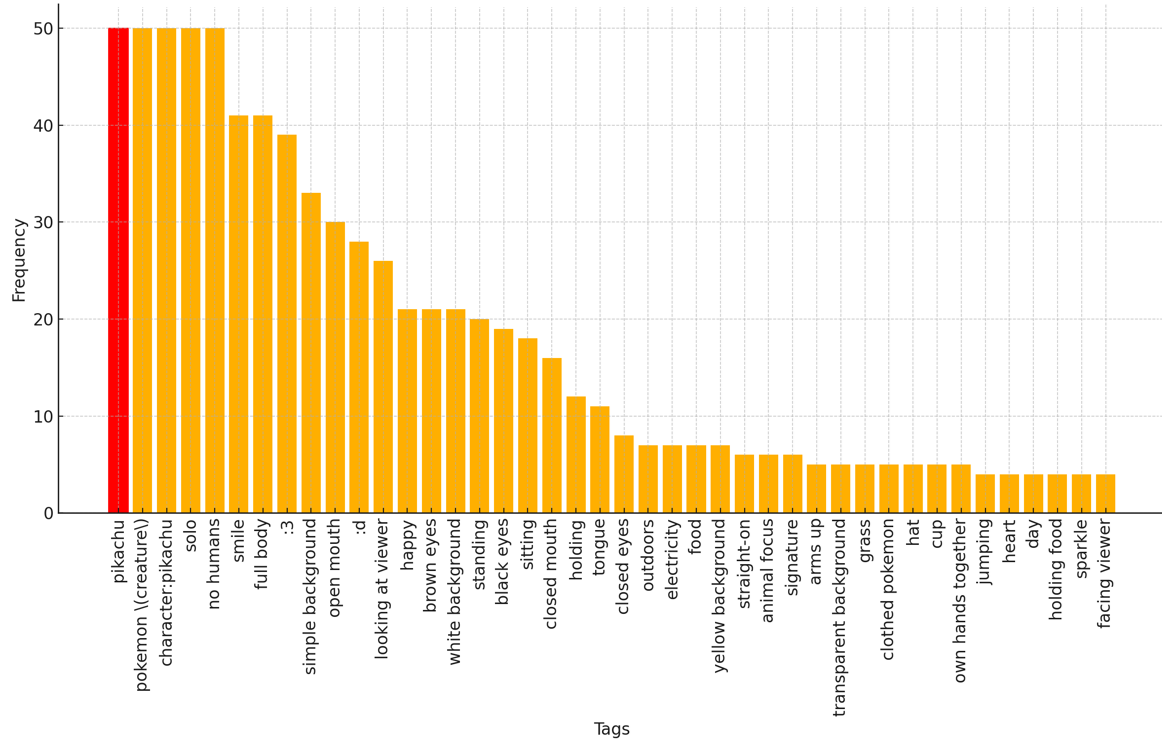}
        \caption{pikachu}
    \end{subfigure}%
    \hfill
    \begin{subfigure}{0.32\linewidth}
        \centering
        \includegraphics[width=\linewidth]{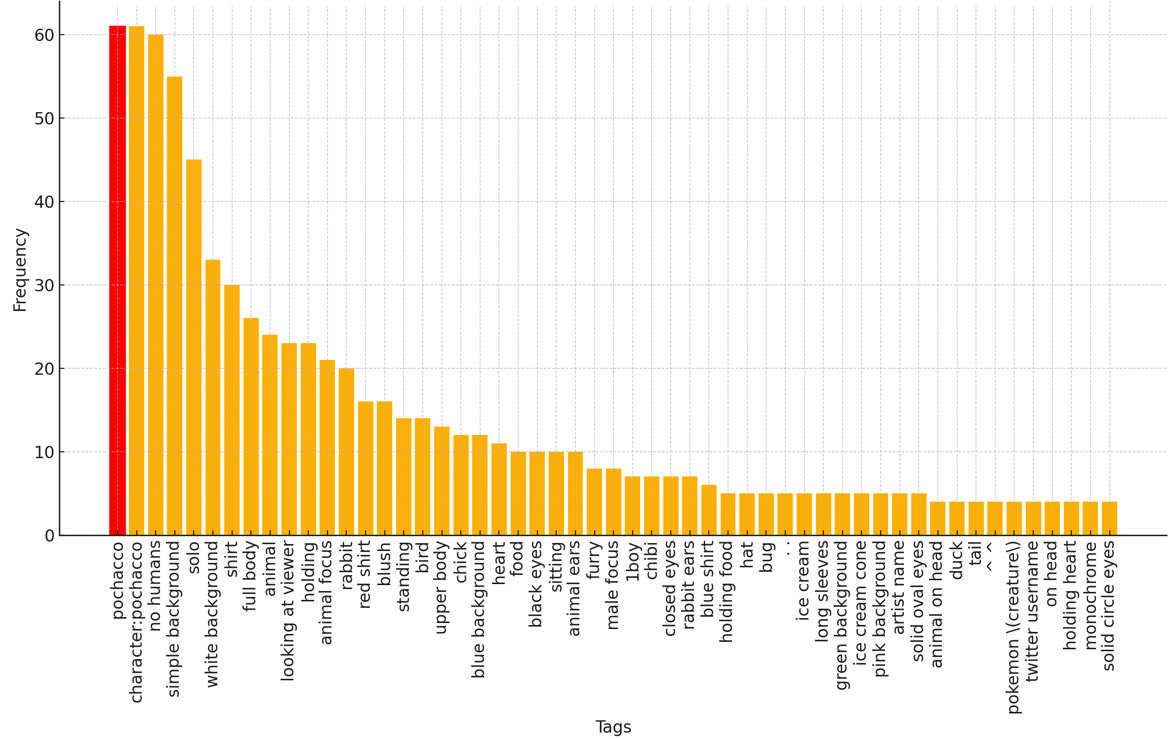}
        \caption{pochacco}
    \end{subfigure}
    \caption{Tag frequency for each dataset. The trigger token is indicated in \textcolor{red}{red}.}
    \label{fig:dataset_analysis}
\end{figure}

\paragraph{Dataset.}
We evaluate our approach on multiple personalization tasks covering both fictional and non-fictional subjects.
Our dataset consists of small image collections for several concepts: four public figures (e.g., Faker, Keanu Reeves, Mr.Beast and Gen Hoshino) and two popular character styles (e.g., Pikachu and Pochacco).
For each concept, we collect a set of 50 images from the web, ensuring diversity in poses, expressions, and backgrounds.
Following the common practice in DreamBooth and Textual Inversion, we manually filter all images to retain only high-quality samples that faithfully represent the subject's appearance or artistic feature.
This curated subset ensures a consistent and objective foundation for evaluating personalization methods.
Each concept is then assigned a unique trigger token and the corresponding text-to-image models are fine-tuned on its curated dataset to allow the trigger token alone to evoke the target feature.
\cref{fig:dataset_analysis} illustrates the tag frequency distribution across all six datasets; higher dropout is applied to more frequent tags, while rare tags receive lower dropout.

\paragraph{Model.}
Our experiments cover multiple diffusion backbones to validate that Frequency-Aware Dropout generalizes beyond a single architecture.
For photo-realistic generation, we adopt Stable Diffusion v1.5 (SD 1.5) and Stable Diffusion XL 1.0 (SDXL).
For anime-style generation, we use Illustrious-XL v2.0~\cite{park2024illustriousopenadvancedillustration}, an SDXL-based model.
In addition, we evaluate our method on more recent diffusion backbones, namely FLUX and Qwen-Image~\cite{qwenimage,flux2024,labs2025flux1kontextflowmatching}, to test robustness across different training recipes and architectures.

For all backbones, all pre-trained weights (including text encoder~\cite{CLIP,Ilharco_Open_Clip_2021} and diffusion U-Net) remain frozen, except for the newly introduced LoRA parameters and the learned trigger-token embedding.
Unless otherwise specified, each concept is fine-tuned for 1,500 steps per concept.
During inference, we generate images with classifier-free guidance~\cite{ho2022classifierfreediffusionguidance} (guidance scale 7.5) using the LoRA-fine-tuned model.

We evaluate three variants: (a) \textbf{Normal Dropout}, the baseline without any dropout; (b) \textbf{FAD}, which applies co-occurrence-based dropout to caption tokens; and (c) \textbf{sFAD}, where FAD gradually increases over training steps.
This last variant corresponds to the whole method described in the method section, where $p_{\text{step}}(i)$ starts from near zero and exponentially increases to 1, thereby applying almost no dropout in early iterations and complete dropout toward the end of training.
The training hyperparameters are summarized in the supplementary material (\cref{tab:hyperparameters_sd,tab:hyperparameters_flux_qwen}).

\paragraph{Metrics.}
To measure the quality of personalization, the following metrics are measured. All reported metric scores are averaged over inference outputs from all training steps.
\textbf{Fr\'{e}chet Inception Distance (FID)}~\cite{heusel2018ganstrainedtimescaleupdate} quantifies the distributional discrepancy in feature space between generated and real images, where lower scores denote higher visual fidelity.
We further report the \textbf{DINO similarity score}~\cite{DINO}, an embedding-based metric that measures how well each output retains the subject's distinctive visual attributes; larger cosine similarities indicate closer resemblance to the reference in feature or identity~\cite{mi2025datasynthesisdiversestyles,song2024diffsimtamingdiffusionmodels}.

For human faces, we use \textbf{InsightFace}~\cite{insightface,Deng_2022,wang2021facexzoopytorchtoolboxface},
defined as the average cosine \emph{distance} between embeddings of generated images and those of the subject's real images
(i.e., $1-\cos(\cdot,\cdot)$), where lower values indicate better identity preservation.
Character identity is evaluated with \textbf{Contrastive Character Image Pretraining (CCIP)}~\cite{CCIP}, where the trained model extracts the CLIP-based feature representation and measures the similarity between reference character and inputs, which represents stronger character preservation. We also utilize \textbf{MLLM-based Subjective Evaluation} to follow common practice of~\cite{meng2024imageregenerationevaluatingtexttoimage}. We employ \textbf{GPT-4.1}~\cite{GPT4}, along with OpenAI's latest Multimodal Large Language Model (MLLM) evaluators for T2I tasks~\cite{mao2024gptevalsurveyassessmentschatgpt,yan2025gptimgevalcomprehensivebenchmarkdiagnosing}.
The evaluation is conducted based on the following two criteria:
\begin{enumerate}
    \item \textbf{Character Similarity}: Measure how closely the generated character resembles the reference character.
    \item \textbf{Composition \& Image quality}: Measures how technical and artistic standards consider composition coherence, deformities, and fidelity in text, lighting, color, and clarity.
\end{enumerate}

We score each criterion on a 10-point Likert scale, assigning 1 to failure cases (e.g., unrecognizable style or severe artifacts) and 10 to excellent fidelity and quality.
For every concept, GPT-4.1 is presented with a small gallery consisting of a reference image and a generated sample, and it assigns independent scores for each pair of images. We then average the scores over 50 pairs per method.
To prevent bias, we shuffle the images and conceal the method labels from the evaluator in a pairwise blind evaluation setting. This protocol follows the official OpenAI vision evaluation guidelines and is consistent with the best practices in the literature.

\subsection{Experiments Results}
As shown in \cref{tab:fid_dino_table} and \cref{tab:ccip_insight_gpt_table}, our method consistently improves personalization quality across multiple diffusion backbones.
On SD 1.5 and SDXL, both FAD and sFAD outperform Normal Dropout in most settings, and sFAD typically yields the strongest overall gains.
Importantly, the same trend is observed on additional backbones (FLUX and Qwen-Image), indicating that Frequency-Aware Dropout generalizes beyond the Stable Diffusion family and remains effective under different architectures and training recipes.

Additionally, we conduct GPT-4.1 Evaluation based on two criteria: Character Similarity and Composition \& Image Quality.
Separating the evaluation into these two dimensions allows us to independently assess a model's ability to preserve the subject's distinctive features---such as facial structure, hairstyle, or eye color---while also evaluating the technical and artistic quality of the generated image. We also report GPT-4.1 evaluations for FLUX and Qwen-Image in \cref{tab:ccip_insight_gpt_table}, which further supports the improved prompt controllability and perceived quality under FAD/sFAD.

These results clearly indicate that the model trained with FAD consistently achieves better performance across multiple evaluation settings.
However, GPT-4.1--based MLLM evaluation shows less improvements compared to other metrics, likely due to its sensitivity to prompt phrasing and the variability in interpreting character-specific features.
A more extensive study with larger image sets and refined evaluation prompts is left as future work to stabilize and better validate MLLM-based assessments.

\subsection{Anchoring to Pretrained Tokens}
Beyond newly introduced trigger tokens, we further study whether our dropout strategy
also improves controllability when using \emph{pretrained} text tokens as anchors.
Here, an \emph{anchor token} refers to a semantically meaningful phrase that is already well-represented
in the text encoder (e.g., \texttt{japanese man}, \texttt{korean man}), rather than a newly initialized trigger token.
In this setting, we fine-tune LoRA using the same image set while replacing the trigger token with an anchor phrase
and evaluate whether the model can still bind the target identity to the anchor at inference time.

Interestingly, we observe that FAD/sFAD consistently improves anchor-based control as well:
the generated samples better preserve the target identity and exhibit more stable prompt behavior compared to the baseline.
This suggests that our method does not merely benefit \emph{new} tokens, but more generally mitigates co-occurrence interference
and strengthens token-level binding, even when the conditioning token is already pretrained.

\begin{figure}[t]
    \centering
    \setlength{\tabcolsep}{2pt}
    \begin{tabular}{cc}
        \includegraphics[width=0.48\linewidth]{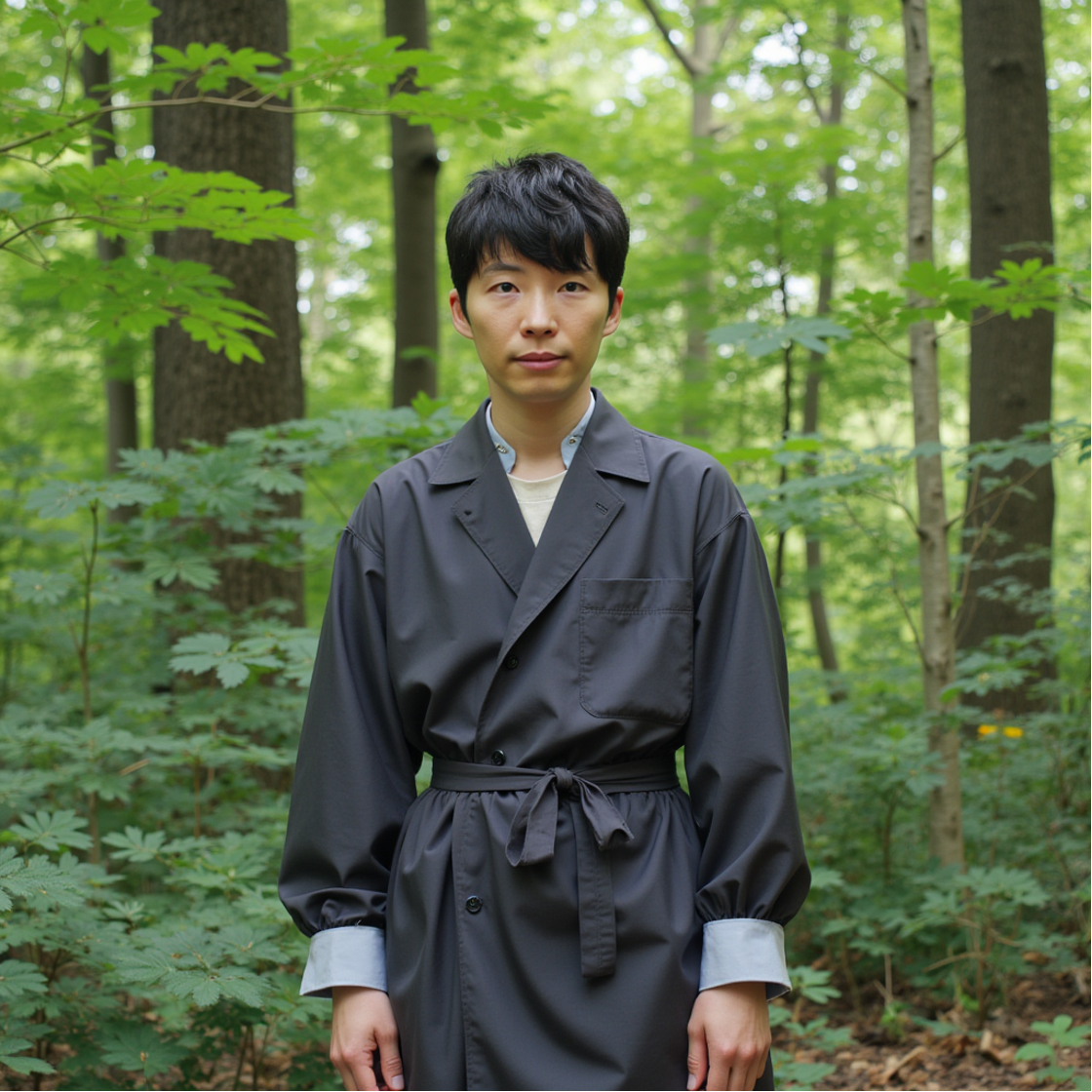} &
        \includegraphics[width=0.48\linewidth]{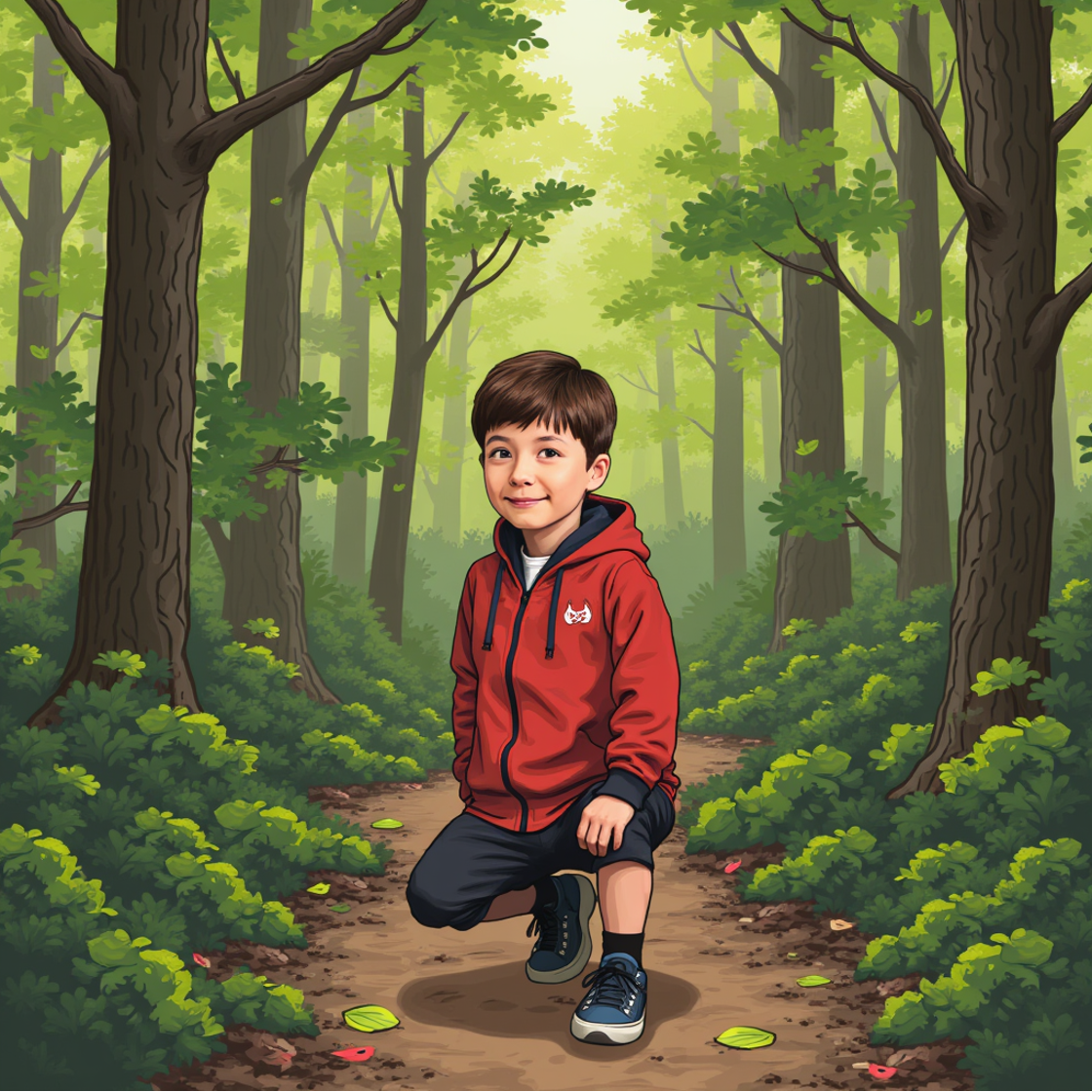} \\
        \scriptsize (a) FAD + \texttt{japanese man} &
        \scriptsize (b) FAD + \texttt{except japanese man} \\[0.3em]
        \includegraphics[width=0.48\linewidth]{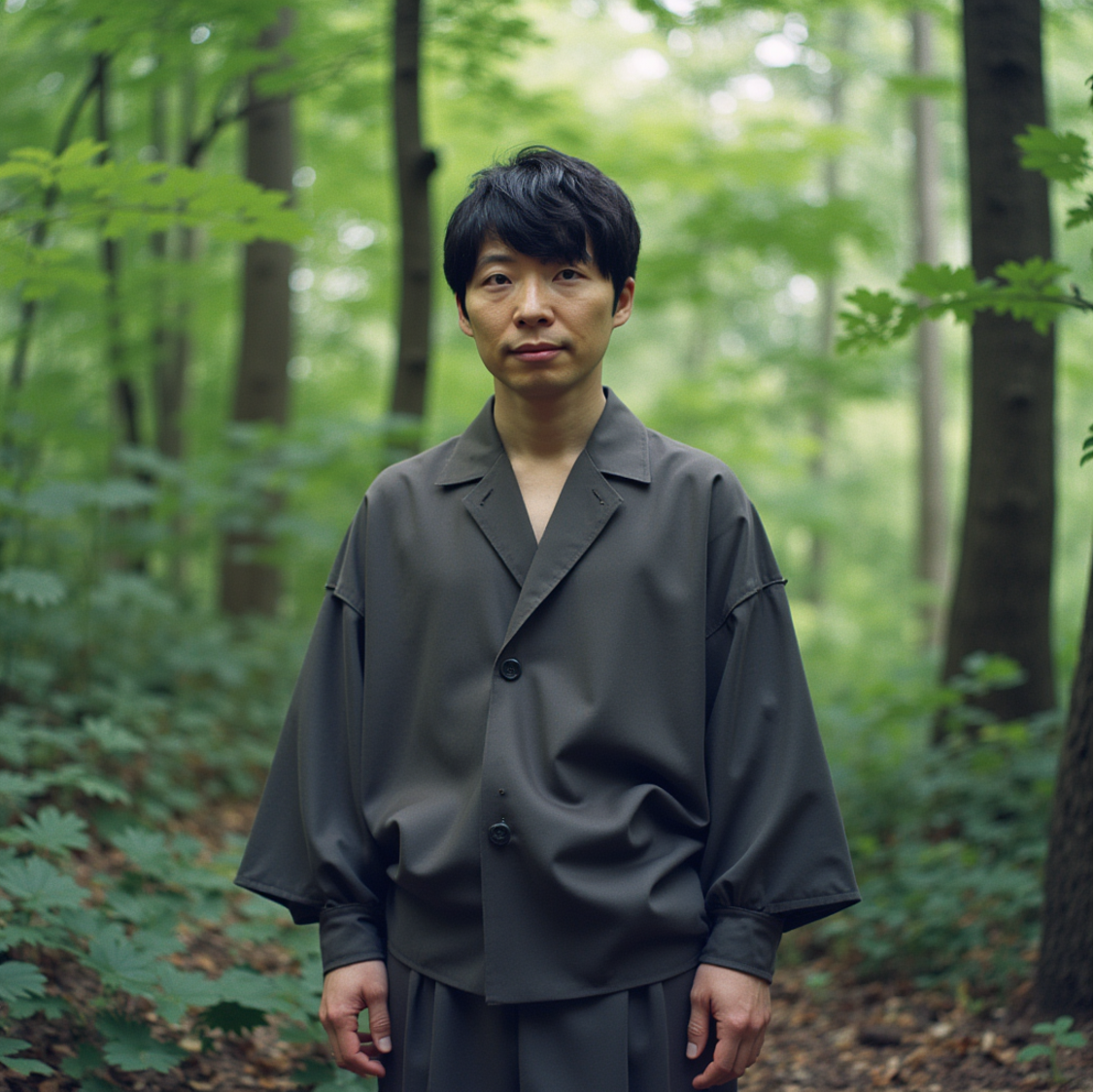} &
        \includegraphics[width=0.48\linewidth]{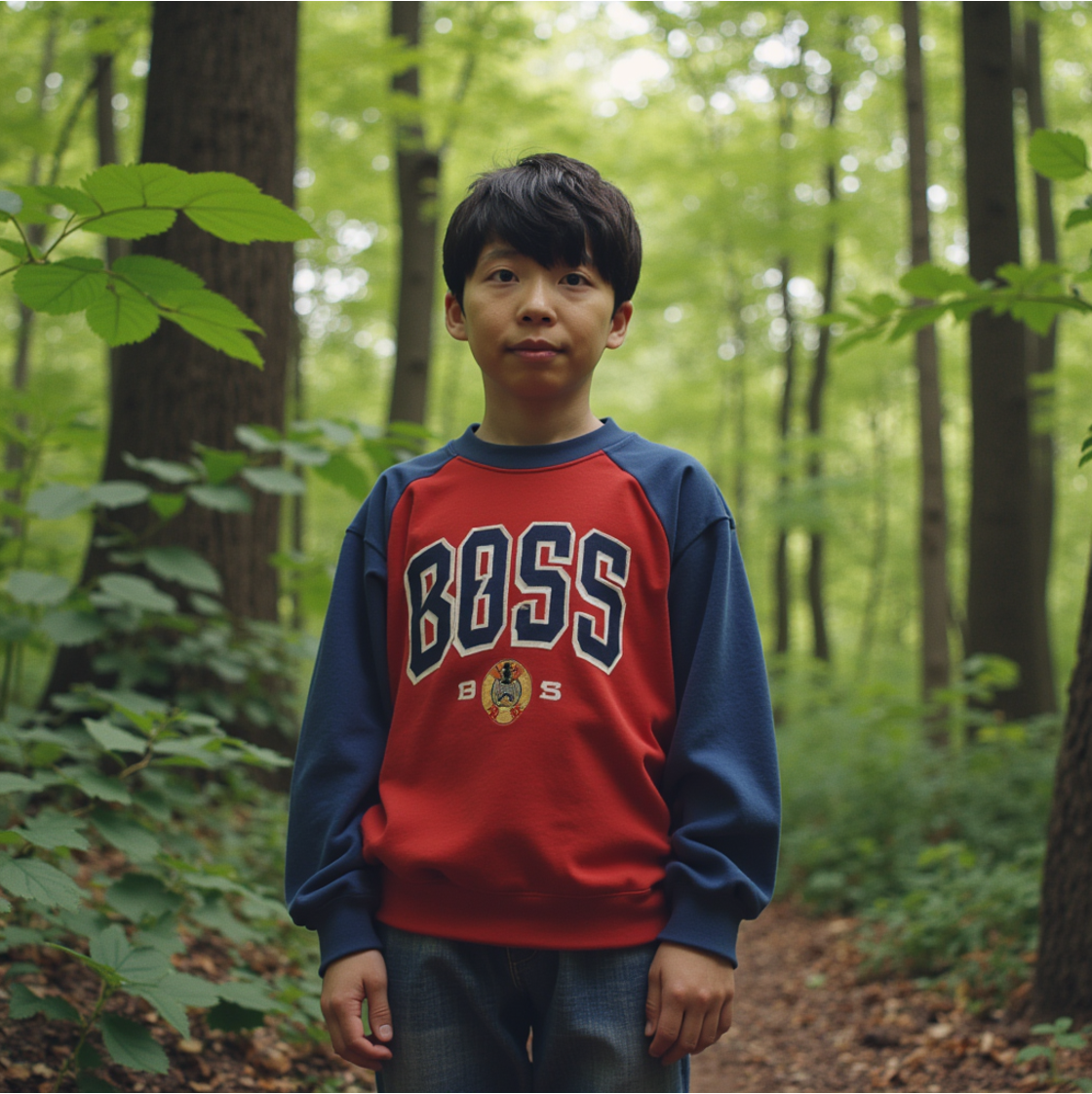} \\
        \scriptsize (c) Normal + \texttt{japanese man} &
        \scriptsize (d) Normal + \texttt{except japanese man} \\
    \end{tabular}
    \caption{
    Anchoring results for \texttt{hsng} trained with anchor token \texttt{japanese man}.
    When the anchor is removed (``except''), FAD/sFAD degenerates to a generic face (often child-like), indicating tighter identity binding.
    The Normal baseline still shows identity leakage under the \texttt{except} prompt, suggesting weaker disentanglement.
    }
    \label{fig:anchoring_results}
\end{figure}

\cref{tab:anchoring_dino_only} reports the quantitative results of the anchoring experiment across DiT-based backbones (FLUX.1-dev and Qwen-Image).
For each dataset, we compare the Normal Dropout baseline and our proposed FAD under the corresponding anchoring token.
Across all four real-person datasets, FAD consistently improves the DINO similarity score, indicating better preservation of the target identity when conditioned on a semantically known token.
Notably, these gains are observed on both latent diffusion (FLUX.1-dev) and natural language--driven DiT (Qwen-Image), demonstrating that our method generalizes across different model families.
The improvements further support the qualitative findings in \cref{fig:anchoring_results}, where FAD reduces identity leakage and enforces tighter binding between the learned concept and the explicit anchor token.

\cref{fig:anchoring_results} illustrates that \textbf{FAD/sFAD} tends to bind the identity more explicitly to the conditioning token: when the anchor is excluded, generations often collapse to a generic face (frequently child-like), implying reduced identity leakage into surrounding context.
In contrast, the \textbf{Random/Normal} baseline still exhibits noticeable identity traces of \texttt{hsng} even under the \texttt{except} condition, suggesting that identity features are more diffusely distributed across the prompt context and are less disentangled.

\begin{table}[t]
\centering
\caption{DINO score comparison under anchoring token training for DiT-based backbones. Best values per dataset are in bold.}
\resizebox{\columnwidth}{!}{%
\begin{tabular}{ll|cc}
\toprule
& & \textbf{FLUX.1-dev} & \textbf{Qwen-Image} \\
\textbf{Dataset} & \textbf{Method} & \textbf{DINO $\uparrow$} & \textbf{DINO $\uparrow$} \\
\midrule
hsng     & Normal                  & 0.5628 & 0.7347 \\
         & FAD (anchor: japanese man)  & \textbf{0.5725} & \textbf{0.7840} \\
\midrule
mbst     & Normal                  & 0.5316 & 0.6784 \\
         & FAD (anchor: american man) & \textbf{0.5453} & \textbf{0.6840} \\
\midrule
faker    & Normal                  & 0.5268 & 0.6373 \\
         & FAD (anchor: korean man) & \textbf{0.5361} & \textbf{0.6529} \\
\midrule
reeves   & Normal                  & 0.5082 & 0.6543 \\
         & FAD (anchor: european man) & \textbf{0.5153} & \textbf{0.6576} \\
\bottomrule
\end{tabular}%
}
\label{tab:anchoring_dino_only}
\end{table}

\section{Analysis}

\begin{figure}[t]
    \centering
    \begin{adjustbox}{max width=\linewidth}
    \begin{tabular}{ccc}
        \toprule
        \multicolumn{3}{c}{\textbf{pikachu}} \\
        \includegraphics[width=0.32\linewidth]{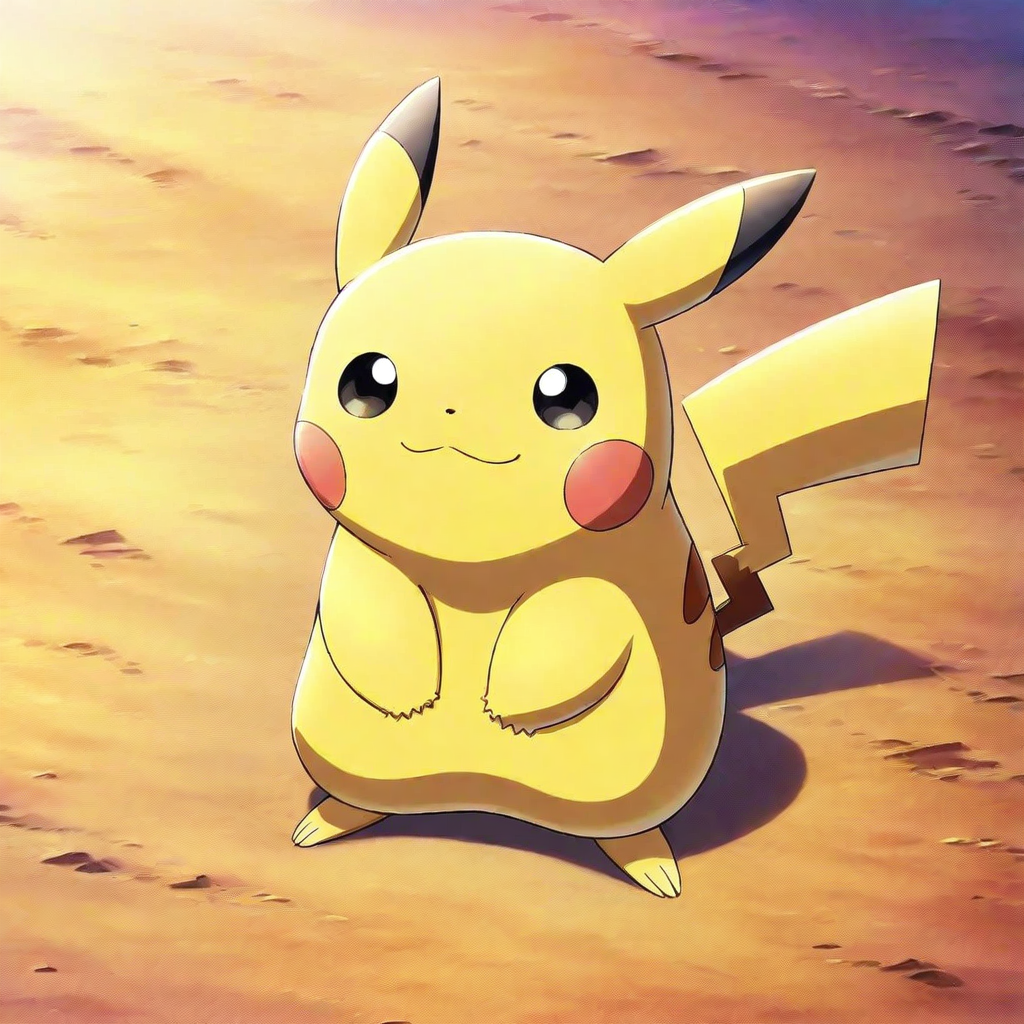} &
        \includegraphics[width=0.32\linewidth]{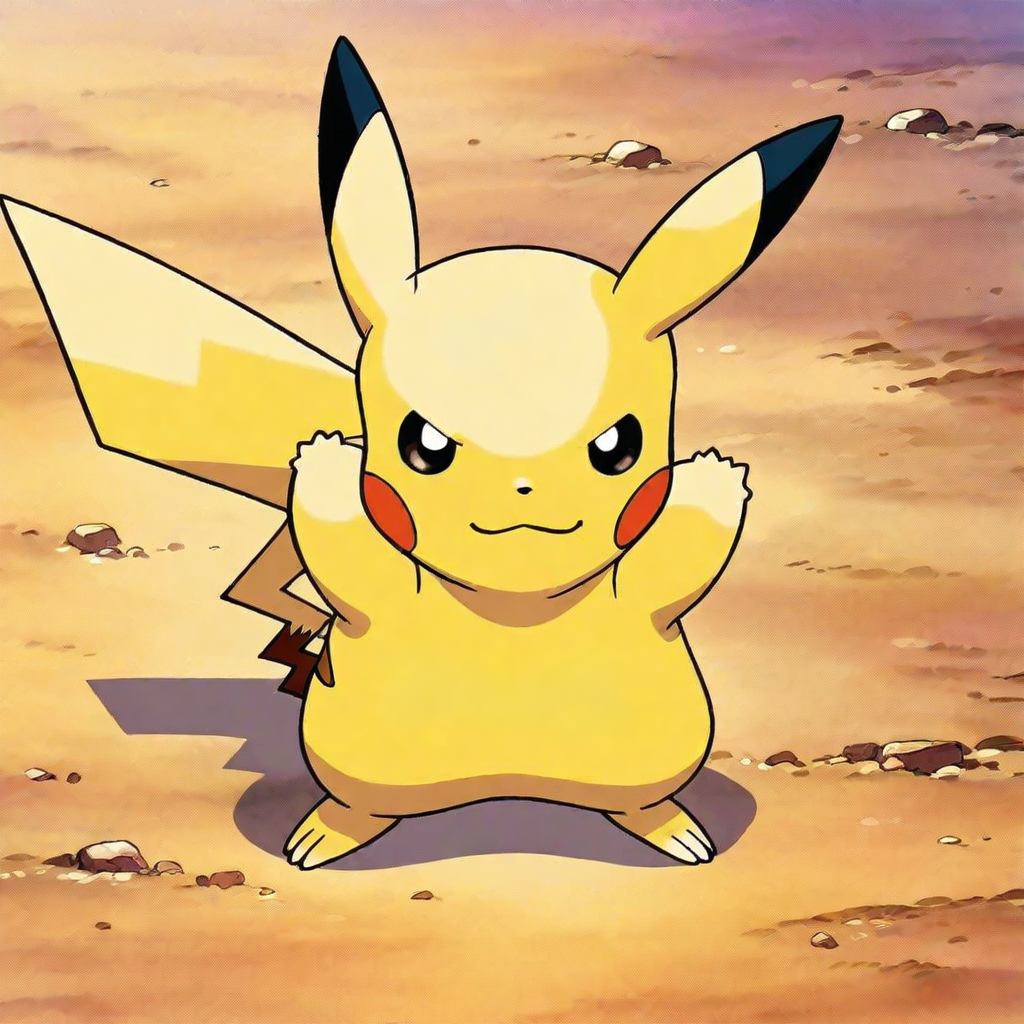} &
        \includegraphics[width=0.32\linewidth]{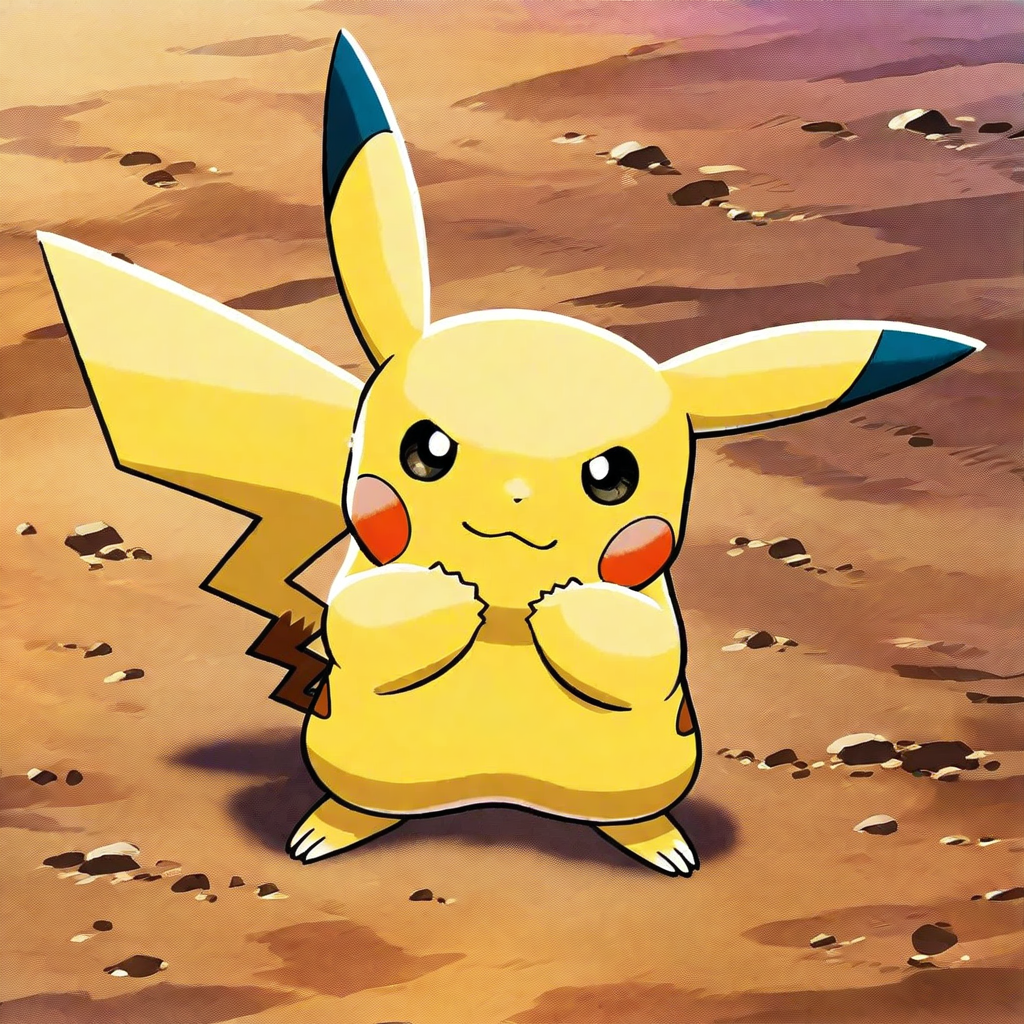} \\
        \includegraphics[width=0.32\linewidth]{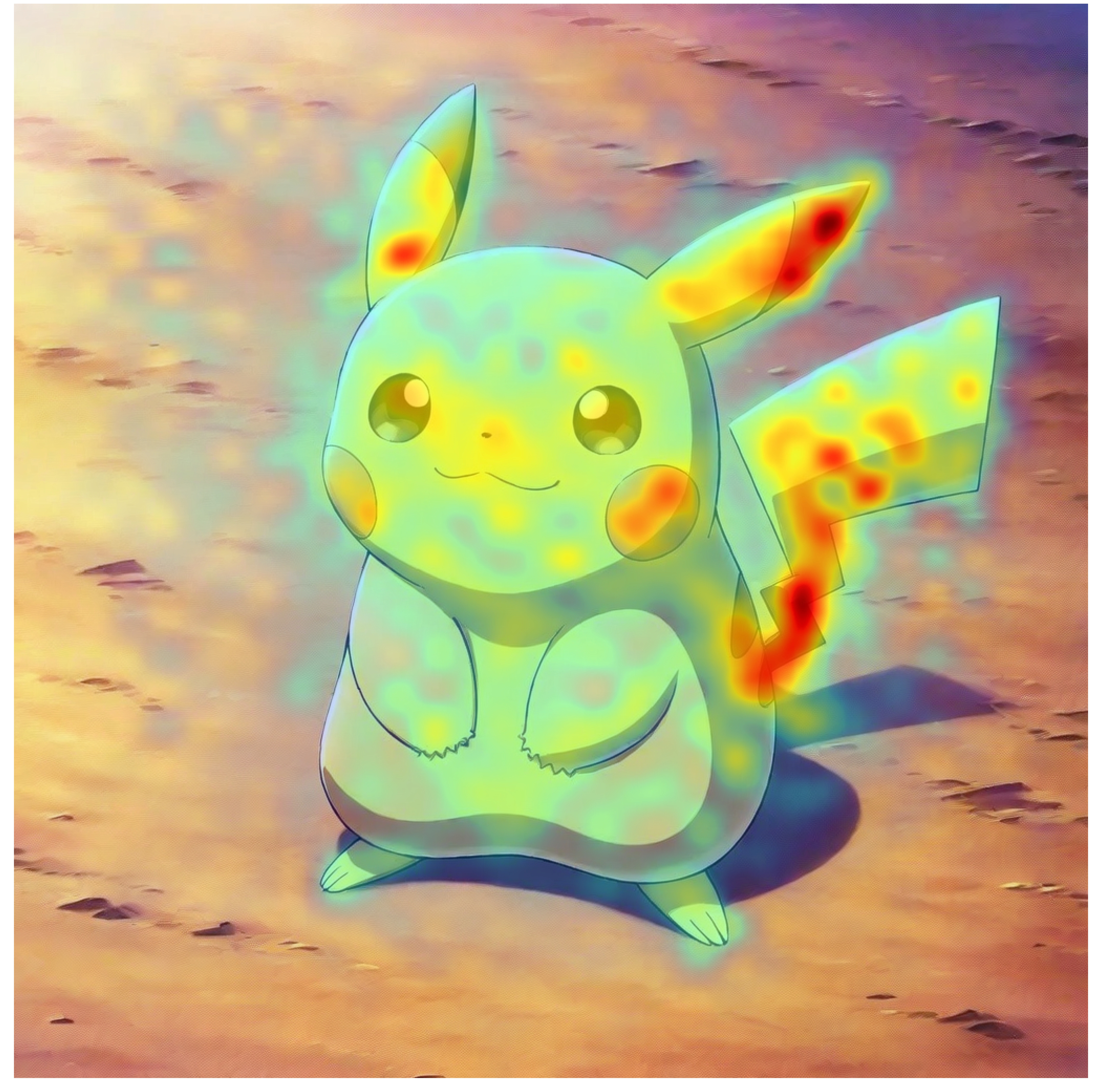} &
        \includegraphics[width=0.32\linewidth]{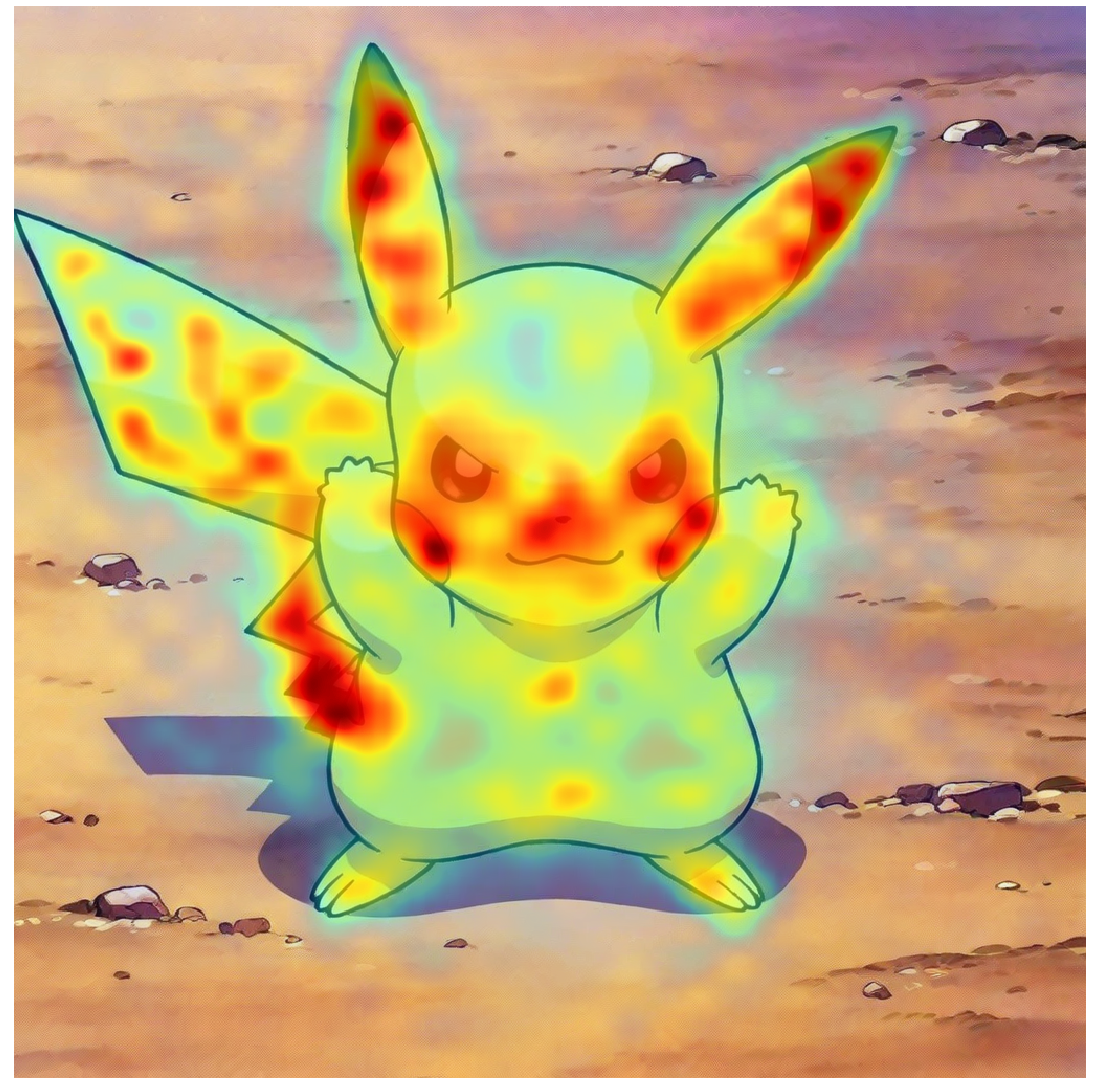} &
        \includegraphics[width=0.32\linewidth]{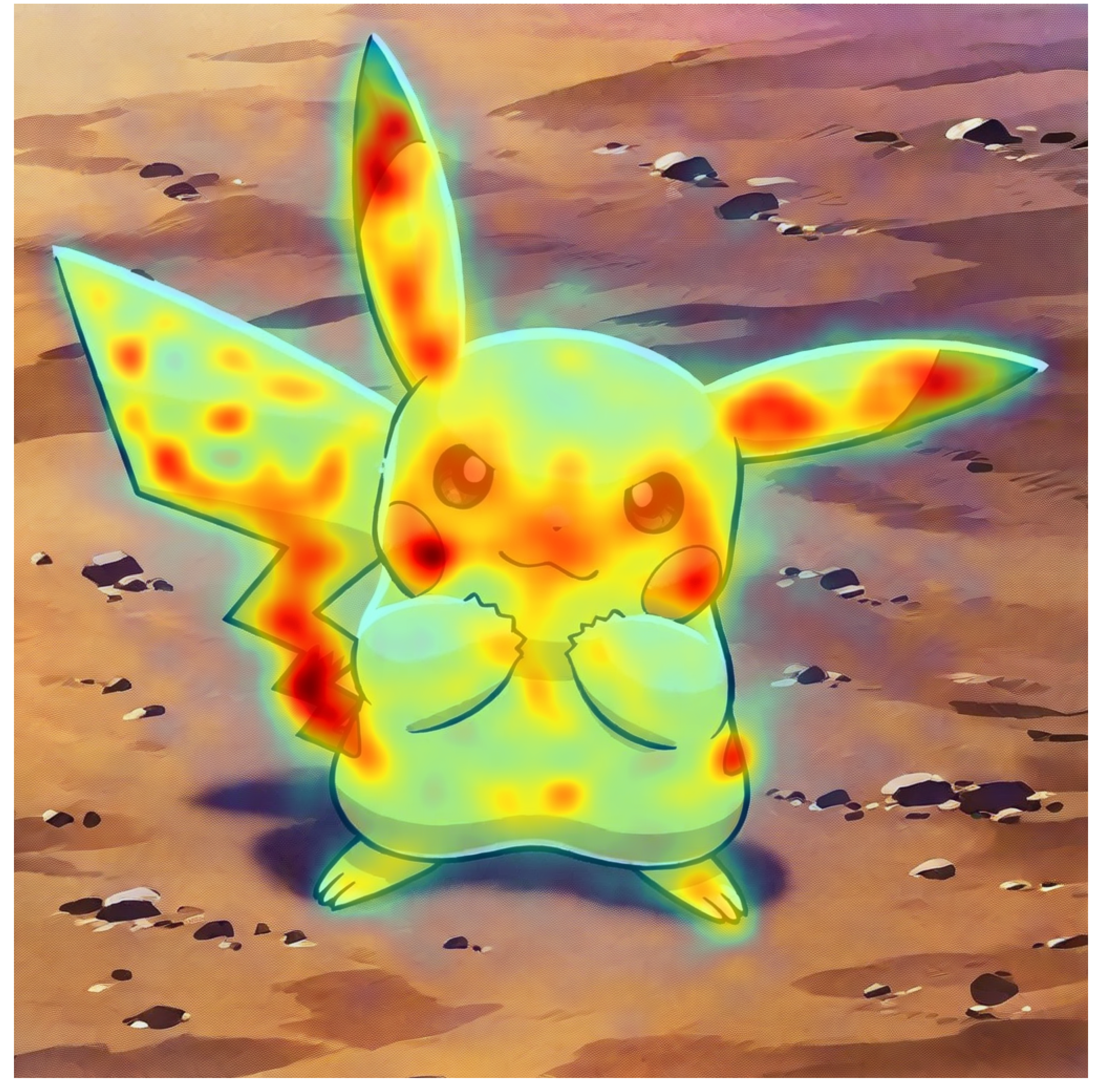} \\
        \midrule
        \multicolumn{3}{c}{\textbf{faker}} \\
        \includegraphics[width=0.32\linewidth]{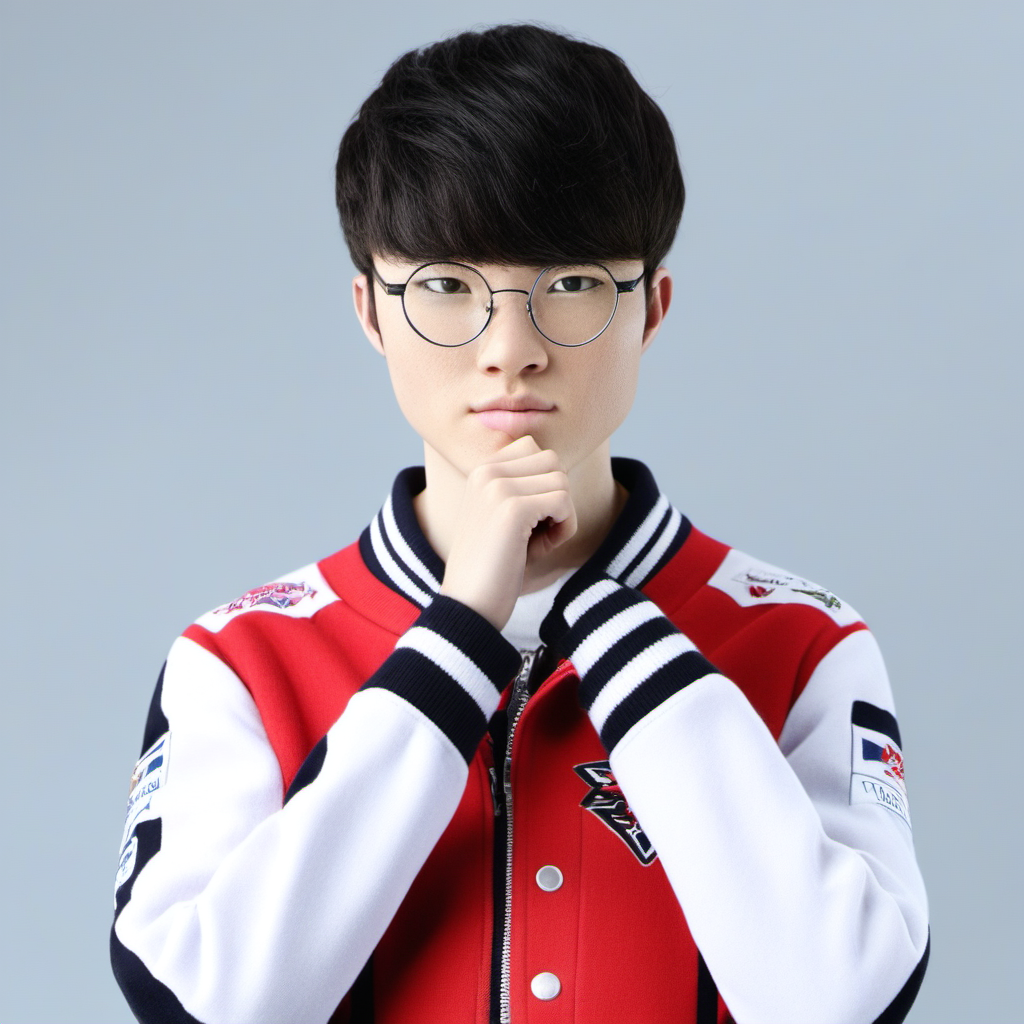} &
        \includegraphics[width=0.32\linewidth]{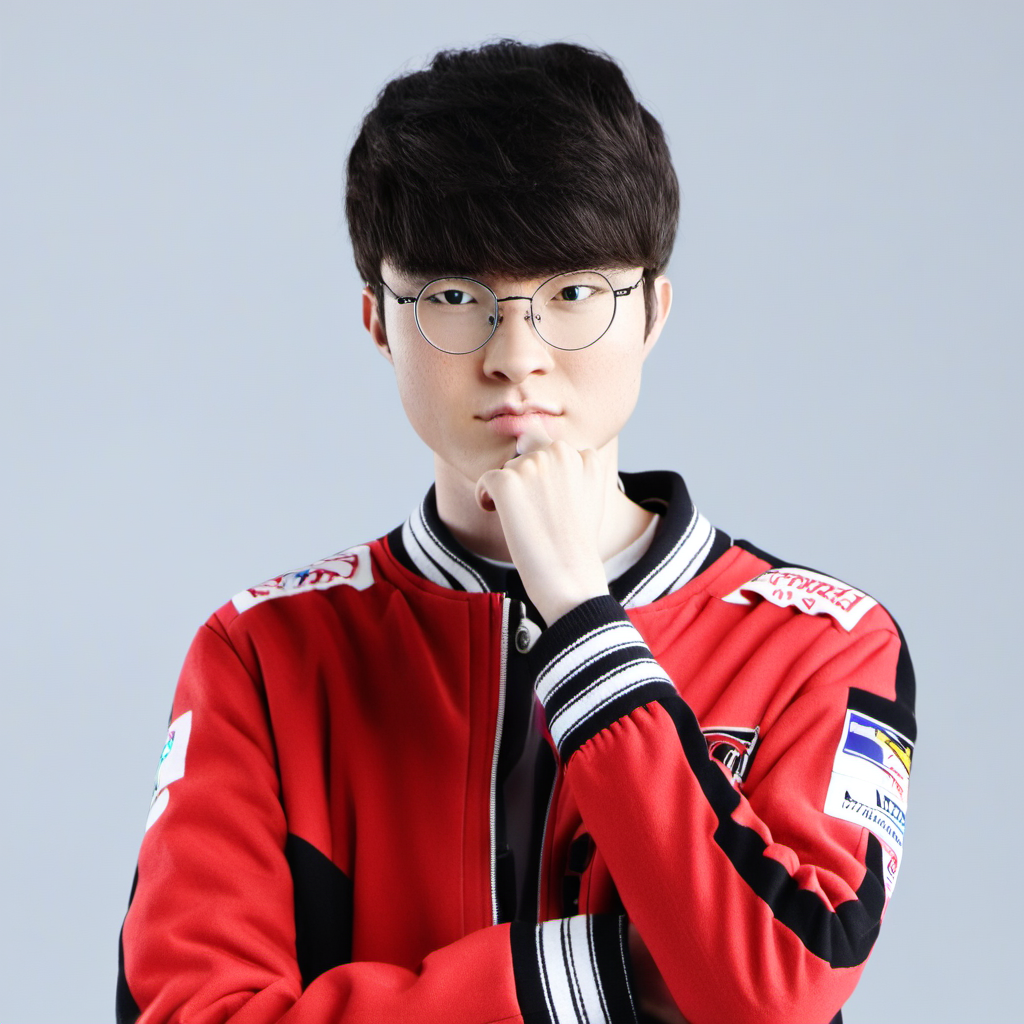} &
        \includegraphics[width=0.32\linewidth]{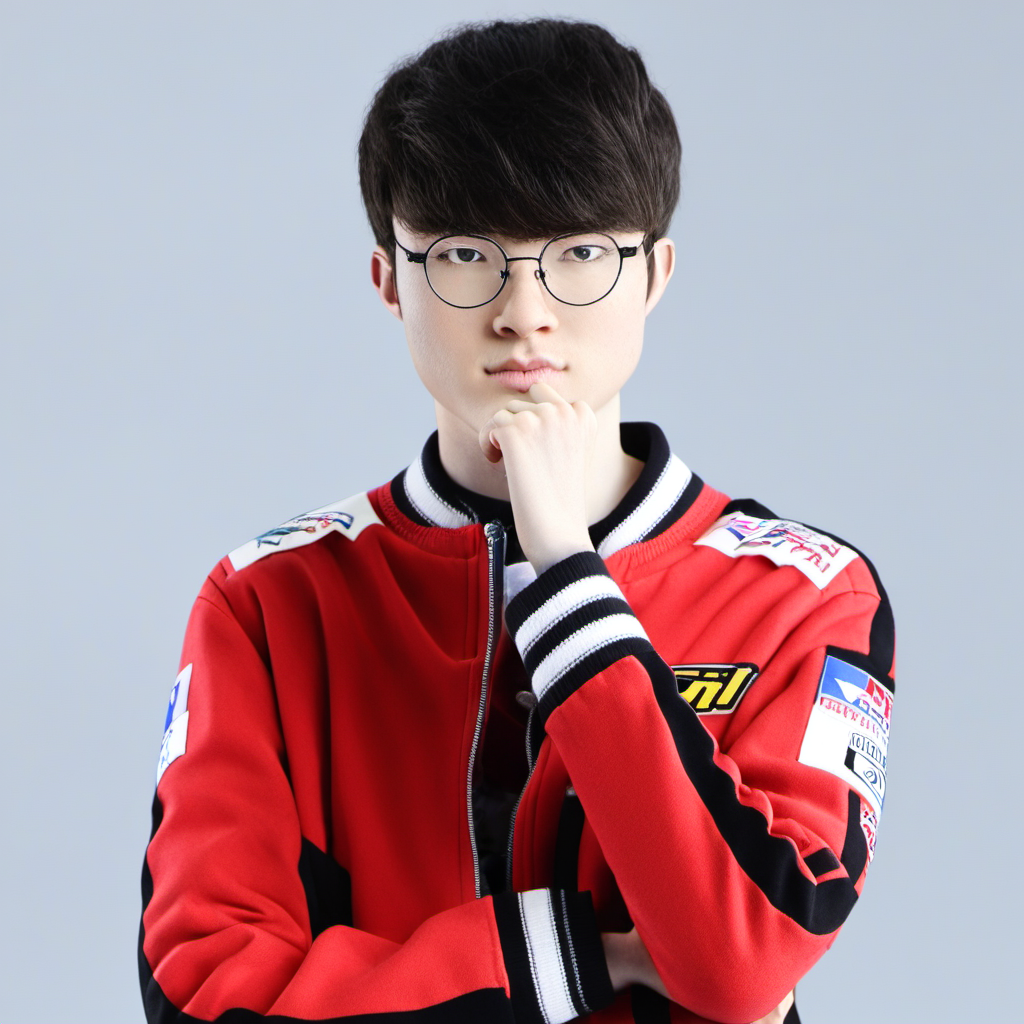} \\
        \includegraphics[width=0.32\linewidth]{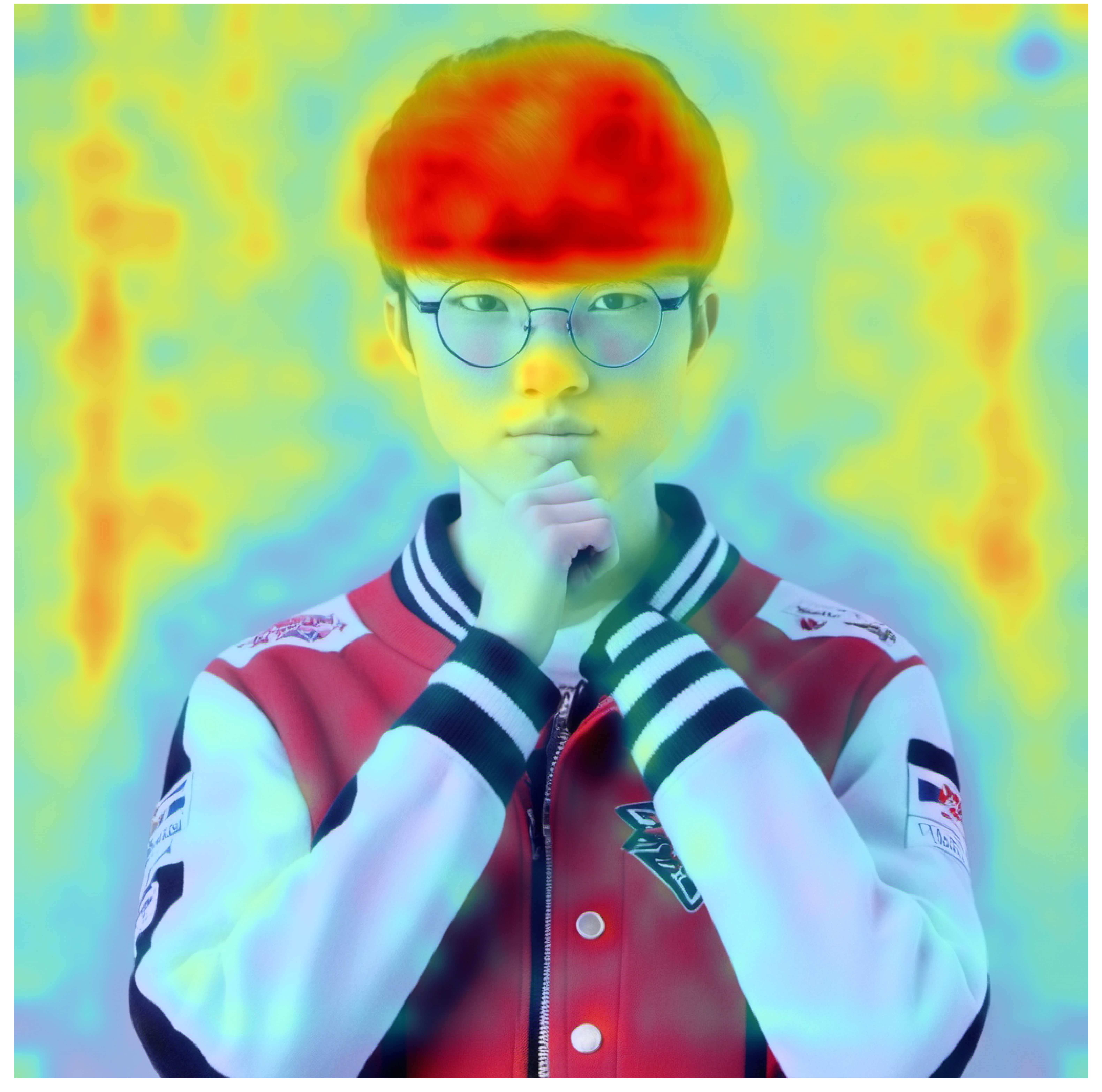} &
        \includegraphics[width=0.32\linewidth]{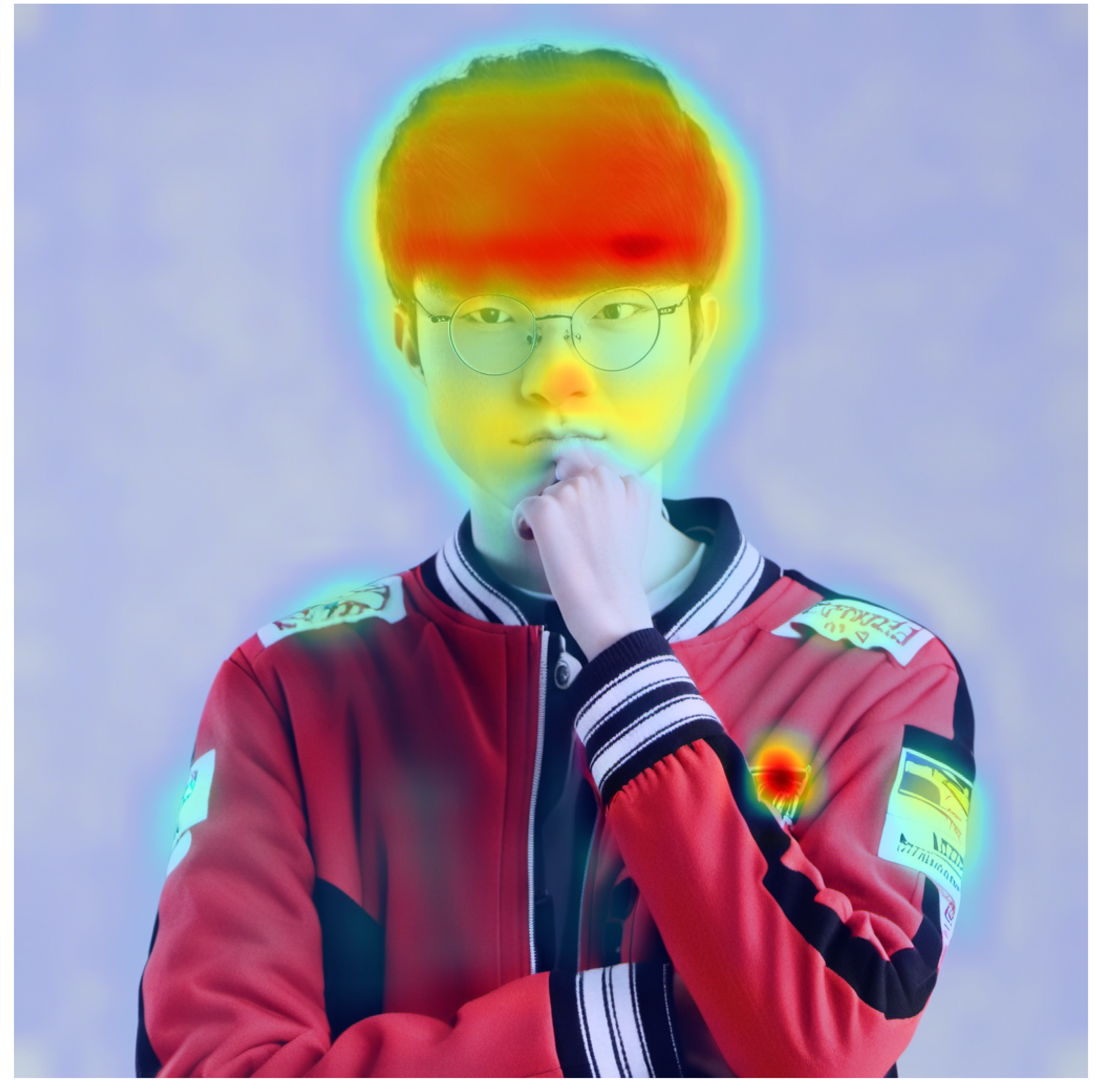} &
        \includegraphics[width=0.32\linewidth]{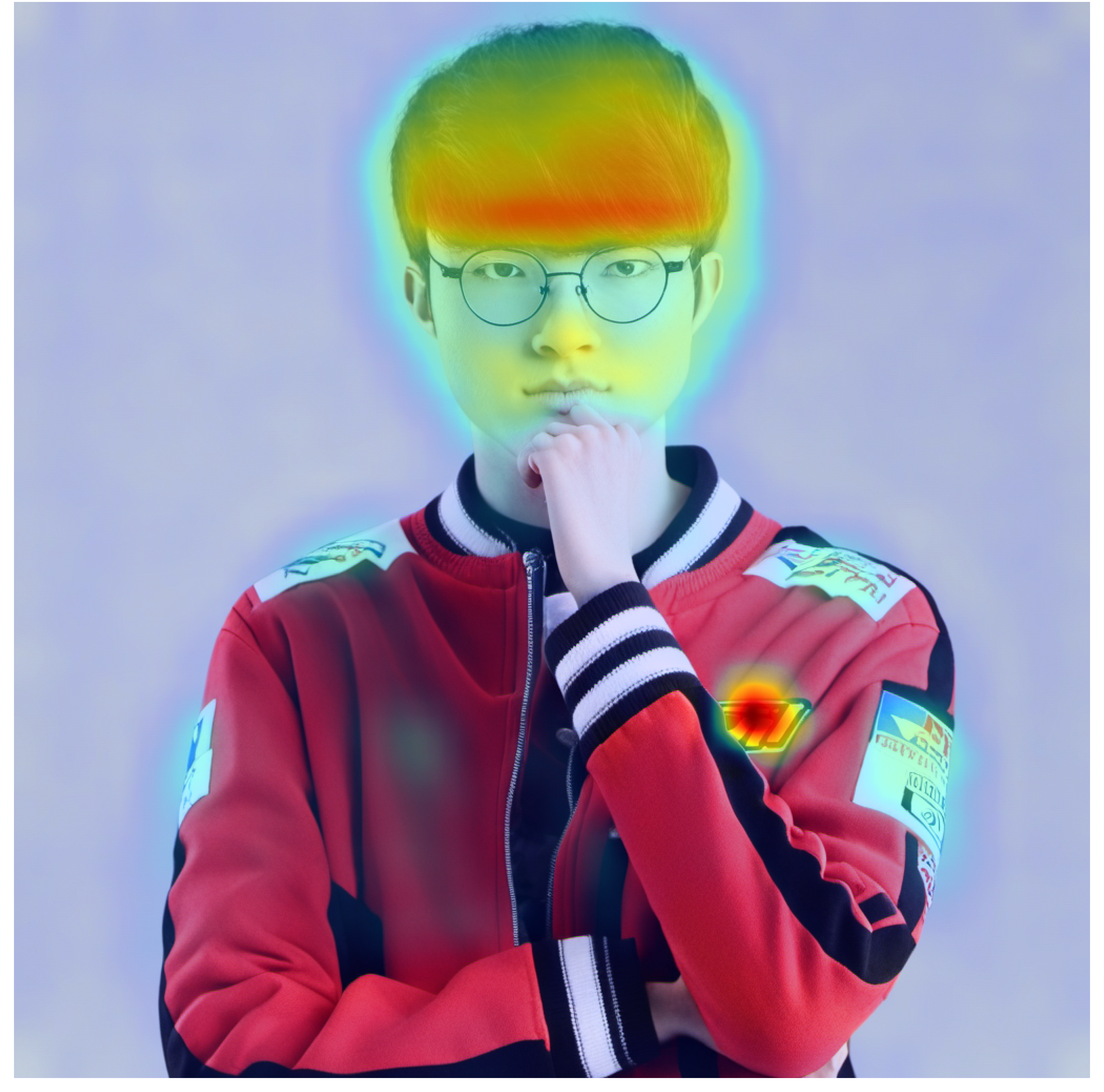} \\
        \bottomrule
    \end{tabular}
    \end{adjustbox}
    \caption{
        Attention map results using trigger token \texttt{pikachu} and \texttt{faker} under three dropout strategies: \textbf{Normal dropout (left)}, \textbf{FAD (middle)} and \textbf{sFAD (right)}. Each image is generated with the following prompts: \\
        \raggedright
        \textbf{pikachu} --- \texttt{pikachu, animal focus, black eyes, closed mouth, full body, looking at viewer, no humans, simple background, smile, solo, standing, straight-on}.
        \\
        \textbf{faker} --- \texttt{faker, 1boy, asian, black eyes, black hair, glasses, grey background, hand on own chin, hand up, looking at viewer, male focus, photorealistic, red jacket, round eyewear, short hair, simple background, solo, upper body}.
        }
    \label{fig:three_attention_maps}
\end{figure}

Employing the Diffusion Attentive Attribution Maps (DAAM)~\cite{DAAM}, we calculate cross-attention maps for the trigger token to characterize the behavior of each LoRA model.
\cref{fig:three_attention_maps} illustrates how the model's attention is distributed in response to the trigger token.
Compared to FAD and sFAD, Normal Dropout gives the trigger token much less focused attention.
Of the three methods, sFAD gives the most focused attention, showing that our method gives the best control over the trigger token.
FAD produces strong heatmaps on the character's main features, Normal Dropout shows much weaker activations.
These results clearly show that FAD disentangles the trigger token from other tokens.
As a result, FAD greatly improves controllability of trigger token in personalized diffusion models and helps the model accurately and reliably generate the desired character features.

\section{Conclusion}
We present Frequency-Aware Dropout (FAD), a lightweight and efficient method that significantly improves trigger token controllability.
By selectively dropping tokens that frequently co-occurrence with the new trigger token during fine-tuning, our method forces the model to encode the unique feature entirely into the trigger token itself.
Without incurring additional computational overhead or requiring architectural changes, our method yields notable performance gains over the traditional method, achieving higher feature fidelity and better identity preservation.
Attention-map analysis further reveals that FAD concentrates cross-attention on the trigger token's region, confirming that the model learns a more disentangled and independent representation of the new concept.

Moreover, the step-wise dropout scheduling (sFAD), which gradually increases the dropout rate during training, provides additional benefits, achieving stable convergence while further strengthening the trigger token's influence.

In future work, we plan to extend FAD to multi-concept personalization and text-to-video diffusion models, where token interference is even more pronounced, and to investigate scalable strategies for large-scale deployment.

\section{Limitations}
One limitation of our method is that its performance is highly dependent on the composition of the training data, including the amount of data, the diversity of paired tags, and the overall attribute distribution. Because FAD relies on the statistical co-occurrence of the trigger token with surrounding tokens, insufficient data diversity or skewed distributions may lead to over- or under-estimated dropout probabilities, which can weaken the expressiveness of the trigger token.
Second, multi-concept or large-scale personalization introduces significant computational overhead, as the co-occurrence matrix grows with the vocabulary size.
The current implementation performs co-occurrence analysis only offline; an online variant that dynamically updates dropout probabilities would be required for real-time or streaming scenarios.

\section*{Acknowledgements}
This work was partly supported by an IITP grant funded by the Korean Government (MSIT) (No.~RS-2020-II201361, Artificial Intelligence Graduate School Program (Yonsei University)) and the TIPS (Tech Incubator Program for Startup) grant funded by the Korea Government (MSS) (No.~RS-2023-00303145).

{
    \small
    \bibliographystyle{ieeenat_fullname}
    \bibliography{main}

@misc{c:22,
      title={Attention Is All You Need}, 
      author={Ashish Vaswani and Noam Shazeer and Niki Parmar and Jakob Uszkoreit and Llion Jones and Aidan N. Gomez and Lukasz Kaiser and Illia Polosukhin},
      year={2017},
      eprint={1706.03762},
      archivePrefix={arXiv},
      primaryClass={cs.CL}
}

@misc{podell2023sdxlimprovinglatentdiffusion,
      title={SDXL: Improving Latent Diffusion Models for High-Resolution Image Synthesis}, 
      author={Dustin Podell and Zion English and Kyle Lacey and Andreas Blattmann and Tim Dockhorn and Jonas Müller and Joe Penna and Robin Rombach},
      year={2023},
      eprint={2307.01952},
      archivePrefix={arXiv},
      primaryClass={cs.CV},
      url={https://arxiv.org/abs/2307.01952}, 
}

@misc{rombach2022highresolutionimagesynthesislatent,
      title={High-Resolution Image Synthesis with Latent Diffusion Models}, 
      author={Robin Rombach and Andreas Blattmann and Dominik Lorenz and Patrick Esser and Björn Ommer},
      year={2022},
      eprint={2112.10752},
      archivePrefix={arXiv},
      primaryClass={cs.CV},
      url={https://arxiv.org/abs/2112.10752}, 
}

@misc{hu2021loralowrankadaptationlarge,
      title={LoRA: Low-Rank Adaptation of Large Language Models}, 
      author={Edward J. Hu and Yelong Shen and Phillip Wallis and Zeyuan Allen-Zhu and Yuanzhi Li and Shean Wang and Lu Wang and Weizhu Chen},
      year={2021},
      eprint={2106.09685},
      archivePrefix={arXiv},
      primaryClass={cs.CL},
      url={https://arxiv.org/abs/2106.09685}, 
}

@misc{textualinversion,
      title={An Image is Worth One Word: Personalizing Text-to-Image Generation using Textual Inversion}, 
      author={Rinon Gal and Yuval Alaluf and Yuval Atzmon and Or Patashnik and Amit H. Bermano and Gal Chechik and Daniel Cohen-Or},
      year={2022},
      eprint={2208.01618},
      archivePrefix={arXiv},
      primaryClass={cs.CV},
      url={https://arxiv.org/abs/2208.01618}, 
}

@misc{dreambooth,
      title={DreamBooth: Fine Tuning Text-to-Image Diffusion Models for Subject-Driven Generation}, 
      author={Nataniel Ruiz and Yuanzhen Li and Varun Jampani and Yael Pritch and Michael Rubinstein and Kfir Aberman},
      year={2023},
      eprint={2208.12242},
      archivePrefix={arXiv},
      primaryClass={cs.CV},
      url={https://arxiv.org/abs/2208.12242}, 
}

@inproceedings{
  lycoris,
  title={Navigating Text-To-Image Customization: From Ly{CORIS} Fine-Tuning to Model Evaluation},
  author={SHIH-YING YEH and Yu-Guan Hsieh and Zhidong Gao and Bernard B W Yang and Giyeong Oh and Yanmin Gong},
  booktitle={The Twelfth International Conference on Learning Representations},
  year={2024},
  url={https://openreview.net/forum?id=wfzXa8e783}
}

@article{liu2024dora,
  title={DoRA: Weight-Decomposed Low-Rank Adaptation},
  author={Liu, Shih-Yang and Wang, Chien-Yi and Yin, Hongxu and Molchanov, Pavlo and Wang, Yu-Chiang Frank and Cheng, Kwang-Ting and Chen, Min-Hung},
  journal={arXiv preprint arXiv:2402.09353},
  year={2024}
}

@misc{navidet,
      title={NaviDet: Efficient Input-level Backdoor Detection on Text-to-Image Synthesis via Neuron Activation Variation}, 
      author={Shengfang Zhai and Jiajun Li and Yue Liu and Huanran Chen and Zhihua Tian and Wenjie Qu and Qingni Shen and Ruoxi Jia and Yinpeng Dong and Jiaheng Zhang},
      year={2025},
      eprint={2503.06453},
      archivePrefix={arXiv},
      primaryClass={cs.CR},
      url={https://arxiv.org/abs/2503.06453}, 
}

@misc{soboleva2025tlorasingleimagediffusion,
      title={T-LoRA: Single Image Diffusion Model Customization Without Overfitting}, 
      author={Vera Soboleva and Aibek Alanov and Andrey Kuznetsov and Konstantin Sobolev},
      year={2025},
      eprint={2507.05964},
      archivePrefix={arXiv},
      primaryClass={cs.CV},
      url={https://arxiv.org/abs/2507.05964}, 
}

@misc{loradropout,
      title={LoRA Dropout as a Sparsity Regularizer for Overfitting Control}, 
      author={Yang Lin and Xinyu Ma and Xu Chu and Yujie Jin and Zhibang Yang and Yasha Wang and Hong Mei},
      year={2024},
      eprint={2404.09610},
      archivePrefix={arXiv},
      primaryClass={cs.LG},
      url={https://arxiv.org/abs/2404.09610}, 
}

@misc{mao2024gptevalsurveyassessmentschatgpt,
      title={GPTEval: A Survey on Assessments of ChatGPT and GPT-4}, 
      author={Rui Mao and Guanyi Chen and Xulang Zhang and Frank Guerin and Erik Cambria},
      year={2024},
      eprint={2308.12488},
      archivePrefix={arXiv},
      primaryClass={cs.AI},
      url={https://arxiv.org/abs/2308.12488}, 
}

@misc{GPT4,
      title={GPT-4 Technical Report}, 
      author={OpenAI and Josh Achiam and Steven Adler and Sandhini Agarwal and Lama Ahmad and Ilge Akkaya and Florencia Leoni Aleman and Diogo Almeida and Janko Altenschmidt and Sam Altman and Shyamal Anadkat and Red Avila and Igor Babuschkin and Suchir Balaji , et al.},   
      year={2024},
      eprint={2303.08774},
      archivePrefix={arXiv},
      primaryClass={cs.CL},
      url={https://arxiv.org/abs/2303.08774}, 
}

@misc{DINO,
      title={Emerging Properties in Self-Supervised Vision Transformers}, 
      author={Mathilde Caron and Hugo Touvron and Ishan Misra and Hervé Jégou and Julien Mairal and Piotr Bojanowski and Armand Joulin},
      year={2021},
      eprint={2104.14294},
      archivePrefix={arXiv},
      primaryClass={cs.CV},
      url={https://arxiv.org/abs/2104.14294}, 
}

@misc{heusel2018ganstrainedtimescaleupdate,
      title={GANs Trained by a Two Time-Scale Update Rule Converge to a Local Nash Equilibrium}, 
      author={Martin Heusel and Hubert Ramsauer and Thomas Unterthiner and Bernhard Nessler and Sepp Hochreiter},
      year={2018},
      eprint={1706.08500},
      archivePrefix={arXiv},
      primaryClass={cs.LG},
      url={https://arxiv.org/abs/1706.08500}, 
}

@misc{DAAM,
      title={What the DAAM: Interpreting Stable Diffusion Using Cross Attention}, 
      author={Raphael Tang and Linqing Liu and Akshat Pandey and Zhiying Jiang and Gefei Yang and Karun Kumar and Pontus Stenetorp and Jimmy Lin and Ferhan Ture},
      year={2022},
      eprint={2210.04885},
      archivePrefix={arXiv},
      primaryClass={cs.CV},
      url={https://arxiv.org/abs/2210.04885}, 
}

@misc{ronneberger2015unetconvolutionalnetworksbiomedical,
      title={U-Net: Convolutional Networks for Biomedical Image Segmentation}, 
      author={Olaf Ronneberger and Philipp Fischer and Thomas Brox},
      year={2015},
      eprint={1505.04597},
      archivePrefix={arXiv},
      primaryClass={cs.CV},
      url={https://arxiv.org/abs/1505.04597}, 
}

@misc{insightface,
      title={Masked Face Recognition Challenge: The InsightFace Track Report}, 
      author={Jiankang Deng and Jia Guo and Xiang An and Zheng Zhu and Stefanos Zafeiriou},
      year={2021},
      eprint={2108.08191},
      archivePrefix={arXiv},
      primaryClass={cs.CV},
      url={https://arxiv.org/abs/2108.08191}, 
}

@misc{ddpm,
      title={Denoising Diffusion Probabilistic Models}, 
      author={Jonathan Ho and Ajay Jain and Pieter Abbeel},
      year={2020},
      eprint={2006.11239},
      archivePrefix={arXiv},
      primaryClass={cs.LG},
      url={https://arxiv.org/abs/2006.11239}, 
}

@misc{ddim,
      title={Improved Denoising Diffusion Probabilistic Models}, 
      author={Alex Nichol and Prafulla Dhariwal},
      year={2021},
      eprint={2102.09672},
      archivePrefix={arXiv},
      primaryClass={cs.LG},
      url={https://arxiv.org/abs/2102.09672}, 
}

@misc{diffusionbeatsgan,
      title={Diffusion Models Beat GANs on Image Synthesis}, 
      author={Prafulla Dhariwal and Alex Nichol},
      year={2021},
      eprint={2105.05233},
      archivePrefix={arXiv},
      primaryClass={cs.LG},
      url={https://arxiv.org/abs/2105.05233}, 
}

@misc{xu2023parameterefficientfinetuningmethodspretrained,
      title={Parameter-Efficient Fine-Tuning Methods for Pretrained Language Models: A Critical Review and Assessment}, 
      author={Lingling Xu and Haoran Xie and Si-Zhao Joe Qin and Xiaohui Tao and Fu Lee Wang},
      year={2023},
      eprint={2312.12148},
      archivePrefix={arXiv},
      primaryClass={cs.CL},
      url={https://arxiv.org/abs/2312.12148}, 
}

@misc{ho2022classifierfreediffusionguidance,
      title={Classifier-Free Diffusion Guidance}, 
      author={Jonathan Ho and Tim Salimans},
      year={2022},
      eprint={2207.12598},
      archivePrefix={arXiv},
      primaryClass={cs.LG},
      url={https://arxiv.org/abs/2207.12598}, 
}

@misc{park2024illustriousopenadvancedillustration,
      title={Illustrious: an Open Advanced Illustration Model}, 
      author={Sang Hyun Park and Jun Young Koh and Junha Lee and Joy Song and Dongha Kim and Hoyeon Moon and Hyunju Lee and Min Song},
      year={2024},
      eprint={2409.19946},
      archivePrefix={arXiv},
      primaryClass={cs.CV},
      url={https://arxiv.org/abs/2409.19946}, 
}

@misc{labs2025flux1kontextflowmatching,
      title={FLUX.1 Kontext: Flow Matching for In-Context Image Generation and Editing in Latent Space}, 
      author={Black Forest Labs and Stephen Batifol and Andreas Blattmann and Frederic Boesel and Saksham Consul and Cyril Diagne and Tim Dockhorn and Jack English and Zion English and Patrick Esser and Sumith Kulal and Kyle Lacey and Yam Levi and Cheng Li and Dominik Lorenz and Jonas Müller and Dustin Podell and Robin Rombach and Harry Saini and Axel Sauer and Luke Smith},
      year={2025},
      eprint={2506.15742},
      archivePrefix={arXiv},
      primaryClass={cs.GR},
      url={https://arxiv.org/abs/2506.15742}, 
}

@misc{mou2023t2iadapterlearningadaptersdig,
      title={T2I-Adapter: Learning Adapters to Dig out More Controllable Ability for Text-to-Image Diffusion Models}, 
      author={Chong Mou and Xintao Wang and Liangbin Xie and Yanze Wu and Jian Zhang and Zhongang Qi and Ying Shan and Xiaohu Qie},
      year={2023},
      eprint={2302.08453},
      archivePrefix={arXiv},
      primaryClass={cs.CV},
      url={https://arxiv.org/abs/2302.08453}, 
}

@misc{gao2024luminat2xtransformingtextmodality,
      title={Lumina-T2X: Transforming Text into Any Modality, Resolution, and Duration via Flow-based Large Diffusion Transformers}, 
      author={Peng Gao and Le Zhuo and Dongyang Liu and Ruoyi Du and Xu Luo and Longtian Qiu and Yuhang Zhang and Chen Lin and Rongjie Huang and Shijie Geng and Renrui Zhang and Junlin Xi and Wenqi Shao and Zhengkai Jiang and Tianshuo Yang and Weicai Ye and He Tong and Jingwen He and Yu Qiao and Hongsheng Li},
      year={2024},
      eprint={2405.05945},
      archivePrefix={arXiv},
      primaryClass={cs.CV},
      url={https://arxiv.org/abs/2405.05945}, 
}

@misc{esser2024scalingrectifiedflowtransformers,
      title={Scaling Rectified Flow Transformers for High-Resolution Image Synthesis}, 
      author={Patrick Esser and Sumith Kulal and Andreas Blattmann and Rahim Entezari and Jonas Müller and Harry Saini and Yam Levi and Dominik Lorenz and Axel Sauer and Frederic Boesel and Dustin Podell and Tim Dockhorn and Zion English and Kyle Lacey and Alex Goodwin and Yannik Marek and Robin Rombach},
      year={2024},
      eprint={2403.03206},
      archivePrefix={arXiv},
      primaryClass={cs.CV},
      url={https://arxiv.org/abs/2403.03206}, 
}

@misc{saharia2022photorealistictexttoimagediffusionmodels,
      title={Photorealistic Text-to-Image Diffusion Models with Deep Language Understanding}, 
      author={Chitwan Saharia and William Chan and Saurabh Saxena and Lala Li and Jay Whang and Emily Denton and Seyed Kamyar Seyed Ghasemipour and Burcu Karagol Ayan and S. Sara Mahdavi and Rapha Gontijo Lopes and Tim Salimans and Jonathan Ho and David J Fleet and Mohammad Norouzi},
      year={2022},
      eprint={2205.11487},
      archivePrefix={arXiv},
      primaryClass={cs.CV},
      url={https://arxiv.org/abs/2205.11487}, 
}

@misc{ramesh2021zeroshottexttoimagegeneration,
      title={Zero-Shot Text-to-Image Generation}, 
      author={Aditya Ramesh and Mikhail Pavlov and Gabriel Goh and Scott Gray and Chelsea Voss and Alec Radford and Mark Chen and Ilya Sutskever},
      year={2021},
      eprint={2102.12092},
      archivePrefix={arXiv},
      primaryClass={cs.CV},
      url={https://arxiv.org/abs/2102.12092}, 
}

@misc{CLIP,
      title={Learning Transferable Visual Models From Natural Language Supervision}, 
      author={Alec Radford and Jong Wook Kim and Chris Hallacy and Aditya Ramesh and Gabriel Goh and Sandhini Agarwal and Girish Sastry and Amanda Askell and Pamela Mishkin and Jack Clark and Gretchen Krueger and Ilya Sutskever},
      year={2021},
      eprint={2103.00020},
      archivePrefix={arXiv},
      primaryClass={cs.CV},
      url={https://arxiv.org/abs/2103.00020}, 
}

@misc{ramesh2022hierarchicaltextconditionalimagegeneration,
      title={Hierarchical Text-Conditional Image Generation with CLIP Latents}, 
      author={Aditya Ramesh and Prafulla Dhariwal and Alex Nichol and Casey Chu and Mark Chen},
      year={2022},
      eprint={2204.06125},
      archivePrefix={arXiv},
      primaryClass={cs.CV},
      url={https://arxiv.org/abs/2204.06125}, 
}

@misc{zhang2024personalizedlorahumancenteredtext,
      title={Personalized LoRA for Human-Centered Text Understanding}, 
      author={You Zhang and Jin Wang and Liang-Chih Yu and Dan Xu and Xuejie Zhang},
      year={2024},
      eprint={2403.06208},
      archivePrefix={arXiv},
      primaryClass={cs.CL},
      url={https://arxiv.org/abs/2403.06208}, 
}

@misc{sohn2023styledroptexttoimagegenerationstyle,
      title={StyleDrop: Text-to-Image Generation in Any Style}, 
      author={Kihyuk Sohn and Nataniel Ruiz and Kimin Lee and Daniel Castro Chin and Irina Blok and Huiwen Chang and Jarred Barber and Lu Jiang and Glenn Entis and Yuanzhen Li and Yuan Hao and Irfan Essa and Michael Rubinstein and Dilip Krishnan},
      year={2023},
      eprint={2306.00983},
      archivePrefix={arXiv},
      primaryClass={cs.CV},
      url={https://arxiv.org/abs/2306.00983}, 
}

@misc{brooks2023instructpix2pixlearningfollowimage,
      title={InstructPix2Pix: Learning to Follow Image Editing Instructions}, 
      author={Tim Brooks and Aleksander Holynski and Alexei A. Efros},
      year={2023},
      eprint={2211.09800},
      archivePrefix={arXiv},
      primaryClass={cs.CV},
      url={https://arxiv.org/abs/2211.09800}, 
}

@misc{dong2025dreamartistcontrollableoneshottexttoimage,
      title={DreamArtist++: Controllable One-Shot Text-to-Image Generation via Positive-Negative Adapter}, 
      author={Ziyi Dong and Pengxu Wei and Liang Lin},
      year={2025},
      eprint={2211.11337},
      archivePrefix={arXiv},
      primaryClass={cs.CV},
      url={https://arxiv.org/abs/2211.11337}, 
}

@misc{houlsby2019parameterefficienttransferlearningnlp,
      title={Parameter-Efficient Transfer Learning for NLP}, 
      author={Neil Houlsby and Andrei Giurgiu and Stanislaw Jastrzebski and Bruna Morrone and Quentin de Laroussilhe and Andrea Gesmundo and Mona Attariyan and Sylvain Gelly},
      year={2019},
      eprint={1902.00751},
      archivePrefix={arXiv},
      primaryClass={cs.LG},
      url={https://arxiv.org/abs/1902.00751}, 
}

@misc{li2021prefixtuningoptimizingcontinuousprompts,
      title={Prefix-Tuning: Optimizing Continuous Prompts for Generation}, 
      author={Xiang Lisa Li and Percy Liang},
      year={2021},
      eprint={2101.00190},
      archivePrefix={arXiv},
      primaryClass={cs.CL},
      url={https://arxiv.org/abs/2101.00190}, 
}

@misc{jiang2024mc2multiconceptguidancecustomized,
      title={MC$^2$: Multi-concept Guidance for Customized Multi-concept Generation}, 
      author={Jiaxiu Jiang and Yabo Zhang and Kailai Feng and Xiaohe Wu and Wenbo Li and Renjing Pei and Fan Li and Wangmeng Zuo},
      year={2024},
      eprint={2404.05268},
      archivePrefix={arXiv},
      primaryClass={cs.CV},
      url={https://arxiv.org/abs/2404.05268}, 
}

@misc{zaken2022bitfitsimpleparameterefficientfinetuning,
      title={BitFit: Simple Parameter-efficient Fine-tuning for Transformer-based Masked Language-models}, 
      author={Elad Ben Zaken and Shauli Ravfogel and Yoav Goldberg},
      year={2022},
      eprint={2106.10199},
      archivePrefix={arXiv},
      primaryClass={cs.LG},
      url={https://arxiv.org/abs/2106.10199}, 
}

@misc{HyperDreamBooth,
      title={HyperDreamBooth: HyperNetworks for Fast Personalization of Text-to-Image Models}, 
      author={Nataniel Ruiz and Yuanzhen Li and Varun Jampani and Wei Wei and Tingbo Hou and Yael Pritch and Neal Wadhwa and Michael Rubinstein and Kfir Aberman},
      year={2024},
      eprint={2307.06949},
      archivePrefix={arXiv},
      primaryClass={cs.CV},
      url={https://arxiv.org/abs/2307.06949}, 
}

@misc{StyleGANT,
      title={StyleGAN-T: Unlocking the Power of GANs for Fast Large-Scale Text-to-Image Synthesis}, 
      author={Axel Sauer and Tero Karras and Samuli Laine and Andreas Geiger and Timo Aila},
      year={2023},
      eprint={2301.09515},
      archivePrefix={arXiv},
      primaryClass={cs.LG},
      url={https://arxiv.org/abs/2301.09515}, 
}

@misc{shen2024promptstealingattackstexttoimage,
      title={Prompt Stealing Attacks Against Text-to-Image Generation Models}, 
      author={Xinyue Shen and Yiting Qu and Michael Backes and Yang Zhang},
      year={2024},
      eprint={2302.09923},
      archivePrefix={arXiv},
      primaryClass={cs.CR},
      url={https://arxiv.org/abs/2302.09923}, 
}

@misc{witteveen2022investigatingpromptengineeringdiffusion,
      title={Investigating Prompt Engineering in Diffusion Models}, 
      author={Sam Witteveen and Martin Andrews},
      year={2022},
      eprint={2211.15462},
      archivePrefix={arXiv},
      primaryClass={cs.CV},
      url={https://arxiv.org/abs/2211.15462}, 
}

@misc{chefer2023hiddenlanguagediffusionmodels,
      title={The Hidden Language of Diffusion Models}, 
      author={Hila Chefer and Oran Lang and Mor Geva and Volodymyr Polosukhin and Assaf Shocher and Michal Irani and Inbar Mosseri and Lior Wolf},
      year={2023},
      eprint={2306.00966},
      archivePrefix={arXiv},
      primaryClass={cs.CV},
      url={https://arxiv.org/abs/2306.00966}, 
}

@misc{yu2024imageworth32tokens,
      title={An Image is Worth 32 Tokens for Reconstruction and Generation}, 
      author={Qihang Yu and Mark Weber and Xueqing Deng and Xiaohui Shen and Daniel Cremers and Liang-Chieh Chen},
      year={2024},
      eprint={2406.07550},
      archivePrefix={arXiv},
      primaryClass={cs.CV},
      url={https://arxiv.org/abs/2406.07550}, 
}

@misc{cheng2025revisitingloralensparameter,
      title={Revisiting LoRA through the Lens of Parameter Redundancy: Spectral Encoding Helps}, 
      author={Jiashun Cheng and Aochuan Chen and Nuo Chen and Ziqi Gao and Yuhan Li and Jia Li and Fugee Tsung},
      year={2025},
      eprint={2506.16787},
      archivePrefix={arXiv},
      primaryClass={cs.LG},
      url={https://arxiv.org/abs/2506.16787}, 
}

@misc{wang2024loragalowrankadaptationgradient,
      title={LoRA-GA: Low-Rank Adaptation with Gradient Approximation}, 
      author={Shaowen Wang and Linxi Yu and Jian Li},
      year={2024},
      eprint={2407.05000},
      archivePrefix={arXiv},
      primaryClass={cs.LG},
      url={https://arxiv.org/abs/2407.05000}, 
}

@misc{yan2025gptimgevalcomprehensivebenchmarkdiagnosing,
      title={GPT-ImgEval: A Comprehensive Benchmark for Diagnosing GPT4o in Image Generation}, 
      author={Zhiyuan Yan and Junyan Ye and Weijia Li and Zilong Huang and Shenghai Yuan and Xiangyang He and Kaiqing Lin and Jun He and Conghui He and Li Yuan},
      year={2025},
      eprint={2504.02782},
      archivePrefix={arXiv},
      primaryClass={cs.CV},
      url={https://arxiv.org/abs/2504.02782}, 
}

@article{Deng_2022,
   title={ArcFace: Additive Angular Margin Loss for Deep Face Recognition},
   volume={44},
   ISSN={1939-3539},
   url={http://dx.doi.org/10.1109/TPAMI.2021.3087709},
   DOI={10.1109/tpami.2021.3087709},
   number={10},
   journal={IEEE Transactions on Pattern Analysis and Machine Intelligence},
   publisher={Institute of Electrical and Electronics Engineers (IEEE)},
   author={Deng, Jiankang and Guo, Jia and Yang, Jing and Xue, Niannan and Kotsia, Irene and Zafeiriou, Stefanos},
   year={2022},
   month=oct, pages={5962–5979} }

@misc{wang2021facexzoopytorchtoolboxface,
      title={FaceX-Zoo: A PyTorch Toolbox for Face Recognition}, 
      author={Jun Wang and Yinglu Liu and Yibo Hu and Hailin Shi and Tao Mei},
      year={2021},
      eprint={2101.04407},
      archivePrefix={arXiv},
      primaryClass={cs.CV},
      url={https://arxiv.org/abs/2101.04407}, 
}

@misc{wu2024contrastivepromptsimprovedisentanglement,
      title={Contrastive Prompts Improve Disentanglement in Text-to-Image Diffusion Models}, 
      author={Chen Wu and Fernando De la Torre},
      year={2024},
      eprint={2402.13490},
      archivePrefix={arXiv},
      primaryClass={cs.CV},
      url={https://arxiv.org/abs/2402.13490}, 
}

@misc{xu2024cusconceptcustomizedvisualconcept,
      title={CusConcept: Customized Visual Concept Decomposition with Diffusion Models}, 
      author={Zhi Xu and Shaozhe Hao and Kai Han},
      year={2024},
      eprint={2410.00398},
      archivePrefix={arXiv},
      primaryClass={cs.CV},
      url={https://arxiv.org/abs/2410.00398}, 
}

@misc{park2025fairgenerationunfairdistortions,
      title={Fair Generation without Unfair Distortions: Debiasing Text-to-Image Generation with Entanglement-Free Attention}, 
      author={Jeonghoon Park and Juyoung Lee and Chaeyeon Chung and Jaeseong Lee and Jaegul Choo and Jindong Gu},
      year={2025},
      eprint={2506.13298},
      archivePrefix={arXiv},
      primaryClass={cs.CV},
      url={https://arxiv.org/abs/2506.13298}, 
}

@misc{wang2024tokencomposetexttoimagediffusiontokenlevel,
      title={TokenCompose: Text-to-Image Diffusion with Token-level Supervision}, 
      author={Zirui Wang and Zhizhou Sha and Zheng Ding and Yilin Wang and Zhuowen Tu},
      year={2024},
      eprint={2312.03626},
      archivePrefix={arXiv},
      primaryClass={cs.CV},
      url={https://arxiv.org/abs/2312.03626}, 
}

@misc{Ilharco_Open_Clip_2021,
author = {Ilharco, Gabriel and Wortsman, Mitchell and Carlini, Nicholas and Taori, Rohan and Dave, Achal and Shankar, Vaishaal and Namkoong, Hongseok and Miller, John and Hajishirzi, Hannaneh and Farhadi, Ali and Schmidt, Ludwig},
month = {7},
title = {Open Clip},
year = {2021}
}

@misc{CCIP,
  title={Contrastive Anime Character Image Pre-Training},
  author={deepghs},
  year={2024},
  howpublished={\url{https://huggingface.co/deepghs/ccip}}
}

@misc{kocmi2017explorationwordembeddinginitialization,
      title={An Exploration of Word Embedding Initialization in Deep-Learning Tasks}, 
      author={Tom Kocmi and Ondřej Bojar},
      year={2017},
      eprint={1711.09160},
      archivePrefix={arXiv},
      primaryClass={cs.CL},
      url={https://arxiv.org/abs/1711.09160}, 
}

@inproceedings{Dobler_2023,
   title={FOCUS: Effective Embedding Initialization for Monolingual Specialization of Multilingual Models},
   url={http://dx.doi.org/10.18653/v1/2023.emnlp-main.829},
   DOI={10.18653/v1/2023.emnlp-main.829},
   booktitle={Proceedings of the 2023 Conference on Empirical Methods in Natural Language Processing},
   publisher={Association for Computational Linguistics},
   author={Dobler, Konstantin and de Melo, Gerard},
   year={2023},
   pages={13440–13454} }

@misc{pang2023crossinitializationpersonalizedtexttoimage,
      title={Cross Initialization for Personalized Text-to-Image Generation}, 
      author={Lianyu Pang and Jian Yin and Haoran Xie and Qiping Wang and Qing Li and Xudong Mao},
      year={2023},
      eprint={2312.15905},
      archivePrefix={arXiv},
      primaryClass={cs.CV},
      url={https://arxiv.org/abs/2312.15905}, 
}

@misc{mi2025datasynthesisdiversestyles,
      title={Data Synthesis with Diverse Styles for Face Recognition via 3DMM-Guided Diffusion}, 
      author={Yuxi Mi and Zhizhou Zhong and Yuge Huang and Qiuyang Yuan and Xuan Zhao and Jianqing Xu and Shouhong Ding and ShaoMing Wang and Rizen Guo and Shuigeng Zhou},
      year={2025},
      eprint={2504.00430},
      archivePrefix={arXiv},
      primaryClass={cs.CV},
      url={https://arxiv.org/abs/2504.00430}, 
}

@misc{razavi2019generatingdiversehighfidelityimages,
      title={Generating Diverse High-Fidelity Images with VQ-VAE-2}, 
      author={Ali Razavi and Aaron van den Oord and Oriol Vinyals},
      year={2019},
      eprint={1906.00446},
      archivePrefix={arXiv},
      primaryClass={cs.LG},
      url={https://arxiv.org/abs/1906.00446}, 
}

@misc{meng2024imageregenerationevaluatingtexttoimage,
      title={Image Regeneration: Evaluating Text-to-Image Model via Generating Identical Image with Multimodal Large Language Models}, 
      author={Chutian Meng and Fan Ma and Jiaxu Miao and Chi Zhang and Yi Yang and Yueting Zhuang},
      year={2024},
      eprint={2411.09449},
      archivePrefix={arXiv},
      primaryClass={cs.CV},
      url={https://arxiv.org/abs/2411.09449}, 
}

@misc{song2024diffsimtamingdiffusionmodels,
      title={DiffSim: Taming Diffusion Models for Evaluating Visual Similarity}, 
      author={Yiren Song and Xiaokang Liu and Mike Zheng Shou},
      year={2024},
      eprint={2412.14580},
      archivePrefix={arXiv},
      primaryClass={cs.CV},
      url={https://arxiv.org/abs/2412.14580}, 
}

@misc{loshchilov2019decoupledweightdecayregularization,
      title={Decoupled Weight Decay Regularization}, 
      author={Ilya Loshchilov and Frank Hutter},
      year={2019},
      eprint={1711.05101},
      archivePrefix={arXiv},
      primaryClass={cs.LG},
      url={https://arxiv.org/abs/1711.05101}, 
}

@misc{ren2025unveilingmitigatingmemorizationtexttoimage,
      title={Unveiling and Mitigating Memorization in Text-to-image Diffusion Models through Cross Attention}, 
      author={Jie Ren and Yaxin Li and Shenglai Zeng and Han Xu and Lingjuan Lyu and Yue Xing and Jiliang Tang},
      year={2025},
      eprint={2403.11052},
      archivePrefix={arXiv},
      primaryClass={cs.CV},
      url={https://arxiv.org/abs/2403.11052}, 
}

@misc{zhang2024enhancingsemanticfidelitytexttoimage,
      title={Enhancing Semantic Fidelity in Text-to-Image Synthesis: Attention Regulation in Diffusion Models}, 
      author={Yang Zhang and Teoh Tze Tzun and Lim Wei Hern and Tiviatis Sim and Kenji Kawaguchi},
      year={2024},
      eprint={2403.06381},
      archivePrefix={arXiv},
      primaryClass={cs.CV},
      url={https://arxiv.org/abs/2403.06381}, 
}

@misc{zhong2024multiloracompositionimagegeneration,
      title={Multi-LoRA Composition for Image Generation}, 
      author={Ming Zhong and Yelong Shen and Shuohang Wang and Yadong Lu and Yizhu Jiao and Siru Ouyang and Donghan Yu and Jiawei Han and Weizhu Chen},
      year={2024},
      eprint={2402.16843},
      archivePrefix={arXiv},
      primaryClass={cs.CV},
      url={https://arxiv.org/abs/2402.16843}, 
}

@misc{qwenimage,
      title={Qwen-Image Technical Report}, 
      author={Chenfei Wu and Jiahao Li and Jingren Zhou and Junyang Lin and Kaiyuan Gao and Kun Yan and Sheng-ming Yin and Shuai Bai and Xiao Xu and Yilei Chen and Yuxiang Chen and Zecheng Tang and Zekai Zhang and Zhengyi Wang and An Yang and Bowen Yu and Chen Cheng and Dayiheng Liu and Deqing Li and Hang Zhang and Hao Meng and Hu Wei and Jingyuan Ni and Kai Chen and Kuan Cao and Liang Peng and Lin Qu and Minggang Wu and Peng Wang and Shuting Yu and Tingkun Wen and Wensen Feng and Xiaoxiao Xu and Yi Wang and Yichang Zhang and Yongqiang Zhu and Yujia Wu and Yuxuan Cai and Zenan Liu},
      year={2025},
      eprint={2508.02324},
      archivePrefix={arXiv},
      primaryClass={cs.CV},
      url={https://arxiv.org/abs/2508.02324}, 
}

@misc{qwen,
      title={Qwen Technical Report}, 
      author={Jinze Bai and Shuai Bai and Yunfei Chu and Zeyu Cui and Kai Dang and Xiaodong Deng and Yang Fan and Wenbin Ge and Yu Han and Fei Huang and Binyuan Hui and Luo Ji and Mei Li and Junyang Lin and Runji Lin and Dayiheng Liu and Gao Liu and Chengqiang Lu and Keming Lu and Jianxin Ma and Rui Men and Xingzhang Ren and Xuancheng Ren and Chuanqi Tan and Sinan Tan and Jianhong Tu and Peng Wang and Shijie Wang and Wei Wang and Shengguang Wu and Benfeng Xu and Jin Xu and An Yang and Hao Yang and Jian Yang and Shusheng Yang and Yang Yao and Bowen Yu and Hongyi Yuan and Zheng Yuan and Jianwei Zhang and Xingxuan Zhang and Yichang Zhang and Zhenru Zhang and Chang Zhou and Jingren Zhou and Xiaohuan Zhou and Tianhang Zhu},
      year={2023},
      eprint={2309.16609},
      archivePrefix={arXiv},
      primaryClass={cs.CL},
      url={https://arxiv.org/abs/2309.16609}, 
}

@misc{flux2024,
    author={Black Forest Labs},
    title={FLUX},
    year={2024},
    howpublished={\url{https://github.com/black-forest-labs/flux}},
}
}

\clearpage
\setcounter{page}{1}
\maketitlesupplementary

\section{Training Hyperparameters}

We summarize the training hyperparameters used in our experiments in
\cref{tab:hyperparameters_sd} for SD~1.5 and SDXL, and in
\cref{tab:hyperparameters_flux_qwen} for FLUX and Qwen-Image.
For the SD~1.5 and SDXL experiments, we train U-Net LoRA using a batch size of 8 and
64 gradient accumulation steps on a single RTX~3090 GPU.
We adopt the AdamW optimizer~\cite{loshchilov2019decoupledweightdecayregularization}
with weight decay $\lambda = 0.01$ and $\beta=(0.9, 0.999)$,
and set the learning rate to $5\times10^{-5}$.
For the DiT-based backbones (FLUX and Qwen-Image), we follow their official LoRA
training recipes.

\begin{table}[hbtp]
    \centering
    \footnotesize
    \begin{tabular}{l|c}
        \toprule
        \textbf{Hyperparameter} & \textbf{Value} \\
        \midrule
        Train batch size           & 8     \\
        Gradient accumulation steps& 64    \\
        Max train steps            & 1500  \\
        Min adaptive dropout       & 0.35  \\
        Max adaptive dropout       & 1.0   \\
        Step dropout start         & 0.1   \\
        Step dropout end           & 0.8   \\
        Step dropout warmup ratio  & 0.1   \\
        \bottomrule
    \end{tabular}
    \caption{Training hyperparameters for SD~1.5 and SDXL.}
    \label{tab:hyperparameters_sd}
\end{table}

\begin{table}[hbtp]
    \centering
    \footnotesize
    \begin{tabular}{lcc}
        \toprule
        \textbf{Hyperparameter} & \textbf{FLUX} & \textbf{Qwen} \\
        \midrule
        Base model        & FLUX.1-dev & Qwen-Image \\
        LoRA rank         & 32  & 32 \\
        Learning rate     & $1\!\times\!10^{-4}$ & $1\!\times\!10^{-4}$ \\
        Num epochs        & 2   & 2 \\
        Dataset repeat    & 50  & 50 \\
        Grad.\ accum.\ steps & 1 & 1 \\
        \bottomrule
    \end{tabular}
    \caption{Training hyperparameters for FLUX and Qwen-Image.}
    \label{tab:hyperparameters_flux_qwen}
\end{table}

\begin{figure}[htbp]
    \centering
    \begin{tabular}{cc}
        \includegraphics[width=0.47\linewidth]{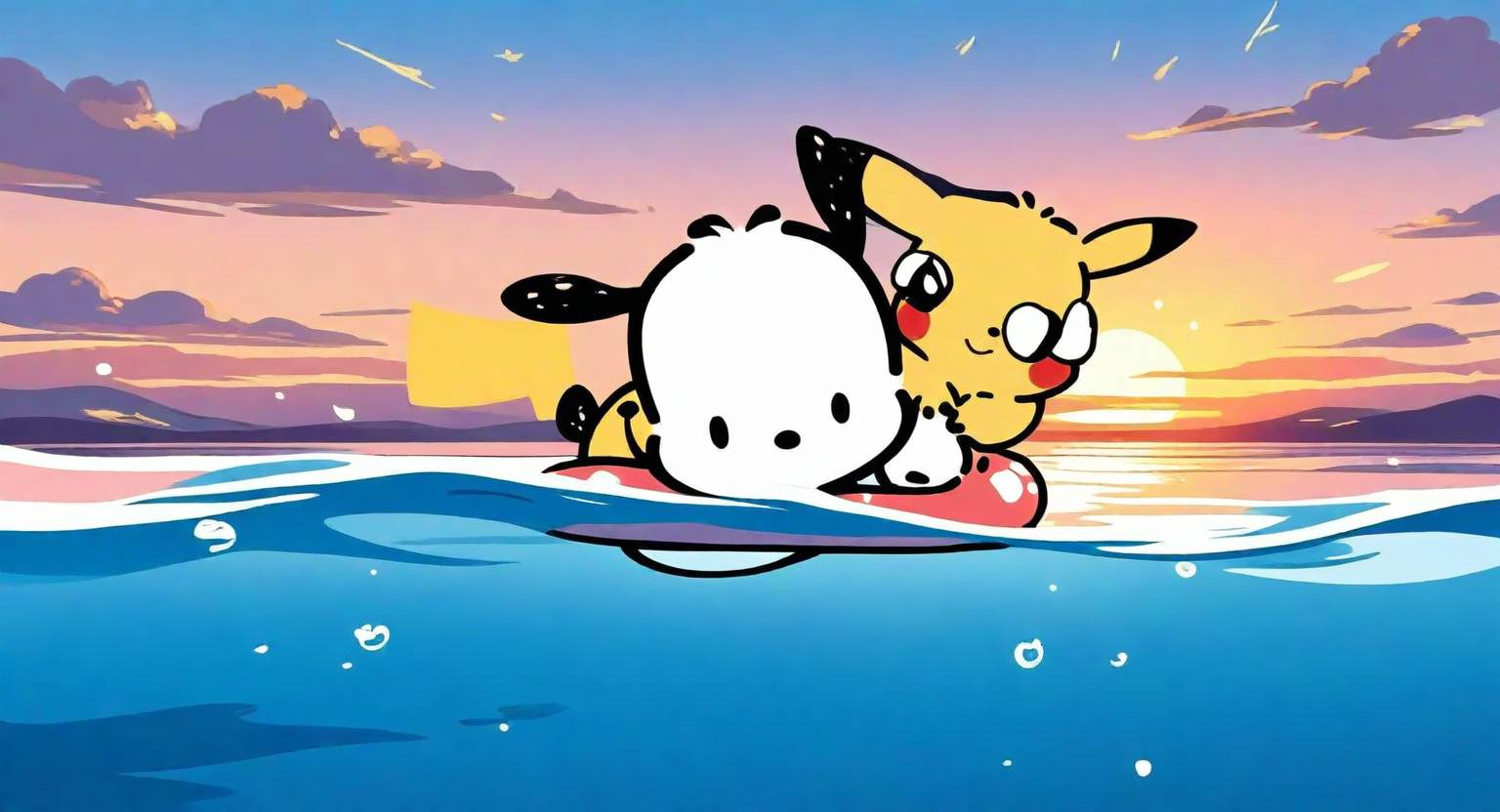} &
        \includegraphics[width=0.47\linewidth]{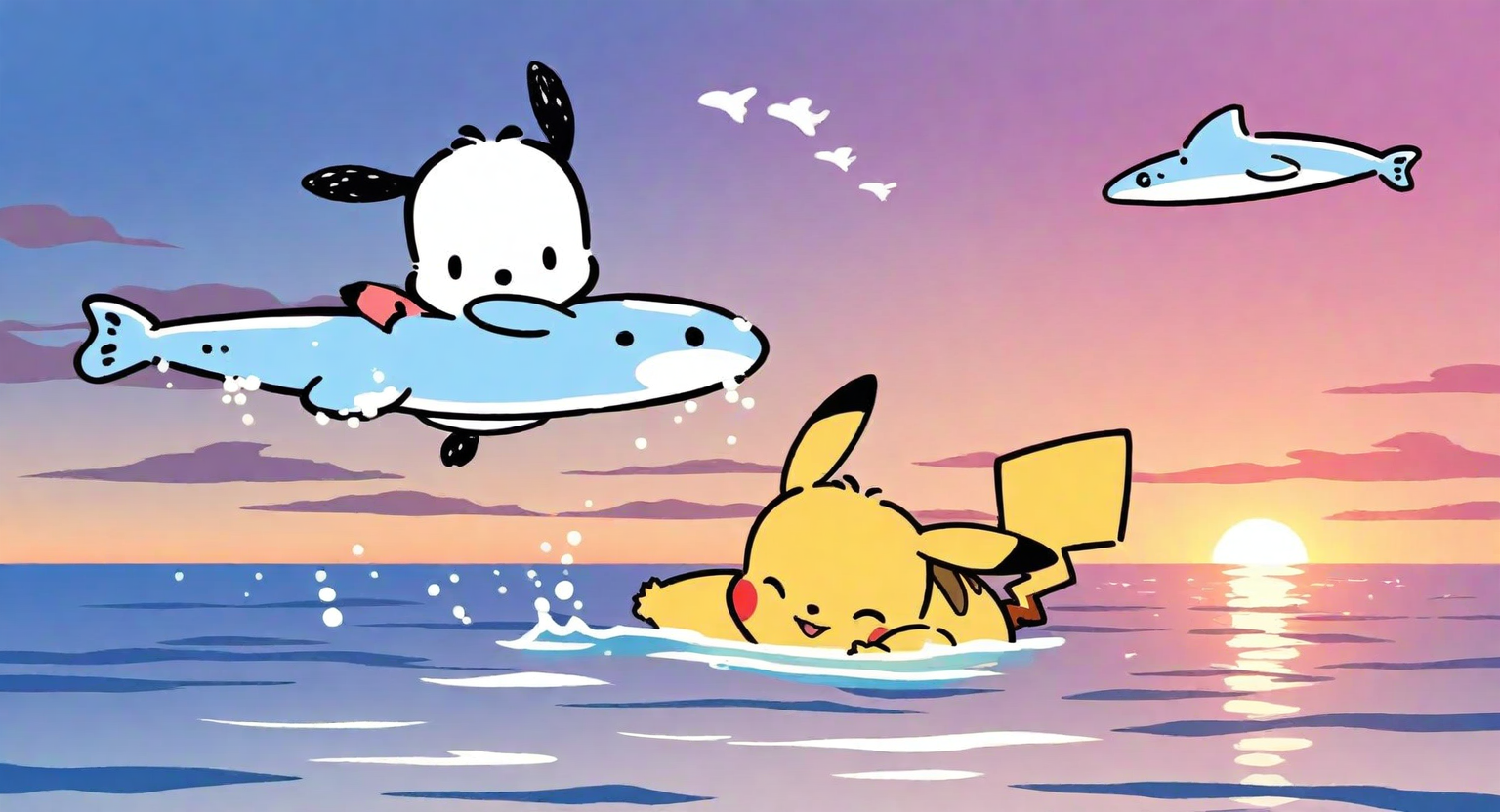} \\
        \small Normal Dropout & \small sFAD \\
    \end{tabular}
    \vspace{0.3em}
    {\footnotesize \textbf{Prompt:} \texttt{2 object, no humans, masterpiece, right is character- pikachu, left is character- pochacco, swimming, sunset}}

    \vspace{0.8em}
    \begin{tabular}{cc}
        \includegraphics[width=0.47\linewidth]{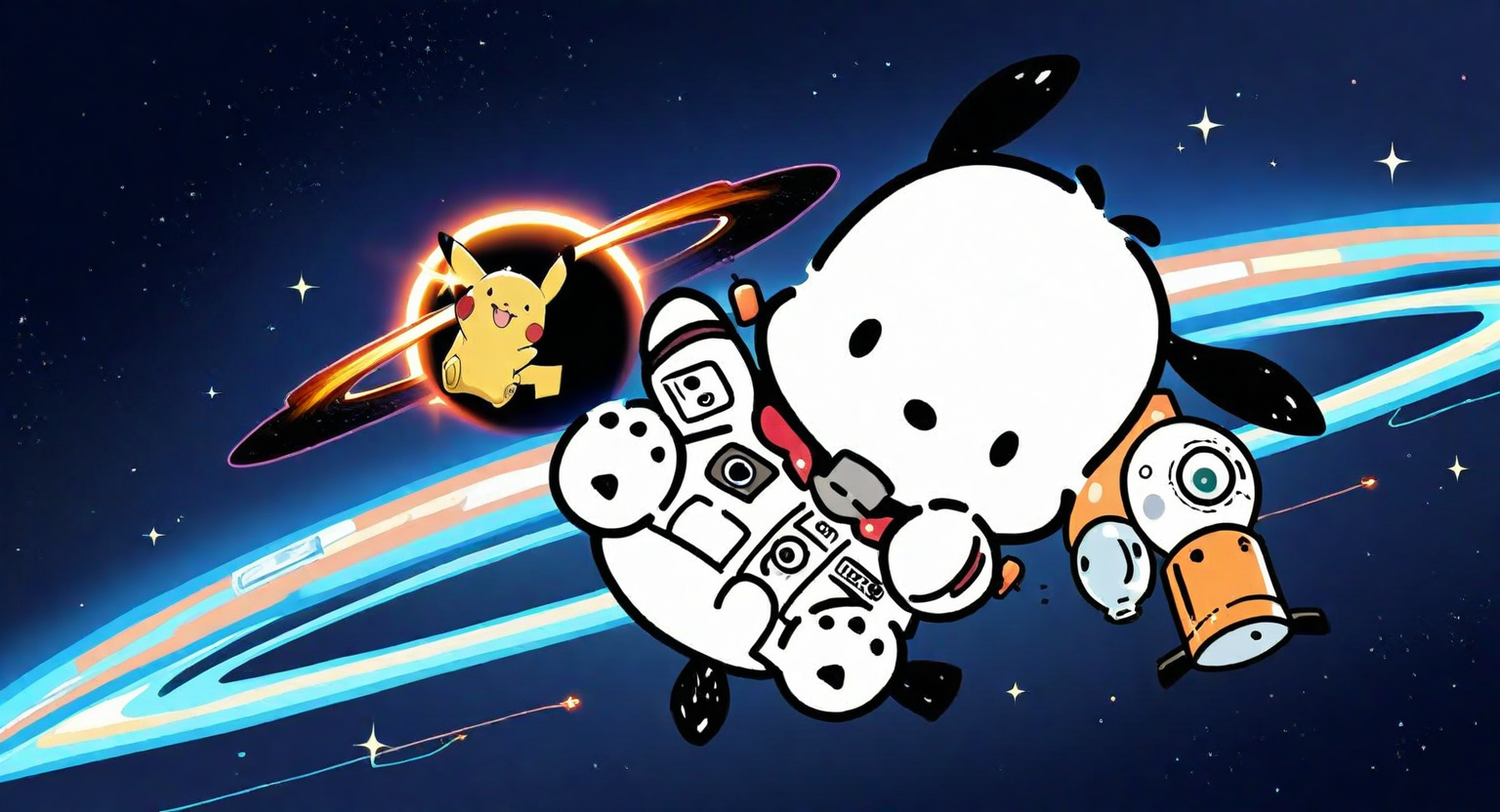} &
        \includegraphics[width=0.47\linewidth]{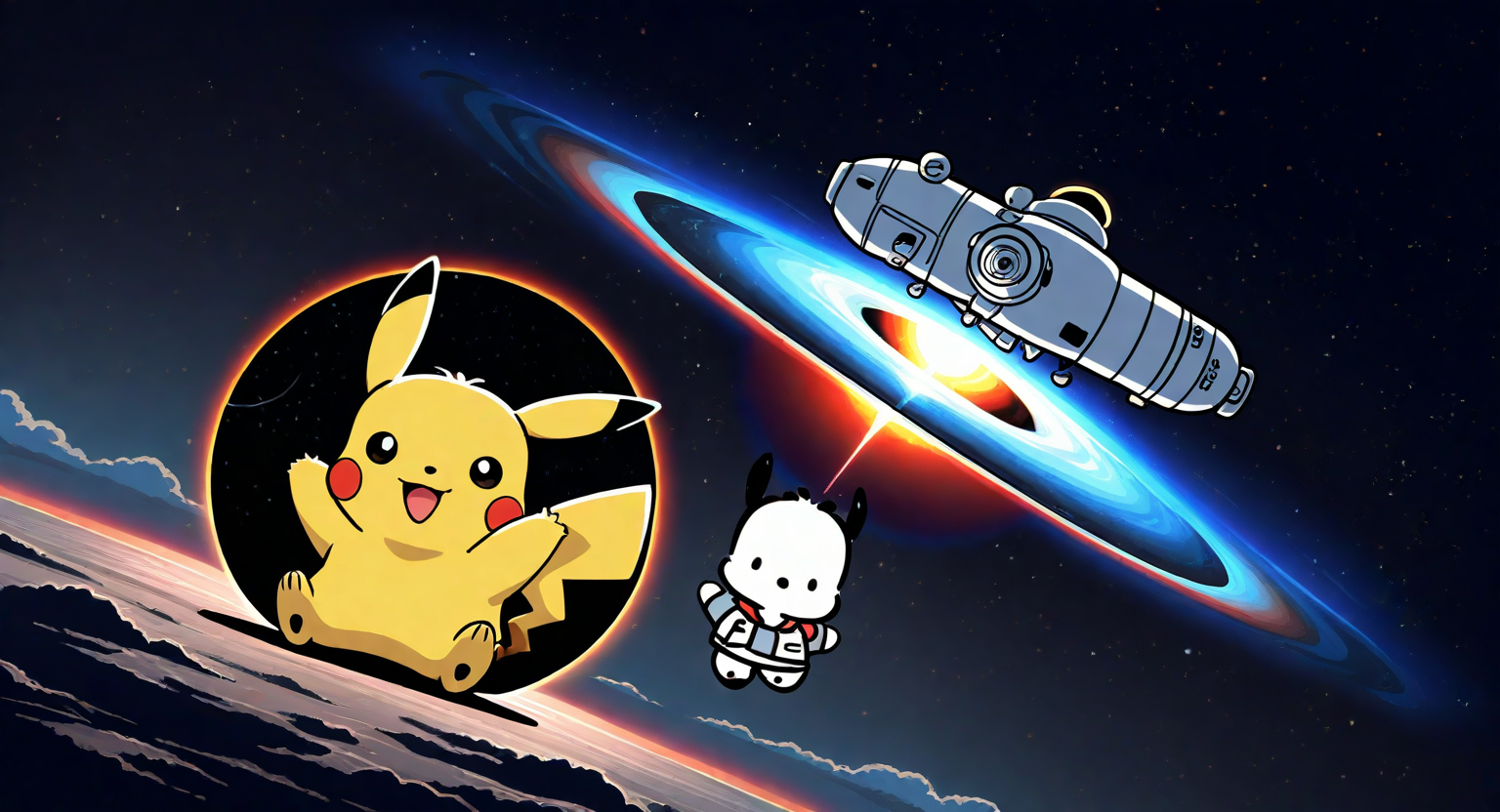} \\
        \small Normal Dropout & \small sFAD \\
    \end{tabular}
    \vspace{0.3em}
    {\footnotesize \textbf{Prompt:} \texttt{2 object, no humans, masterpiece, right is character- pochacco, left is character- pikachu, astronaut, black hole, rocket}}

    \caption{Multi-concept Training: Normal Dropout (left) vs.\ sFAD (right). The model trained with sFAD successfully reproduces both characters, whereas Normal Dropout results in noticeable deformations.}
    \label{fig:multi_eval}
\end{figure}

\section{Effect of Trigger Token on Inference Performance}
Although improving the generation capacity of LoRA models is crucial, it is equally important to prevent catastrophic forgetting of the inherent knowledge of the base model. To investigate this, we conduct an ablation experiment in which we deliberately exclude the trigger token during inference. By omitting the trigger token, we aim to observe whether the model's ability to reproduce the target identity or style diminishes, thereby assessing how tightly the learned features are bound to the trigger token. The quantitative results are reported in \cref{tab:without_trig_DINO,tab:without_trig_fid,tab:without_trig_insightface,tab:without_trig_CCIP}.

Interestingly, for the Normal Dropout setting, the scores without the trigger token are often similar to---or occasionally even higher than---those with the trigger token. This suggests that character-related features are diffusely distributed across all tokens, rather than being strongly bound to the trigger token. In contrast, for FAD and sFAD, the scores drop drastically when the trigger token is omitted, demonstrating that our method successfully concentrates the critical character features into the trigger token itself. This confirms that FAD and sFAD effectively strengthen the association between the trigger token and the intended identity or style, while Normal Dropout does not.

\begin{table*}[ht]
\centering
\small
\caption{DINO ($\uparrow$) scores with and without trigger token. Bold indicates large change of $\Delta$ (x - o).}
\begin{adjustbox}{max width=\textwidth}
\begin{tabular}{llccc|ccc|ccc|ccc|ccc|ccc}
\toprule
\textbf{Model} & \textbf{Method}
& \multicolumn{3}{c|}{\textbf{faker}}
& \multicolumn{3}{c|}{\textbf{reeves}}
& \multicolumn{3}{c|}{\textbf{hsng}}
& \multicolumn{3}{c|}{\textbf{mbst}}
& \multicolumn{3}{c|}{\textbf{pikachu}}
& \multicolumn{3}{c}{\textbf{pochacco}} \\
&
& o & x & $\Delta$
& o & x & $\Delta$
& o & x & $\Delta$
& o & x & $\Delta$
& o & x & $\Delta$
& o & x & $\Delta$ \\
\midrule
\multirow{3}{*}{SD 1.5}
& Normal
  & 0.897 & 0.899 & +0.002
  & 0.924 & 0.924 & +0.000
  & 0.896 & 0.898 & +0.002
  & 0.898 & 0.898 & +0.000
  & 0.915 & 0.919 & +0.004
  & 0.936 & 0.937 & +0.001 \\
& FAD
  & 0.907 & 0.902 & \textbf{-0.005}
  & 0.929 & 0.926 & \textbf{-0.003}
  & 0.893 & 0.893 & \textbf{+0.000}
  & 0.903 & 0.902 & \textbf{-0.001}
  & 0.917 & 0.918 & \textbf{+0.001}
  & 0.940 & 0.940 & \textbf{+0.000} \\
& sFAD
  & 0.910 & 0.906 & -0.004
  & 0.932 & 0.930 & -0.002
  & 0.893 & 0.893 & \textbf{+0.000}
  & 0.907 & 0.906 & \textbf{-0.001}
  & 0.922 & 0.924 & +0.002
  & 0.944 & 0.944 & \textbf{+0.000} \\
\midrule
\multirow{3}{*}{SDXL}
& Normal
  & 0.9037 & 0.9029 & -0.0008
  & 0.9238 & 0.9224 & -0.0014
  & 0.9013 & 0.8960 & \textbf{-0.0053}
  & 0.9025 & 0.9017 & -0.0008
  & 0.8509 & 0.8592   & +0.0083
  & 0.8582 & 0.8561   & -0.0024 \\
& FAD
  & 0.9061 & 0.9058 & -0.0003
  & 0.9322 & 0.9296 & \textbf{-0.0026}
  & 0.9005 & 0.9028 & +0.0023
  & 0.9115 & 0.9094 & \textbf{-0.0021}
  & 0.8628 & 0.8522   & \textbf{-0.0106}
  & 0.8828 & 0.8671   & \textbf{-0.0157} \\
& sFAD
  & 0.9107 & 0.9096 & \textbf{-0.0011}
  & 0.9325 & 0.9307 & -0.0018
  & 0.9034 & 0.9035 & +0.0001
  & 0.9124 & 0.9111   & -0.0013
  & 0.8671 & 0.8585   & -0.0086
  & 0.8668 & 0.8673   & +0.0005 \\
\bottomrule
\end{tabular}
\label{tab:without_trig_DINO}
\end{adjustbox}
\end{table*}

\begin{table*}[ht]
\centering
\small
\caption{FID ($\downarrow$) scores with and without trigger token. Bold indicates the largest change of $\Delta$ (x - o).}
\begin{adjustbox}{max width=\textwidth}
\begin{tabular}{llccc|ccc|ccc|ccc|ccc|ccc}
\toprule
\textbf{Model} & \textbf{Method}
& \multicolumn{3}{c|}{\textbf{faker}}
& \multicolumn{3}{c|}{\textbf{reeves}}
& \multicolumn{3}{c|}{\textbf{hsng}}
& \multicolumn{3}{c|}{\textbf{mbst}}
& \multicolumn{3}{c|}{\textbf{pikachu}}
& \multicolumn{3}{c}{\textbf{pochacco}} \\
&
& o & x & $\Delta$
& o & x & $\Delta$
& o & x & $\Delta$
& o & x & $\Delta$
& o & x & $\Delta$
& o & x & $\Delta$ \\
\midrule
\multirow{3}{*}{SD 1.5}
& Normal
  & 207.103 & 206.449 & -0.654
  & 210.532 & 210.463 & -0.069
  & 164.259 & 163.539 & -0.720
  & 137.473 & 136.017 & -1.456
  & 137.473 & 135.702 & -1.771
  & 136.017 & 135.404 & -0.613 \\
& Adaptive
  & 197.462 & 204.428 & \textbf{+6.966}
  & 203.077 & 204.407 & +1.330
  & 161.200 & 162.812 & +1.612
  & 137.760 & 133.723 & \textbf{-4.037}
  & 137.760 & 138.203 & +0.443
  & 133.723 & 137.205 & \textbf{+3.482} \\
& Step
  & 195.863 & 197.097 & +1.234
  & 198.524 & 200.642 & \textbf{+2.118}
  & 160.478 & 161.665 & \textbf{+1.187}
  & 136.746 & 131.513 & \textbf{-5.233}
  & 136.746 & 137.327 & \textbf{+0.581}
  & 131.513 & 134.811 & +3.298 \\
\midrule
\multirow{3}{*}{SDXL}
& Normal
  & 199.310 & 199.970 & +0.660
  & 208.110 & 209.600 & \textbf{+1.490}
  & 166.350 & 166.390 & \textbf{+0.040}
  & 236.945 & 234.35 & -2.595
  & 170.046 & 164.853 & -5.194
  & 193.026 & 194.358 & +1.332 \\
& Adaptive
  & 194.690 & 199.730 & \textbf{+5.040}
  & 196.300 & 196.150 & -0.150
  & 160.800 & 159.750 & -1.050
  & 216.498 & 218.28 & \textbf{+1.782}
  & 160.946 & 165.081 & \textbf{+4.134}
  & 182.566 & 193.153 & \textbf{+10.587} \\
& Step
  & 197.760 & 197.720 & -0.040
  & 199.090 & 197.320 & -1.770
  & 162.650 & 161.010 & -1.640
  & 217.34 & 216.95 & -0.39
  & 164.224 & 165.339 & +1.115
  & 184.087 & 193.974 & +9.887 \\
\bottomrule
\end{tabular}
\label{tab:without_trig_fid}
\end{adjustbox}
\end{table*}

\begin{table*}[htb]
\centering
\small
\caption{InsightFace ($\downarrow$) scores with and without trigger token. Bold indicates the largest increase in $\Delta$ (x - o).}
\begin{adjustbox}{max width=\textwidth}
\begin{tabular}{llccc|ccc|ccc|ccc}
\toprule
\textbf{Model} & \textbf{Method}
& \multicolumn{3}{c|}{\textbf{faker}}
& \multicolumn{3}{c|}{\textbf{reeves}}
& \multicolumn{3}{c|}{\textbf{hsng}}
& \multicolumn{3}{c|}{\textbf{mbst}} \\
&
& o & x & $\Delta$
& o & x & $\Delta$
& o & x & $\Delta$
& o & x & $\Delta$ \\
\midrule
\multirow{3}{*}{SD 1.5}
& Normal
  & 28.064 & 28.202 & +0.138
  & 30.619 & 30.999 & +0.380
  & 27.792 & 27.655 & -0.137
  & 28.015 & 28.087 & +0.072  \\
& Adaptive
  & 26.997 & 27.826 & \textbf{+0.829}
  & 30.236 & 31.131 & \textbf{+0.895}
  & 27.511 & 27.596 & +0.085
  & 28.058 & 28.098 & +0.040 \\
& Step
  & 26.954 & 27.655 & +0.701
  & 30.192 & 30.922 & +0.730
  & 27.433 & 27.522 & \textbf{+0.089}
  & 27.835 & 27.908 & \textbf{+0.073} \\
\midrule
\multirow{3}{*}{SDXL}
& Normal
  & 25.510 & 25.560 & +0.050
  & 27.070 & 26.570 & -0.500
  & 24.960 & 24.990 & +0.030
  & 27.280 & 27.300 & \textbf{+0.020} \\
& Adaptive
  & 24.470 & 25.320 & \textbf{+0.850}
  & 26.290 & 26.960 & \textbf{+0.670}
  & 24.410 & 24.820 & \textbf{+0.410}
  & 25.440 & 25.410 & -0.030 \\
& Step
  & 25.130 & 25.160 & +0.030
  & 25.930 & 26.250 & +0.320
  & 24.530 & 24.490 & -0.040
  & 24.550 & 24.570 & \textbf{+0.020} \\
\bottomrule
\end{tabular}
\label{tab:without_trig_insightface}
\end{adjustbox}
\end{table*}

\begin{table}[htb]
\centering
\caption{CCIP ($\uparrow$) scores with and without trigger token. Bold indicates the largest decrease in $\Delta$ (x - o).}
\resizebox{\columnwidth}{!}{%
\begin{tabular}{llccc|ccc}
\toprule
\textbf{Model} & \textbf{Method}
& \multicolumn{3}{c|}{\textbf{Pikachu}}
& \multicolumn{3}{c}{\textbf{Pochacco}} \\
&
& o & x & $\Delta$
& o & x & $\Delta$ \\
\midrule
\multirow{3}{*}{SD 1.5}
& Normal
  & 0.9987 & 0.9987 & +0.0000
  & 0.9431 & 0.9350 & \textbf{-0.0081} \\
& Adaptive
  & 0.9987 & 0.9987 & +0.0000
  & 0.9528 & 0.9487 & -0.0041 \\
& Step
  & 0.9987 & 0.9986 & \textbf{-0.0001}
  & 0.9487 & 0.9428 & -0.0059 \\
\midrule
\multirow{3}{*}{SDXL}
& Normal
  & 0.9992 & 0.9972 & -0.0020
  & 0.8996 & 0.8341 & -0.0655 \\
& Adaptive
  & 0.9996 & 0.9451 & -0.0545
  & 0.9091 & 0.7514 & -0.1577 \\
& Step
  & 1.0000 & 0.9412 & \textbf{-0.0588}
  & 0.9295 & 0.7708 & \textbf{-0.1587} \\
\bottomrule
\end{tabular}%
}
\label{tab:without_trig_CCIP}
\end{table}

\section{Detailed Example of GPT-4.1 Evaluation}
For this evaluation, we follow the evaluation prompt from~\cite{zhong2024multiloracompositionimagegeneration} and the representative images are selected from the model outputs at the 1300th training step. \cref{tab:gpt4v_eval_mbst} shows the complete result used for the evaluation, and \cref{tab:gpt4v_eval_result_mbst} presents the complete evaluation result corresponding to \cref{tab:gpt4v_eval_mbst}.

\begin{figure}[htb]
    \centering
    \begin{subfigure}{0.33\linewidth}
        \centering
        \includegraphics[width=\linewidth]{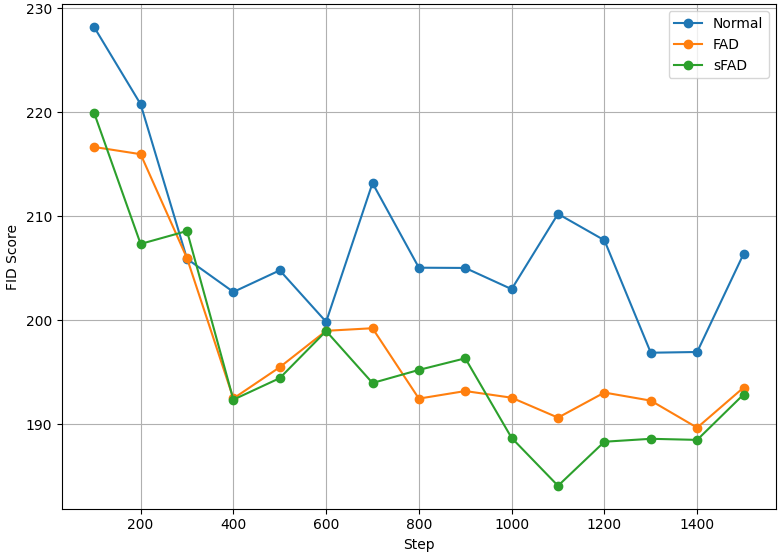}
        \caption{faker}
    \end{subfigure}%
    \begin{subfigure}{0.33\linewidth}
        \centering
        \includegraphics[width=\linewidth]{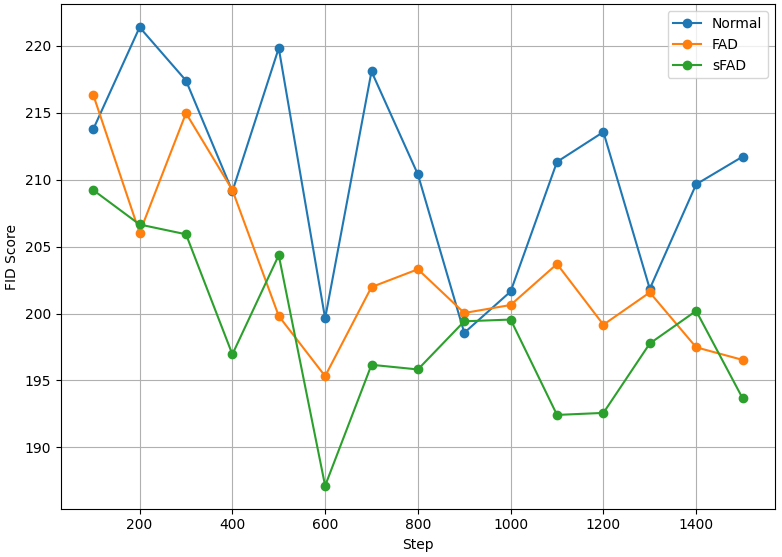}
        \caption{reeves}
    \end{subfigure}%
    \begin{subfigure}{0.33\linewidth}
        \centering
        \includegraphics[width=\linewidth]{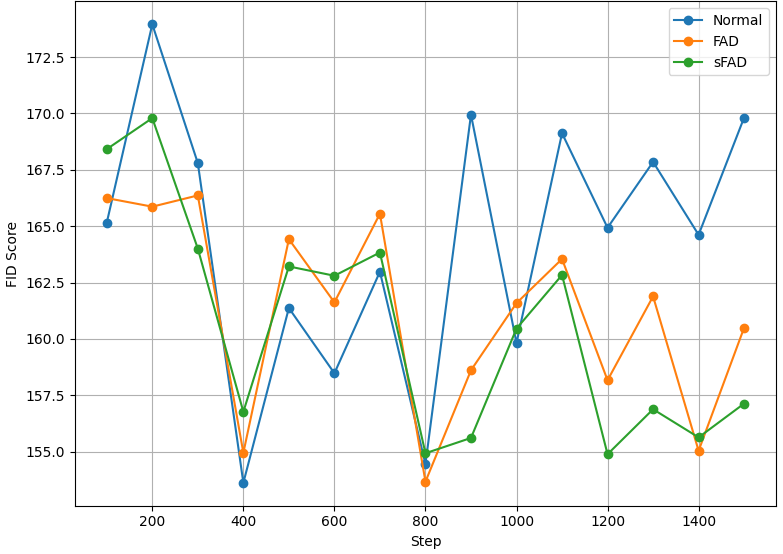}
        \caption{hsng}
    \end{subfigure}

    \begin{subfigure}{0.33\linewidth}
        \centering
        \includegraphics[width=\linewidth]{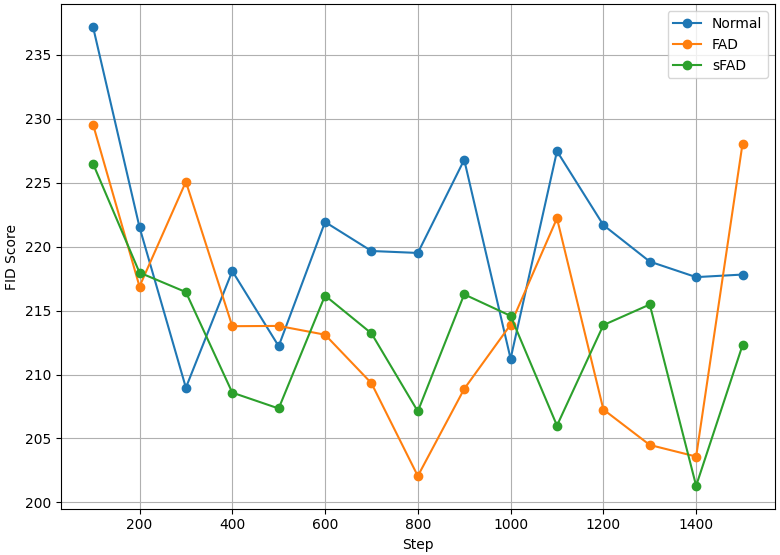}
        \caption{mbst}
    \end{subfigure}%
    \begin{subfigure}{0.33\linewidth}
        \centering
        \includegraphics[width=\linewidth]{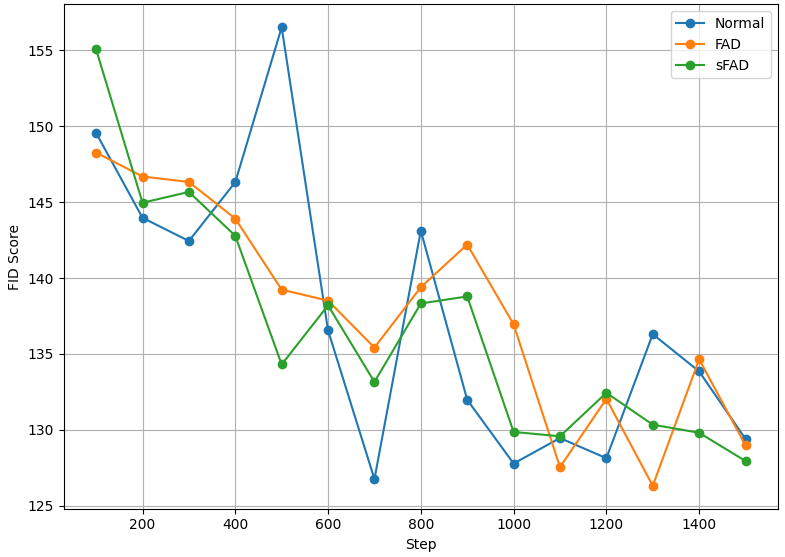}
        \caption{pikachu}
    \end{subfigure}%
    \begin{subfigure}{0.33\linewidth}
        \centering
        \includegraphics[width=\linewidth]{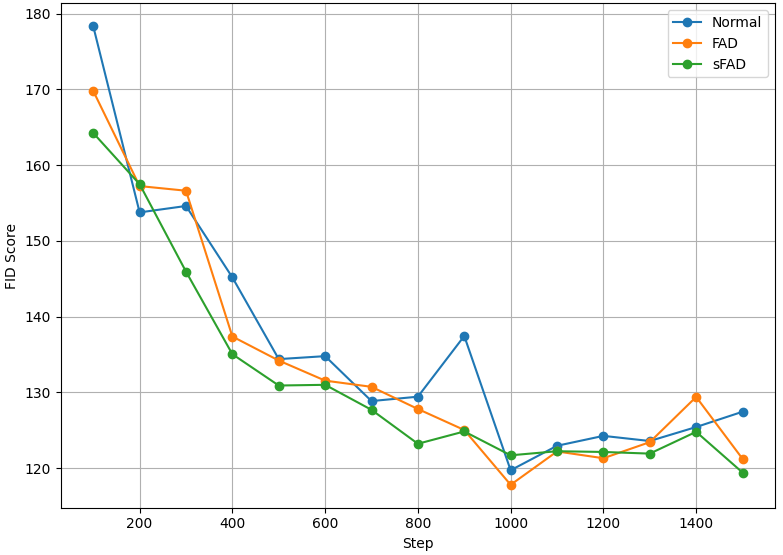}
        \caption{pochacco}
    \end{subfigure}
    \caption{Comparison of FID ($\downarrow$) scores of three methods across all steps for six datasets, extracted from \textbf{SD 1.5}.}
    \label{fig:FID_comparison_15}
\end{figure}

\section{Dropout Scheduling Strategies}

In \cref{tab:linear_experiments}, the proposed sFAD method employs an exponential scheduling function to gradually increase the dropout rate during training, ranging from 0.1 to 0.8. To further validate the robustness of our approach, we also conduct experiments using a linear scheduling strategy.

Beyond the choice of scheduling function, we investigate how the magnitude of dropout probabilities should vary across diffusion timesteps. Specifically, we define two distinct dropout ranges: (1) 0.1 $\to$ 0.8 and (2) 0.8 $\to$ 0.1. In the first setting (0.1 $\to$ 0.8), training begins with a lower dropout rate to prioritize generalization in the early timesteps, and the rate gradually increases to promote disentanglement between the trigger token and surrounding tokens, following the principles of FAD. Conversely, in the second setting (0.8 $\to$ 0.1), we apply a higher dropout rate in the early timesteps to encourage token disentanglement, and then gradually reduce it in later steps to stabilize learning and preserve fine-grained visual fidelity.

As part of future work, we plan to explore additional scheduling strategies, such as cyclical or adaptive schedules, to further improve the balance between generalization and visual fidelity.

\begin{table}[ht]
\centering
\caption{Comparison of SD 1.5 and SDXL for Step (0.1\textasciitilde0.8) vs Step (0.8\textasciitilde0.1) using Linear / Exp\_up methods.}
\resizebox{\columnwidth}{!}{%
\begin{tabular}{cc|l|cc|cc}
\toprule
\textbf{Char.} & \textbf{Step} & \textbf{Metric}
& \multicolumn{2}{c|}{\textbf{SD 1.5}}
& \multicolumn{2}{c}{\textbf{SDXL}} \\
& &  & Linear & Exp\_up & Linear & Exp\_up \\
\midrule
\multirow{6}*{\centering Faker} & \multirow{3}*{\centering 0.1$\sim$0.8}
& FID ($\downarrow$)         & 196.119 & 195.863 & \textbf{191.29} & 197.76 \\
& & DINO ($\uparrow$)        & 0.9100  & 0.9100  & \textbf{0.9131} & 0.9107 \\
& & InsightFace ($\downarrow$) & 26.976  & \textbf{26.954}  & \textbf{24.51}  & 25.13 \\
\cmidrule{2-7}
& \multirow{3}*{\centering 0.8$\sim$0.1}
& FID ($\downarrow$)         & 195.411 & \textbf{194.655} & 196.85 & \textbf{194.93} \\
& & DINO ($\uparrow$)        & \textbf{0.9107}  & 0.9100  & 0.9087 & \textbf{0.9120} \\
& & InsightFace ($\downarrow$) & 26.967  & 26.984  & \textbf{25.23}  & 25.87 \\
\midrule
\multirow{6}*{\centering Reeves} & \multirow{3}*{\centering 0.1$\sim$0.8}
& FID ($\downarrow$)         & 199.778 & \textbf{198.524} & \textbf{194.17} & 199.09 \\
& & DINO ($\uparrow$)        & \textbf{0.9320}  & \textbf{0.9320}  & \textbf{0.9336} & 0.9325 \\
& & InsightFace ($\downarrow$) & 30.038  & 30.192  & \textbf{25.85}  & 25.93 \\
\cmidrule{2-7}
& \multirow{3}*{\centering 0.8$\sim$0.1}
& FID ($\downarrow$)         & 201.095 & 201.838 & \textbf{195.26} & 198.06 \\
& & DINO ($\uparrow$)        & \textbf{0.9320}  & 0.9310  & \textbf{0.9342} & 0.9322 \\
& & InsightFace ($\downarrow$) & \textbf{29.950}  & 29.995  & \textbf{25.53}  & 26.22 \\
\midrule
\multirow{6}*{\centering hsng} & \multirow{3}*{\centering 0.1$\sim$0.8}
& FID ($\downarrow$)         & 161.579 & \textbf{160.478} & \textbf{162.49} & 162.65 \\
& & DINO ($\uparrow$)        & \textbf{0.8930}  & \textbf{0.8930}  & \textbf{0.9062} & 0.9035 \\
& & InsightFace ($\downarrow$) & 27.503  & \textbf{27.433}  & \textbf{24.25}  & 24.53 \\
\cmidrule{2-7}
& \multirow{3}*{\centering 0.8$\sim$0.1}
& FID ($\downarrow$)         & 160.742 & 161.246 & 162.92 & \textbf{158.16} \\
& & DINO ($\uparrow$)        & 0.8940  & \textbf{0.8930}  & \textbf{0.9069} & 0.9034 \\
& & InsightFace ($\downarrow$) & 27.451  & 27.479  & 24.59  & \textbf{24.36} \\
\midrule
\multirow{6}*{\centering mbst} & \multirow{3}*{\centering 0.1$\sim$0.8}
& FID ($\downarrow$)         & 212.874 & 212.874 & \textbf{210.85} & 217.34 \\
& & DINO ($\uparrow$)        & 0.9060  & \textbf{0.9070}  & \textbf{0.9159} & 0.9124 \\
& & InsightFace ($\downarrow$) & 27.955  & \textbf{27.835}  & \textbf{24.27}  & 24.55 \\
\cmidrule{2-7}
& \multirow{3}*{\centering 0.8$\sim$0.1}
& FID ($\downarrow$)         & \textbf{212.153} & \textbf{212.153} & 217.08 & \textbf{213.92} \\
& & DINO ($\uparrow$)        & 0.9040  & 0.9050  & \textbf{0.9099} & 0.8945 \\
& & InsightFace ($\downarrow$) & 27.898  & 27.862  & 25.67  & \textbf{24.31} \\
\bottomrule
\end{tabular}%
}
\label{tab:linear_experiments}
\end{table}

\section{Multi-concept Training}
We also train LoRA using Normal Dropout and sFAD on multiple datasets for SDXL, specifically \textbf{pikachu} and \textbf{pochacco}. As shown in \cref{fig:multi_eval}, the model trained with sFAD successfully reproduces both characters, whereas the model trained with Normal Dropout fails to do so, resulting in noticeable character deformations.

\section{Score Comparison}
\begin{figure}[htb]
    \centering
    \begin{subfigure}{0.33\linewidth}
        \centering
        \includegraphics[width=\linewidth]{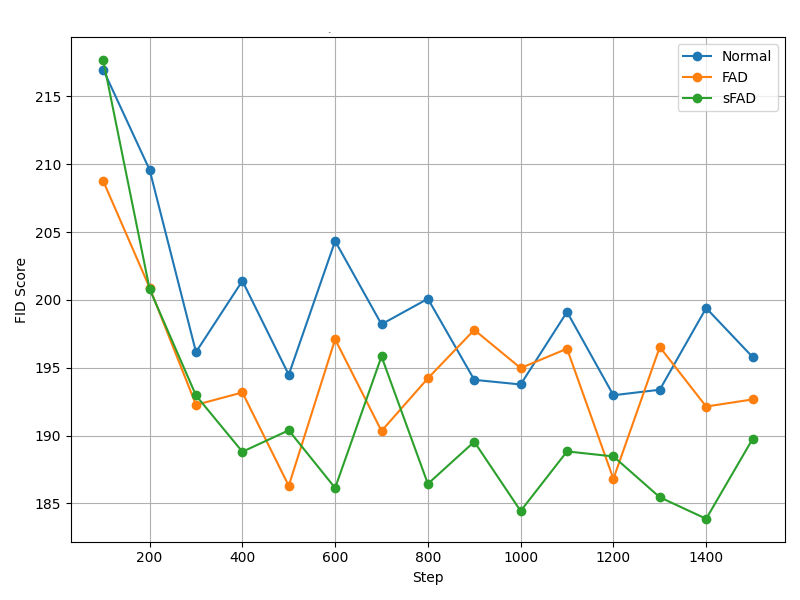}
        \caption{faker}
    \end{subfigure}%
    \begin{subfigure}{0.33\linewidth}
        \centering
        \includegraphics[width=\linewidth]{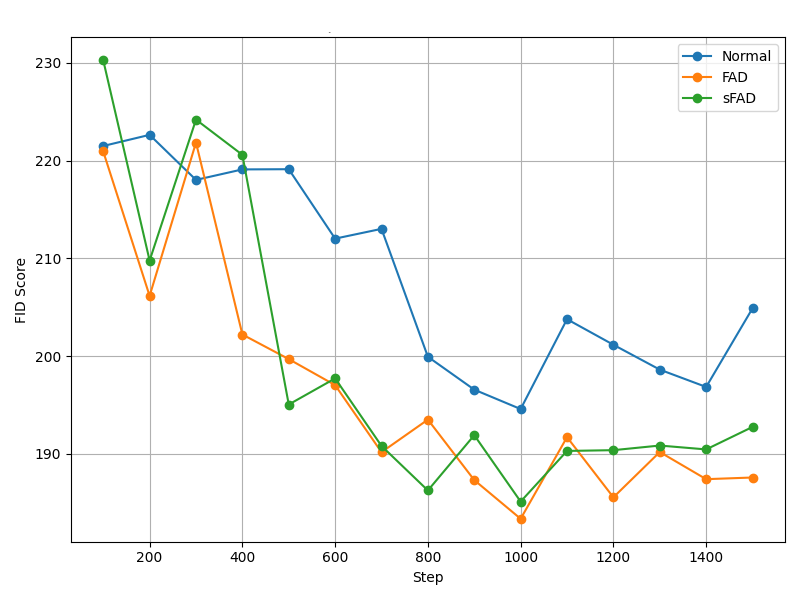}
        \caption{reeves}
    \end{subfigure}%
    \begin{subfigure}{0.33\linewidth}
        \centering
        \includegraphics[width=\linewidth]{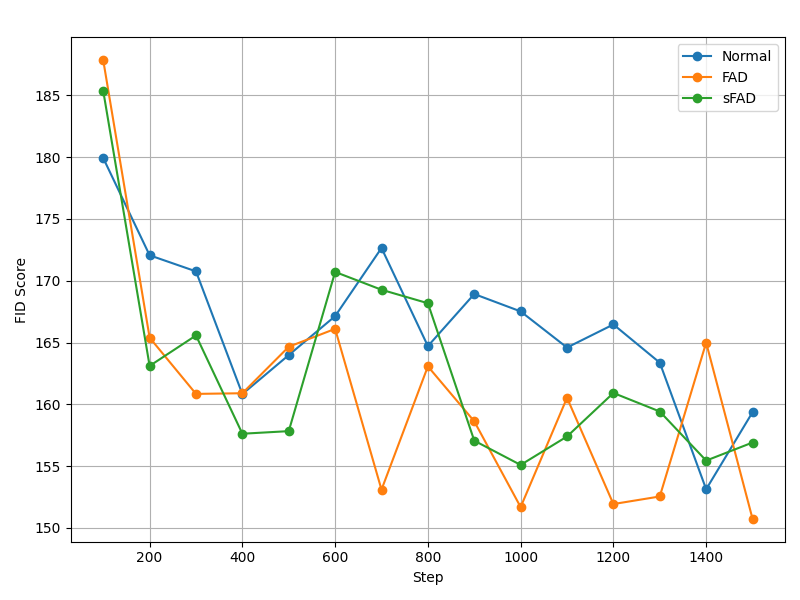}
        \caption{hsng}
    \end{subfigure}

    \begin{subfigure}{0.33\linewidth}
        \centering
        \includegraphics[width=\linewidth]{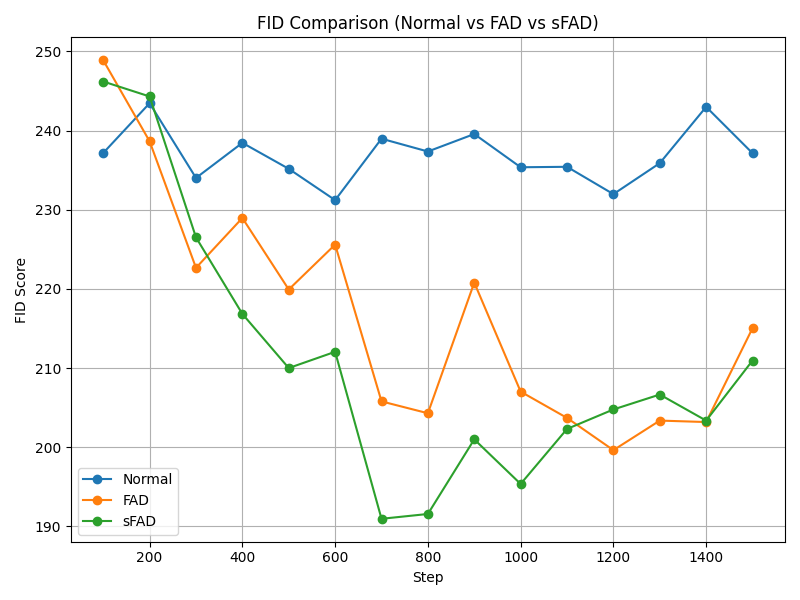}
        \caption{mbst}
    \end{subfigure}%
    \begin{subfigure}{0.33\linewidth}
        \centering
        \includegraphics[width=\linewidth]{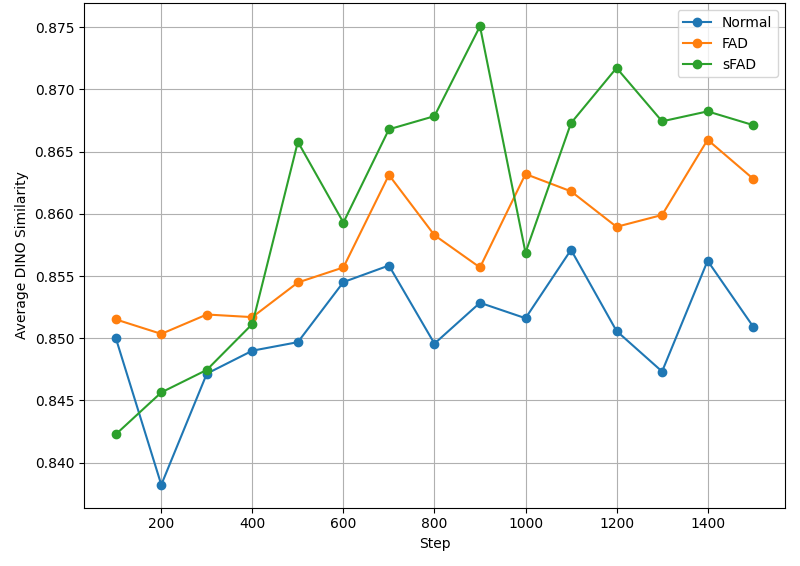}
        \caption{pikachu}
    \end{subfigure}%
    \begin{subfigure}{0.33\linewidth}
        \centering
        \includegraphics[width=\linewidth]{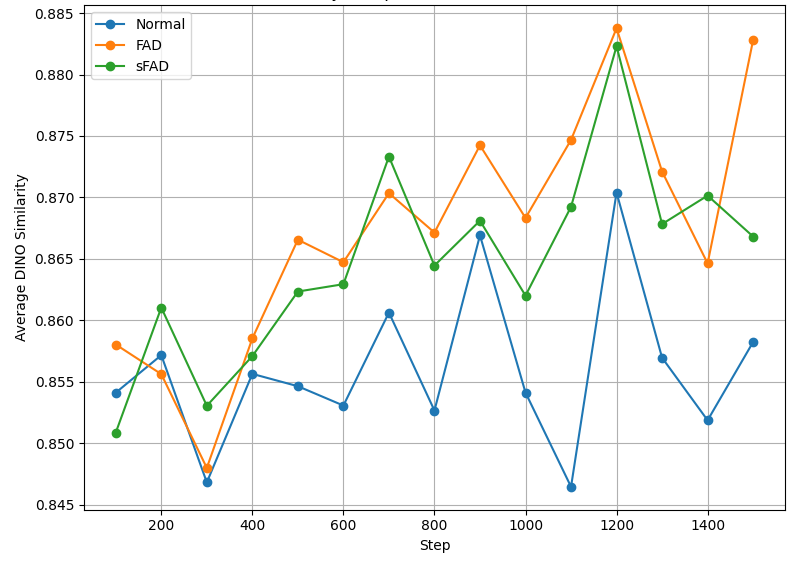}
        \caption{pochacco}
    \end{subfigure}
    \caption{Comparison of FID ($\downarrow$) scores of three methods across all steps for six datasets, extracted from \textbf{SDXL}.}
    \label{fig:FID_comparison_xl}
\end{figure}

\begin{figure}[htb]
    \centering
    \begin{subfigure}{0.33\linewidth}
        \centering
        \includegraphics[width=\linewidth]{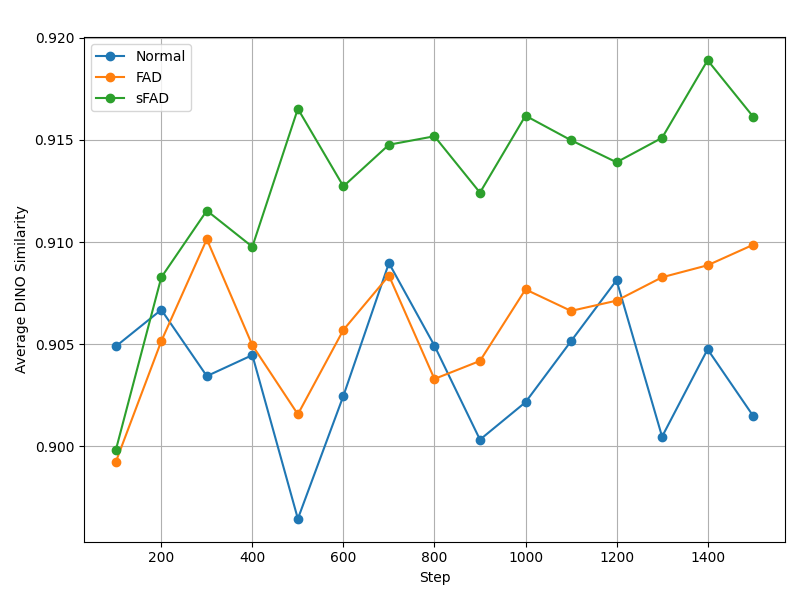}
        \caption{faker}
    \end{subfigure}%
    \begin{subfigure}{0.33\linewidth}
        \centering
        \includegraphics[width=\linewidth]{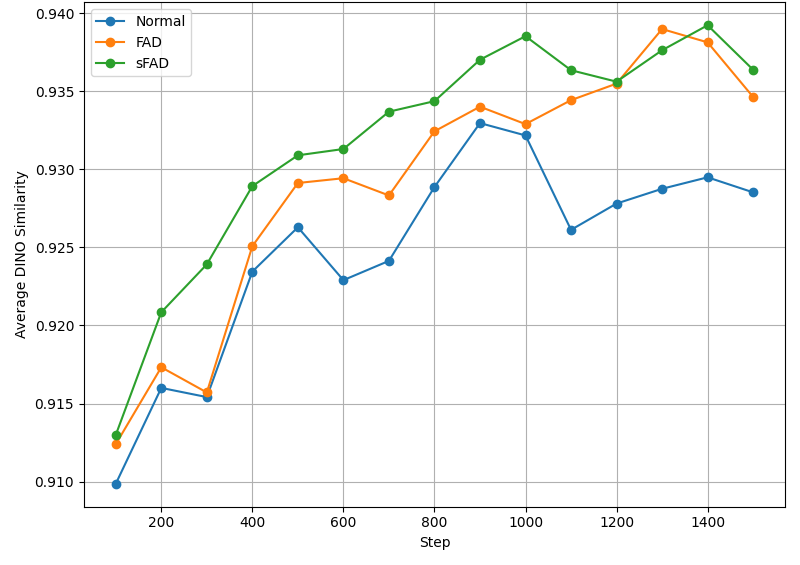}
        \caption{reeves}
    \end{subfigure}%
    \begin{subfigure}{0.33\linewidth}
        \centering
        \includegraphics[width=\linewidth]{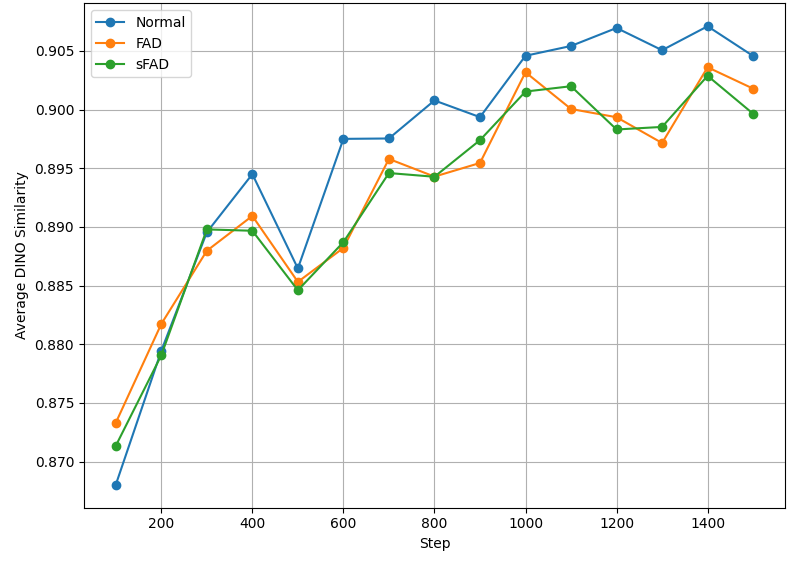}
        \caption{hsng}
    \end{subfigure}

    \begin{subfigure}{0.33\linewidth}
        \centering
        \includegraphics[width=\linewidth]{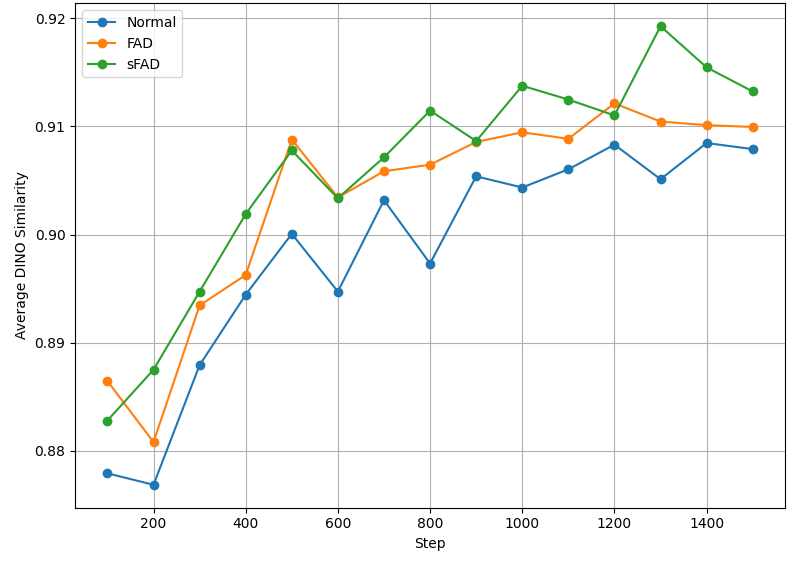}
        \caption{mbst}
    \end{subfigure}%
    \begin{subfigure}{0.33\linewidth}
        \centering
        \includegraphics[width=\linewidth]{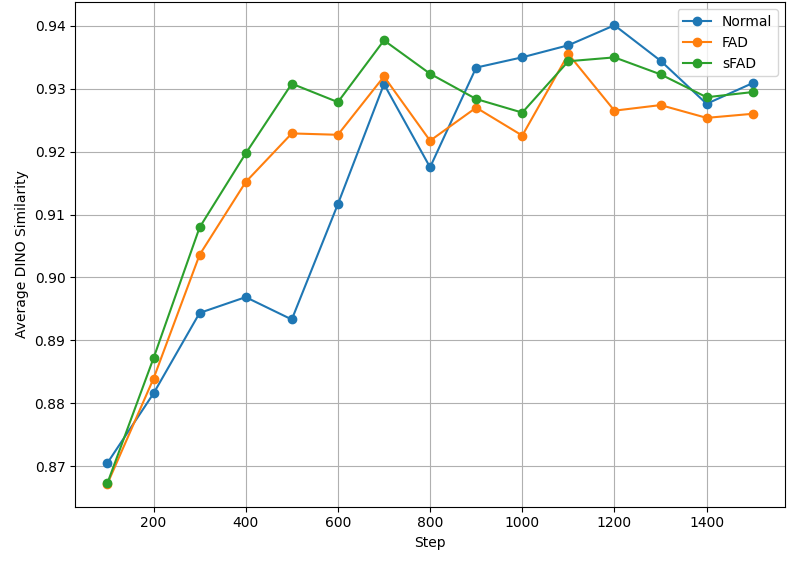}
        \caption{pikachu}
    \end{subfigure}%
    \begin{subfigure}{0.33\linewidth}
        \centering
        \includegraphics[width=\linewidth]{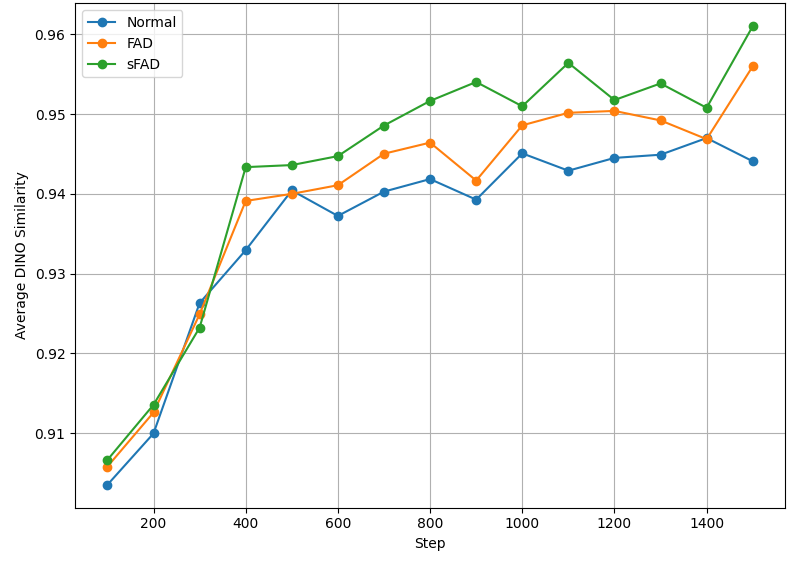}
        \caption{pochacco}
    \end{subfigure}
    \caption{Comparison of DINO ($\uparrow$) scores of three methods across all steps for six datasets, extracted from \textbf{SD 1.5}.}
    \label{fig:DINO_comparison_15}
\end{figure}

\begin{figure}[htb]
    \centering
    \begin{subfigure}{0.33\linewidth}
        \centering
        \includegraphics[width=\linewidth]{figures/DINO_score_graph/faker_dino_comparison.png}
        \caption{faker}
    \end{subfigure}%
    \begin{subfigure}{0.33\linewidth}
        \centering
        \includegraphics[width=\linewidth]{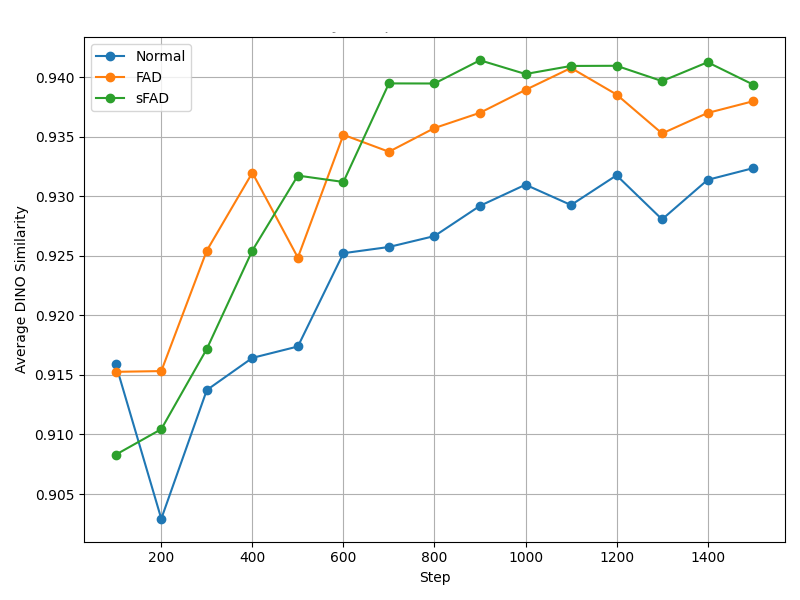}
        \caption{reeves}
    \end{subfigure}%
    \begin{subfigure}{0.33\linewidth}
        \centering
        \includegraphics[width=\linewidth]{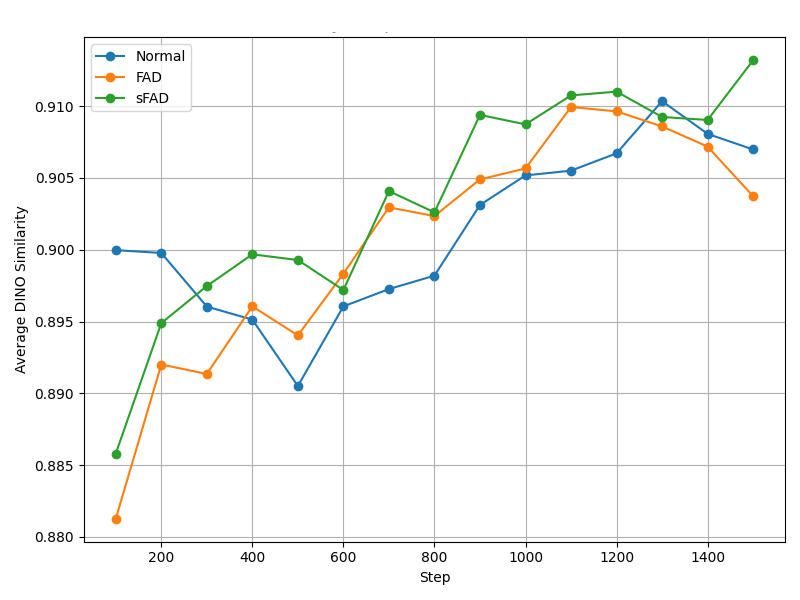}
        \caption{hsng}
    \end{subfigure}

    \begin{subfigure}{0.33\linewidth}
        \centering
        \includegraphics[width=\linewidth]{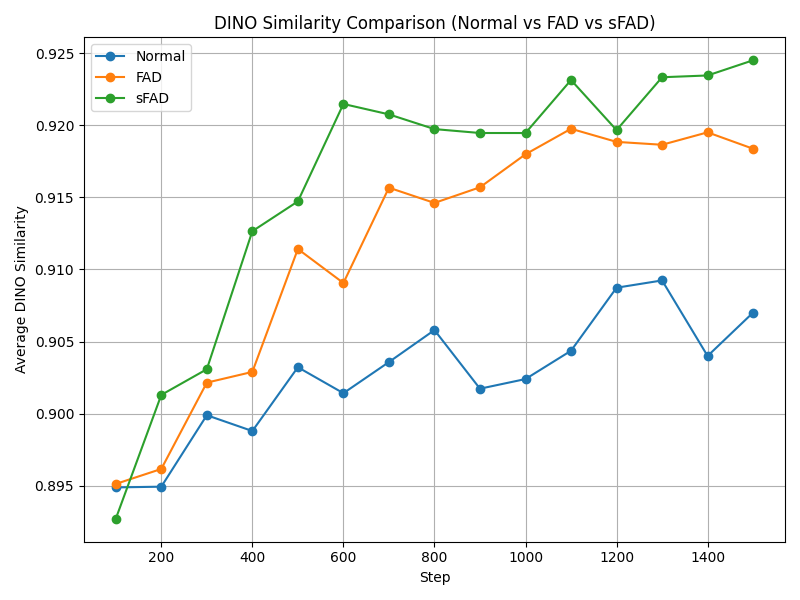}
        \caption{mbst}
    \end{subfigure}%
    \begin{subfigure}{0.33\linewidth}
        \centering
        \includegraphics[width=\linewidth]{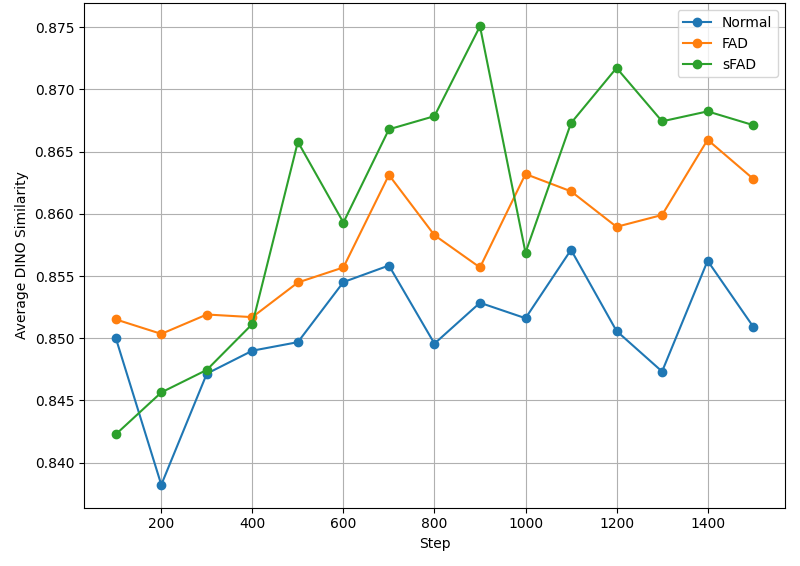}
        \caption{pikachu}
    \end{subfigure}%
    \begin{subfigure}{0.33\linewidth}
        \centering
        \includegraphics[width=\linewidth]{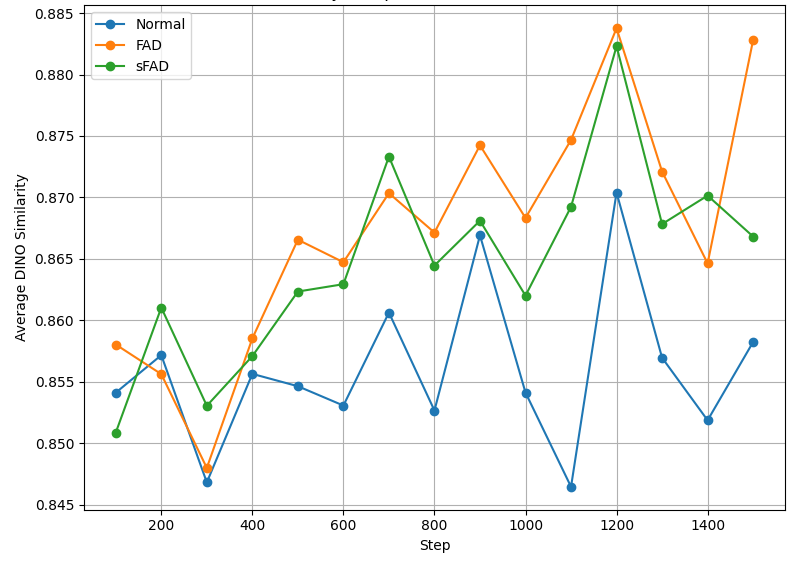}
        \caption{pochacco}
    \end{subfigure}
    \caption{Comparison of DINO ($\uparrow$) scores of three methods across all steps for six datasets, extracted from \textbf{SDXL}.}
    \label{fig:DINO_comparison_xl}
\end{figure}

\begin{figure}[htb]
    \centering
    \begin{subfigure}{0.43\linewidth}
        \centering
        \includegraphics[width=\linewidth]{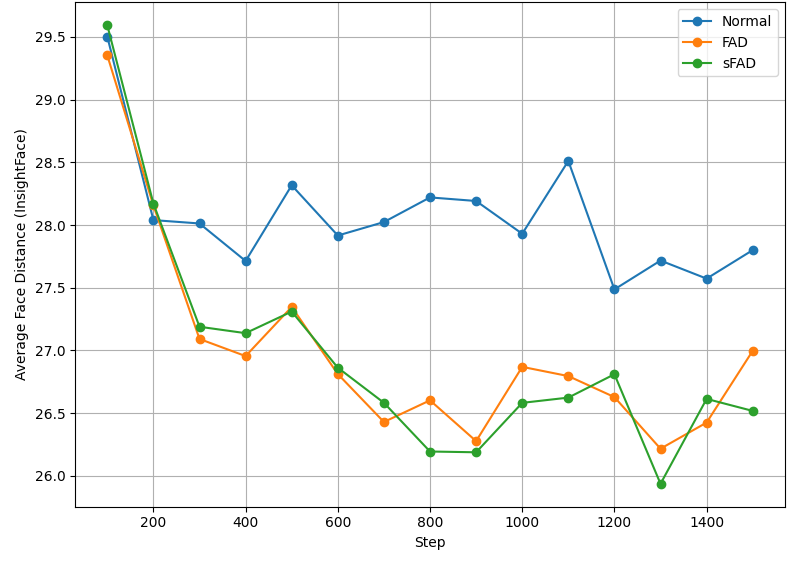}
        \caption{faker}
    \end{subfigure}%
    \begin{subfigure}{0.43\linewidth}
        \centering
        \includegraphics[width=\linewidth]{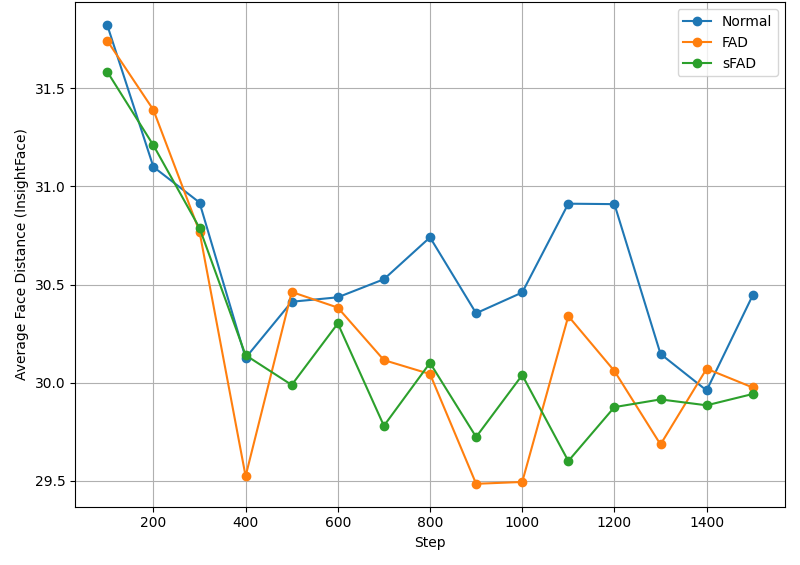}
        \caption{reeves}
    \end{subfigure}%

    \begin{subfigure}{0.43\linewidth}
        \centering
        \includegraphics[width=\linewidth]{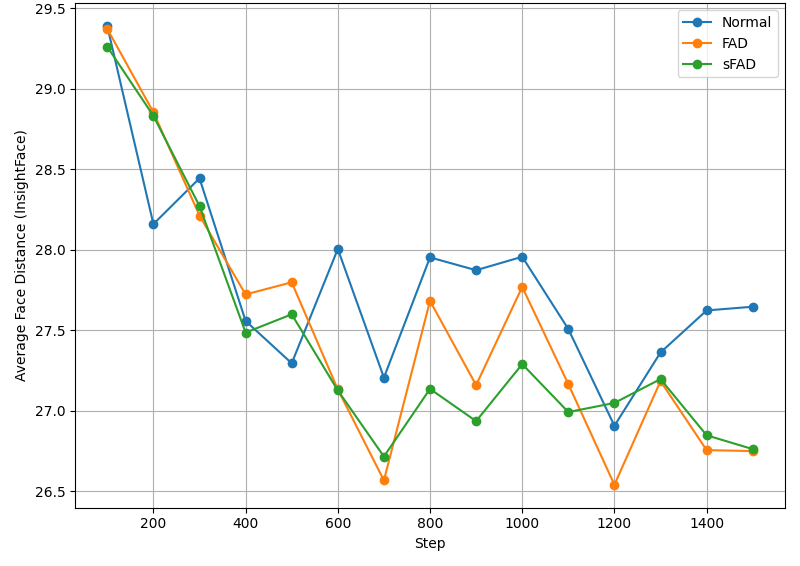}
        \caption{hsng}
    \end{subfigure}%
    \begin{subfigure}{0.43\linewidth}
        \centering
        \includegraphics[width=\linewidth]{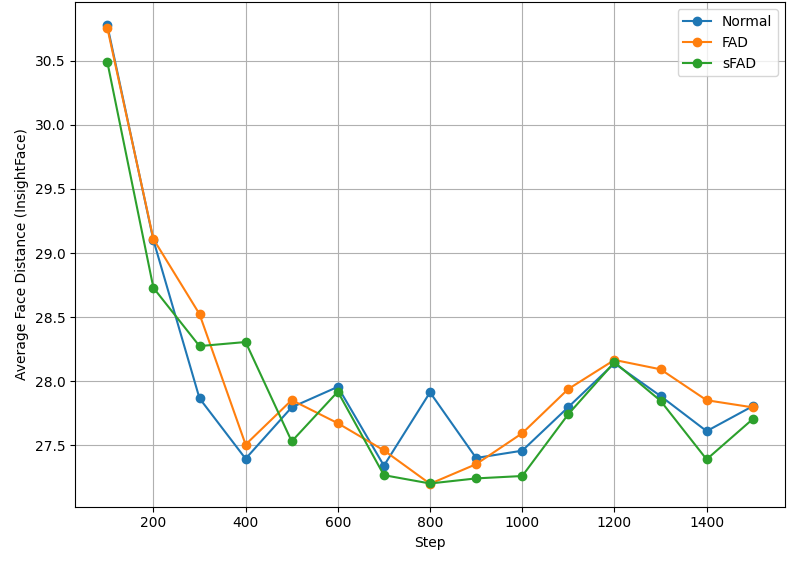}
        \caption{mbst}
    \end{subfigure}%
    \caption{Comparison of InsightFace ($\downarrow$) scores of three methods across all steps for four human datasets, extracted from \textbf{SD 1.5}.}
    \label{fig:insightface_comparison_15}
\end{figure}

\begin{figure}[htb]
    \centering
    \begin{subfigure}{0.45\linewidth}
        \centering
        \includegraphics[width=\linewidth]{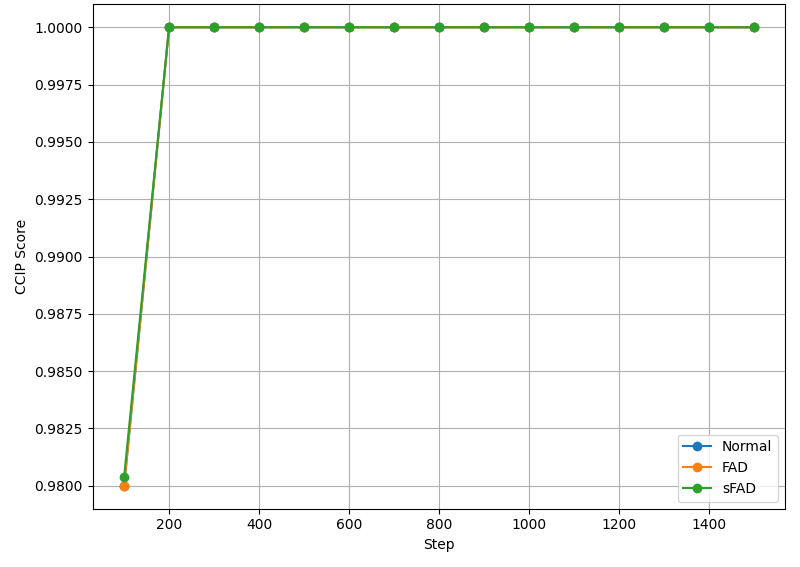}
        \caption{pikachu}
    \end{subfigure}
    \hspace{0.05\linewidth}
    \begin{subfigure}{0.45\linewidth}
        \centering
        \includegraphics[width=\linewidth]{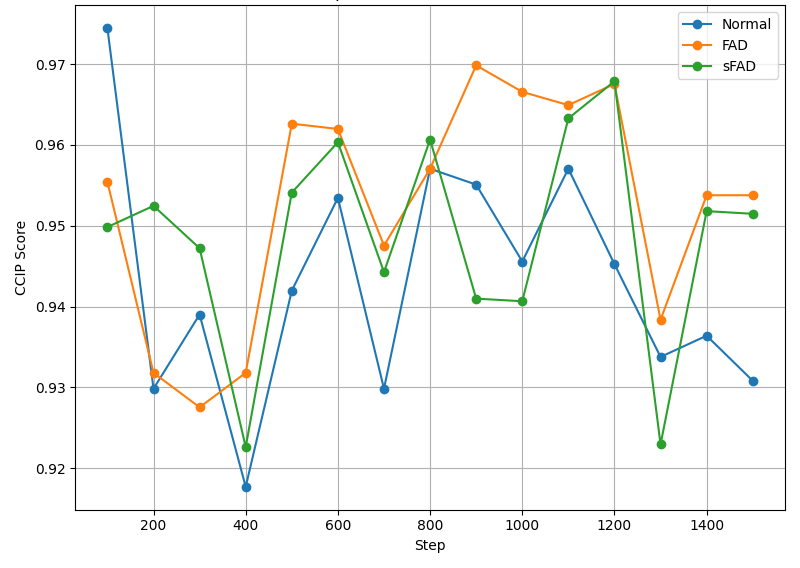}
        \caption{pochacco}
    \end{subfigure}

    \caption{Comparison of CCIP ($\uparrow$) scores of three methods across all steps for two character datasets, extracted from \textbf{SD 1.5}.}
    \label{fig:ccip_comparison_15}
\end{figure}

\begin{figure}[htb]
    \centering
    \begin{subfigure}{0.43\linewidth}
        \centering
        \includegraphics[width=\linewidth]{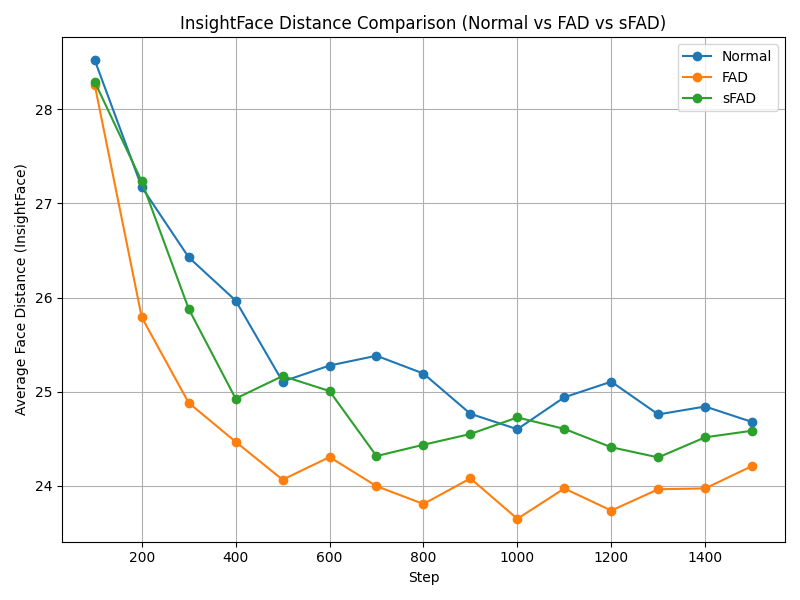}
        \caption{faker}
    \end{subfigure}%
    \begin{subfigure}{0.43\linewidth}
        \centering
        \includegraphics[width=\linewidth]{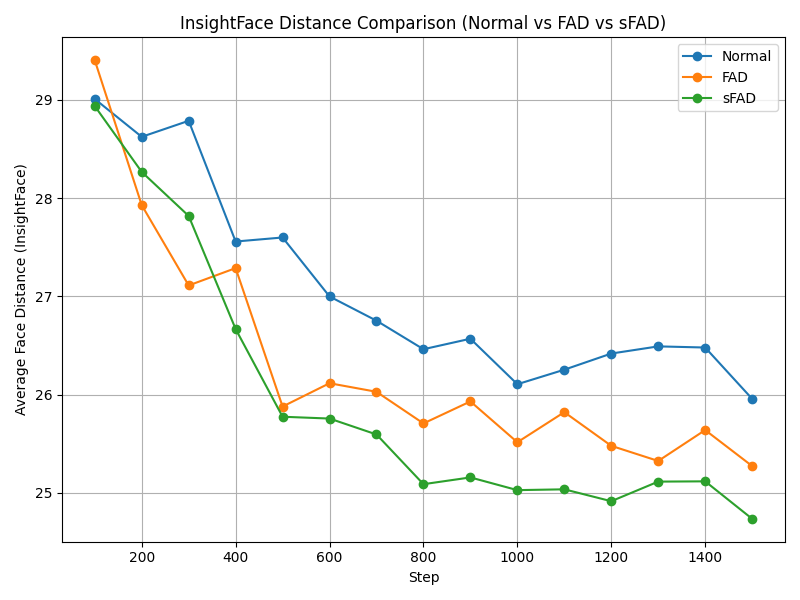}
        \caption{reeves}
    \end{subfigure}%

    \begin{subfigure}{0.43\linewidth}
        \centering
        \includegraphics[width=\linewidth]{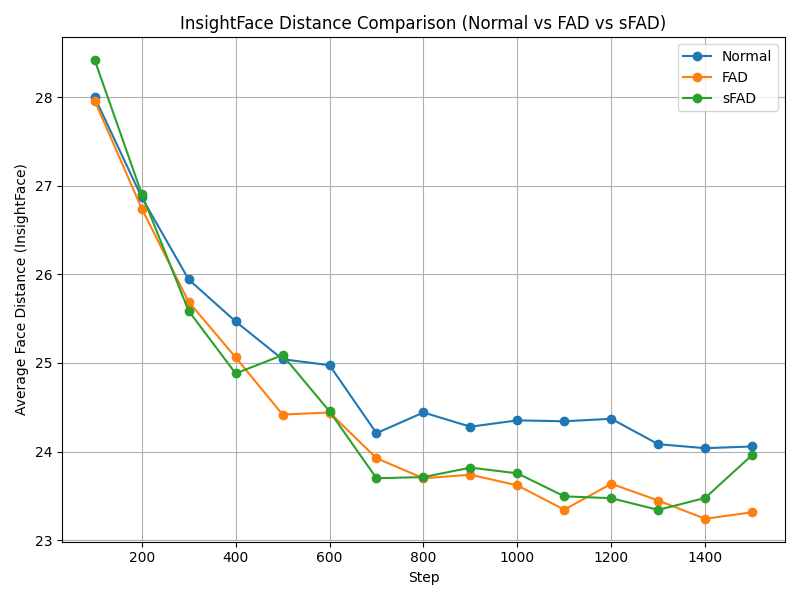}
        \caption{hsng}
    \end{subfigure}%
    \begin{subfigure}{0.43\linewidth}
        \centering
        \includegraphics[width=\linewidth]{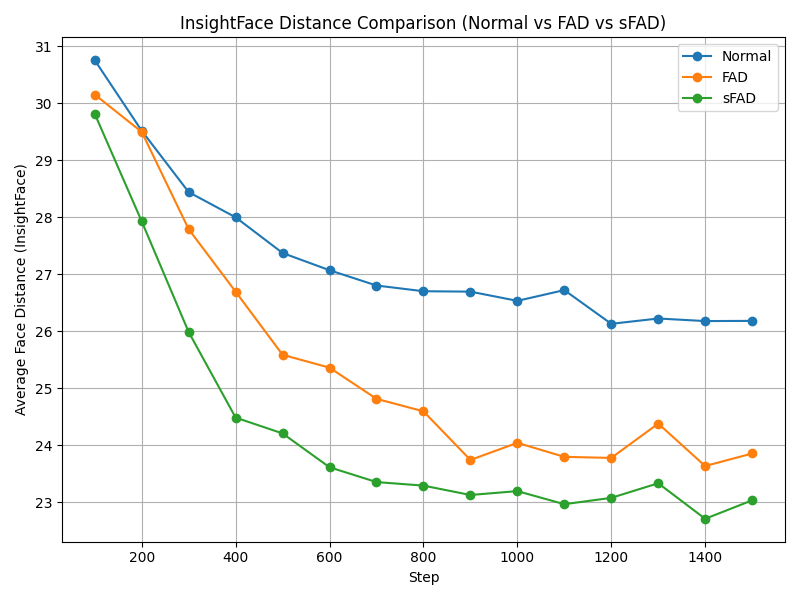}
        \caption{mbst}
    \end{subfigure}%
    \caption{Comparison of InsightFace ($\downarrow$) scores of three methods across all steps for four human datasets, extracted from \textbf{SD XL}.}
    \label{fig:insightface_comparison_xl}
\end{figure}

\begin{figure}[htb]
    \centering
    \begin{subfigure}{0.45\linewidth}
        \centering
        \includegraphics[width=\linewidth]{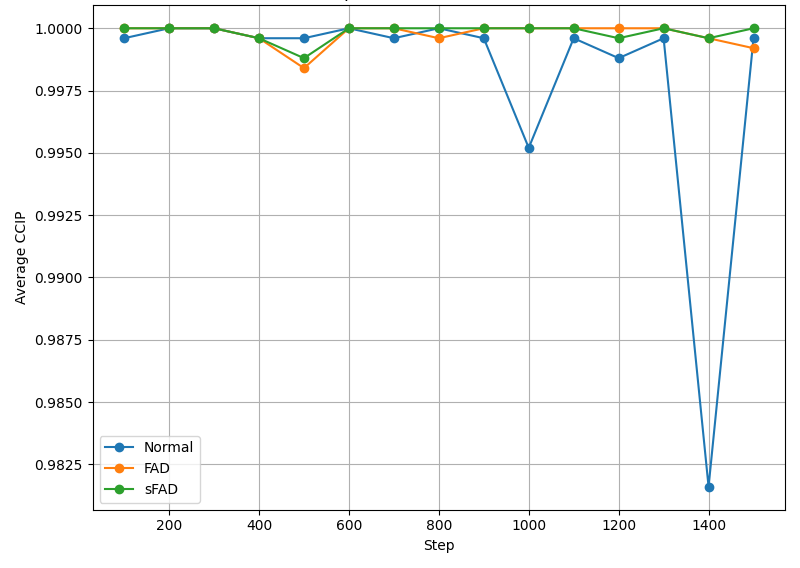}
        \caption{pikachu}
    \end{subfigure}
    \hspace{0.05\linewidth}
    \begin{subfigure}{0.45\linewidth}
        \centering
        \includegraphics[width=\linewidth]{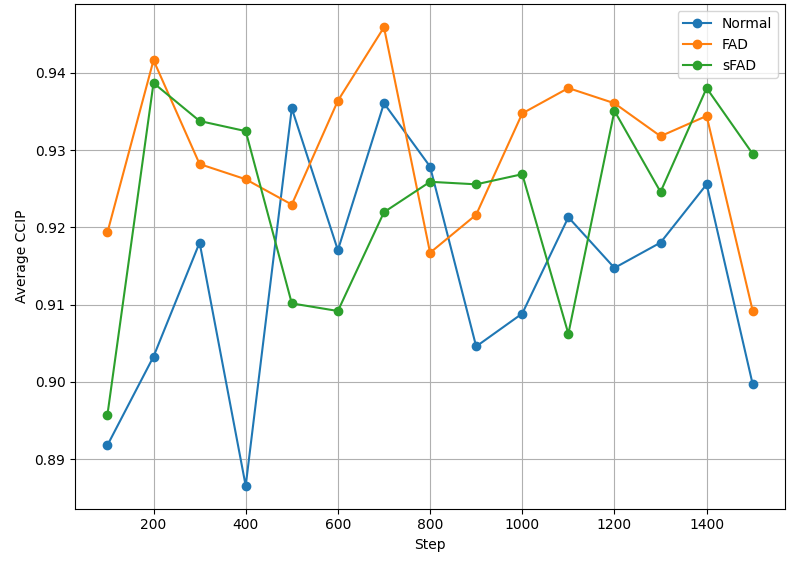}
        \caption{pochacco}
        \end{subfigure}

    \caption{Comparison of CCIP ($\uparrow$) scores of three methods across all steps for two character datasets, extracted from \textbf{SDXL}.}
    \label{fig:ccip_comparison_xl}
\end{figure}

We present graphs comparing the scores of Normal Dropout, FAD, and sFAD across all datasets over 100 to 1500 training steps for the metrics FID, DINO, InsightFace and CCIP.
Overall, FAD and sFAD consistently achieve better scores than Normal Dropout, although the degree of improvement varies across datasets.
The curves exhibit significant fluctuations depending on the dataset, and there are intervals where the ranking of scores swap or score drops, indicating potential overfitting during training. Graphs are shown in \cref{fig:FID_comparison_15,fig:FID_comparison_xl} for FID, \cref{fig:DINO_comparison_15,fig:DINO_comparison_xl} for DINO, \cref{fig:insightface_comparison_15,fig:insightface_comparison_xl} for InsightFace, and \cref{fig:ccip_comparison_15,fig:ccip_comparison_xl} for CCIP. However, CCIP scores appear to be saturated and less discriminative because the CCIP model already has strong prior knowledge of characters such as Pikachu. This makes CCIP less reliable for evaluating subtle differences in character reproduction for these datasets.

\begin{table*}[htbp]
      \centering
      \caption{The full version of evaluation prompts for comparative evaluation with GPT-4.1}
      \begin{tabular}{c}
        \toprule
        \multicolumn{1}{c}{%
          \begin{minipage}{0.95\linewidth}
            \centering
            \includegraphics[width=0.15\linewidth]{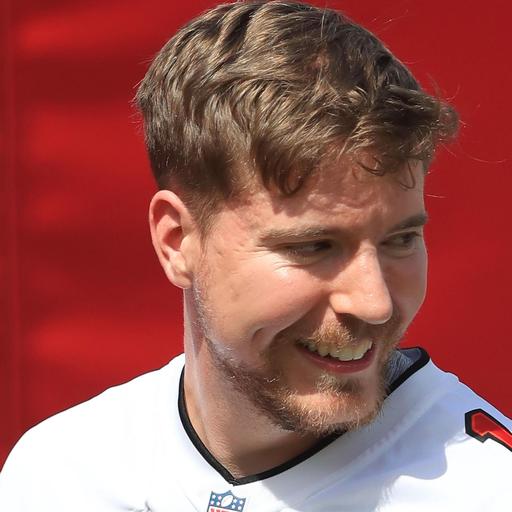}\hspace{0.03\linewidth}
            \includegraphics[width=0.15\linewidth]{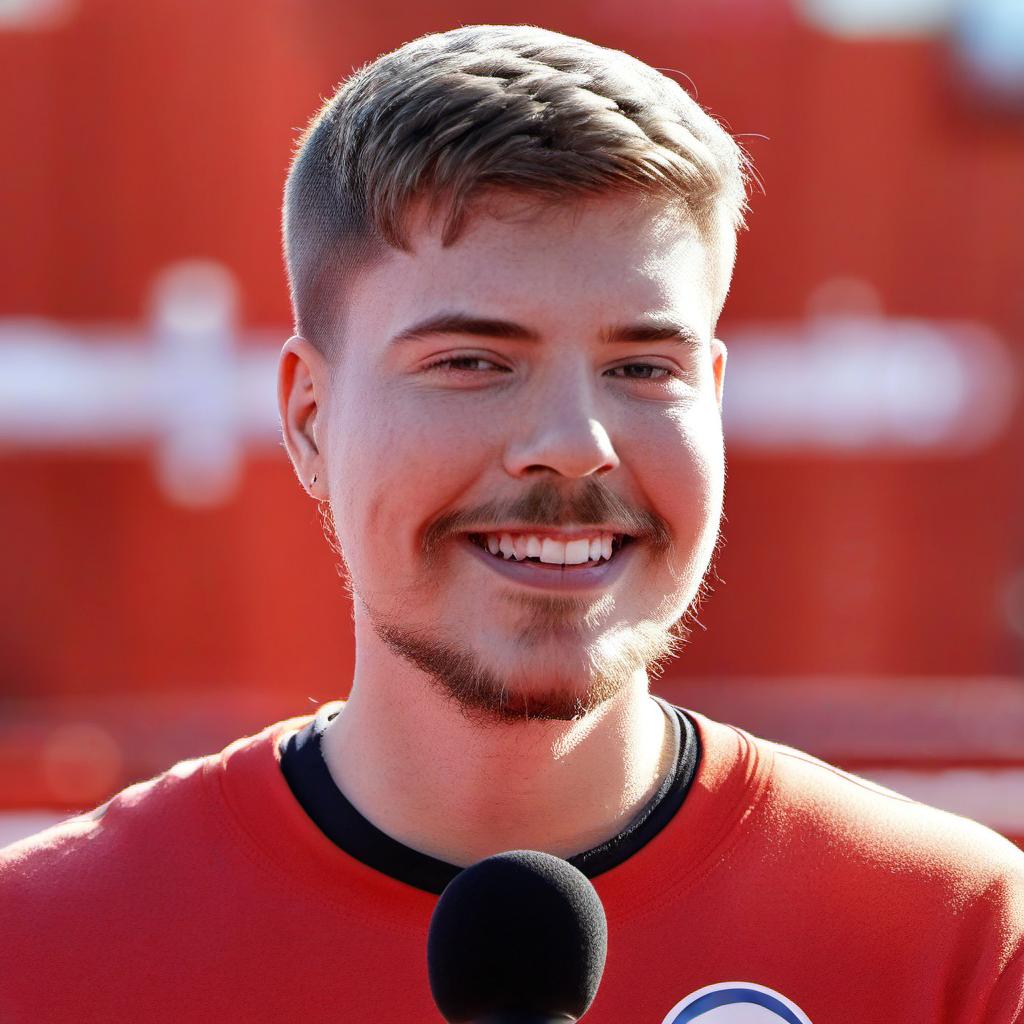}
          \end{minipage}
        } \\
        \midrule
        \multicolumn{1}{p{0.95\linewidth}}{
          \begin{minipage}[t]{0.95\linewidth}
            \textbf{Evaluation Prompt:} \\
            You will be shown TWO images: \\
            \textbf{- Image 1}: the ground truth character image (reference only) \\
            \textbf{- Image 2}: a generated image that should depict the same character.
            \\[0.5em]
            Your task is to critically evaluate how well \textbf{Image 2} resembles \textbf{Image 1},
            considering the character identity and overall image quality.
            \\[0.5em]
            \textbf{Ground Truth Character key features (from Image 1)}: \\
            Character (mbst): mbst, beard, brown hair, short hair
            \\[0.8em]
            \textbf{Evaluation Dimensions} \\
            \underline{Character Similarity}:
            \begin{itemize}
                \item How closely does the generated character (Image 2) resemble the reference character (Image 1)?
                \item Consider race, facial structure, hairstyle, eye color, and other distinctive features.
            \end{itemize}
            \underline{Composition \& Image Quality}:
            \vspace{0.2em}

            Assess technical and artistic quality of Image 2, including:
            \begin{itemize}
                \item Composition coherence
                \item Deformities (e.g., extra limbs, distorted face)
                \item Texture, lighting, color accuracy, clarity
            \end{itemize}
            \vspace{0.5em}
            \textbf{Scoring Criteria} \\
            \underline{Character Similarity (10-point scale)}:
            \vspace{0.2em}

            10 = highly similar, 1 = completely different
            \begin{itemize}
                \item Deduct 2 points for incorrect race or major facial mismatch
                \item Deduct 1 point for missing or incorrect key features (e.g., hair/eye color)
            \end{itemize}
            \underline{Composition \& Image Quality (10-point scale)}:
            \vspace{0.2em}

            10 = excellent, 1 = very poor
            \begin{itemize}
                \item Deduct 2 points for major deformities
                \item Deduct 1 point for lighting/color/texture issues or minor artifacts
            \end{itemize}
            \vspace{0.5em}
            \textbf{Output Format (strict)} \\
            Image2 Evaluation: \\
            - [Concise bullet points listing observed issues or mismatches]
            \vspace{0.2em}

            Scores: \\
            Character Similarity: [score] / 10 \\
            Composition \& Image Quality: [score] / 10
            \\[0.8em]
            \textbf{Notes}
            \begin{itemize}
                \item Do \textbf{not} evaluate Image 1.
                \item Be critical and specific.
                \item Start your answer immediately after this instruction.
            \end{itemize}
            \vspace{0.2em}
            \end{minipage}
        } \\
    \bottomrule
  \end{tabular}
  \label{tab:gpt4v_eval_mbst}
\end{table*}

\begin{table*}[htbp]
    \centering
    \caption{The full version of evaluation result for comparative evaluation with GPT-4.1}
    \begin{tabular}{cc}
        \toprule
        \multicolumn{2}{c}{%
            \includegraphics[width=0.15\linewidth]{figures/images/mbst_gt.jpg}\hspace{0.03\linewidth}
            \includegraphics[width=0.15\linewidth]{figures/images/mbst_adaptive.jpg}
        }\\
        \midrule
        \multicolumn{2}{p{0.95\linewidth}}{
            \begin{minipage}[t]{\linewidth}
            \textbf{Image 2 Evaluation:}

            \underline{Character Similarity}:
            \begin{itemize}
                \item Hairstyle is similar (short, brown hair, styled to the side), but appears slightly neater and more stylized than in Image 1 \textbf{(\textcolor{red}{-1 points})}.
                \item Beard is present, but the style is somewhat different \textbf{(\textcolor{red}{-1 points})}.
                \item General facial structure is fairly close but the face in Image 2 appears slimmer and more angular \textbf{(\textcolor{red}{-1 points})}.
            \end{itemize}

            \underline{Composition \& Image Quality}:
            \begin{itemize}
                \item No major issues with race or overall appearance.
                \item Image 2 shows the person holding a microphone, altering focus compared to the original \textbf{(\textcolor{red}{-1 points})}.
                \item No major deformities or artifacts detected.
                \item Lighting and texture are clean, but the image feels more artificial and softened, losing some realism compared to the reference.
            \end{itemize}
            \end{minipage}
        } \\

        \multicolumn{2}{p{0.95\linewidth}}{
            \vspace{0.5em}
            \begin{minipage}[t]{\linewidth}
            \textbf{Scores:} \\
            Character Similarity: \textbf{\textcolor{blue}{7 / 10}} \quad
            Composition \& Image Quality: \textbf{\textcolor{blue}{9 / 10}}
            \end{minipage}
        } \\
        \bottomrule
    \end{tabular}
    \label{tab:gpt4v_eval_result_mbst}
\end{table*}

\end{document}